\newcommand\mypound{\scalebox{1}{\raisebox{0.1ex}{\#}}}

\documentclass[10pt,journal,compsoc]{IEEEtran}
\ifCLASSOPTIONcompsoc
  \usepackage[nocompress]{cite}
\else
  \usepackage{cite}
\fi
\ifCLASSINFOpdf
   \usepackage[pdftex]{graphicx}
\else
   \usepackage[dvips]{graphicx}
\fi
\usepackage{amssymb}% http://ctan.org/pkg/amssymb
\usepackage{pifont}% http://ctan.org/pkg/pifont
\newcommand{\cmark}{\ding{51}}%
\newcommand{\xmark}{\ding{55}}%

\usepackage{amsmath}
\usepackage{bbold}
\usepackage{array}
\usepackage{xcolor}
 \usepackage[caption=false,font=footnotesize,labelfont=sf,textfont=sf]{subfig}

\usepackage{mathtools}

\usepackage[colorlinks=true,linkcolor=black,urlcolor=black,bookmarksopen=true]{hyperref}

\usepackage{bookmark}

\usepackage{babel,blindtext}

\usepackage{enumitem}
\usepackage{titlesec}
\usepackage{multirow}

\usepackage{makecell}

\usepackage{amsthm}

\usepackage[parfill]{parskip}

\newcommand{\cmmnt}[1]{}

\newtheorem{definition}{Definition}

\titleformat{\paragraph}
{\normalfont\normalsize\bfseries}{\theparagraph}{1em}{}
\titlespacing*{\paragraph}
{0pt}{3.25ex plus 1ex minus .2ex}{1.5ex plus .2ex}

\newcommand{\norm}[1]{\left\lVert#1\right\rVert}

\begin{document}
%
% paper title
% Titles are generally capitalized except for words such as a, an, and, as,
% at, but, by, for, in, nor, of, on, or, the, to and up, which are usually
% not capitalized unless they are the first or last word of the title.
% Linebreaks \\ can be used within to get better formatting as desired.
% Do not put math or special symbols in the title.

% \title{Action Detection in Untrimmed Videos with Deep Learning Models: A Survey}
%
\title{Deep Learning-based Action Detection in Untrimmed Videos: A Survey}

% \title{The Spectrum of Supervision for Action Detection in Untrimmed Videos: A Survey}

%
% author names and IEEE memberships
% note positions of commas and nonbreaking spaces ( ~ ) LaTeX will not break
% a structure at a ~ so this keeps an author's name from being broken across
% two lines.
% use \thanks{} to gain access to the first footnote area
% a separate \thanks must be used for each paragraph as LaTeX2e's \thanks
% was not built to handle multiple paragraphs
%
%
%\IEEEcompsocitemizethanks is a special \thanks that produces the bulleted
% lists the Computer Society journals use for "first footnote" author
% affiliations. Use \IEEEcompsocthanksitem which works much like \item
% for each affiliation group. When not in compsoc mode,
% \IEEEcompsocitemizethanks becomes like \thanks and
% \IEEEcompsocthanksitem becomes a line break with idention. This
% facilitates dual compilation, although admittedly the differences in the
% desired content of \author between the different types of papers makes a
% one-size-fits-all approach a daunting prospect. For instance, compsoc 
% journal papers have the author affiliations above the "Manuscript
% received ..."  text while in non-compsoc journals this is reversed. Sigh.

\author{Elahe~Vahdani and
        Yingli~Tian${}^\ast$,~\IEEEmembership{Fellow,~IEEE}

\IEEEcompsocitemizethanks{\IEEEcompsocthanksitem  E. Vahdani is with the Department of Computer Science, The Graduate Center, The City University of New York, NY, 10016.  E-mail: evahdani@gradcenter.cuny.edu \protect\\ 
\IEEEcompsocthanksitem Y. Tian is with the Department of Electrical Engineering, The City College, and the Department of Computer Science, the Graduate Center, the City University of New York, NY, 10031. E-mail:ytian@ccny.cuny.edu ${}^\ast$Corresponding author}
% note need leading \protect in front of \\ to get a newline within \thanks as
% \\ is fragile and will error, could use \hfil\break instead.
% <-this % stops a space
\thanks{This material is based upon work supported by the National Science Foundation under award number IIS-2041307.}}

% note the % following the last \IEEEmembership and also \thanks - 
% these prevent an unwanted space from occurring between the last author name
% and the end of the author line. i.e., if you had this:
% 
% \author{....lastname \thanks{...} \thanks{...} }
%                     ^------------^------------^----Do not want these spaces!
%
% a space would be appended to the last name and could cause every name on that
% line to be shifted left slightly. This is one of those "LaTeX things". For
% instance, "\textbf{A} \textbf{B}" will typeset as "A B" not "AB". To get
% "AB" then you have to do: "\textbf{A}\textbf{B}"
% \thanks is no different in this regard, so shield the last } of each \thanks
% that ends a line with a % and do not let a space in before the next \thanks.
% Spaces after \IEEEmembership other than the last one are OK (and needed) as
% you are supposed to have spaces between the names. For what it is worth,
% this is a minor point as most people would not even notice if the said evil
% space somehow managed to creep in.

% The paper headers
\markboth{}%
{Shell \MakeLowercase{\textit{et al.}}: Bare Advanced Demo of IEEEtran.cls for IEEE Computer Society Journals}
% The only time the second header will appear is for the odd numbered pages
% after the title page when using the twoside option.
% 
% *** Note that you probably will NOT want to include the author's ***
% *** name in the headers of peer review papers.                   ***
% You can use \ifCLASSOPTIONpeerreview for conditional compilation here if
% you desire.

% The publisher's ID mark at the bottom of the page is less important with
% Computer Society journal papers as those publications place the marks
% outside of the main text columns and, therefore, unlike regular IEEE
% journals, the available text space is not reduced by their presence.
% If you want to put a publisher's ID mark on the page you can do it like
% this:
%\IEEEpubid{0000--0000/00\$00.00~\copyright~2015 IEEE}
% or like this to get the Computer Society new two part style.
%\IEEEpubid{\makebox[\columnwidth]{\hfill 0000--0000/00/\$00.00~\copyright~2015 IEEE}%
%\hspace{\columnsep}\makebox[\columnwidth]{Published by the IEEE Computer Society\hfill}}
% Remember, if you use this you must call \IEEEpubidadjcol in the second
% column for its text to clear the IEEEpubid mark (Computer Society journal
% papers don't need this extra clearance.)

% use for special paper notices
%\IEEEspecialpapernotice{(Invited Paper)}

% for Computer Society papers, we must declare the abstract and index terms
% PRIOR to the title within the \IEEEtitleabstractindextext IEEEtran
% command as these need to go into the title area created by \maketitle.
% As a general rule, do not put math, special symbols or citations
% in the abstract or keywords.

\IEEEtitleabstractindextext{%

\begin{abstract}
% general motive
Understanding human behavior and activity facilitates advancement of numerous real-world applications, and is critical for video analysis. 
% comparison with action recognition 
Despite the progress of action recognition algorithms in trimmed videos, the majority of real-world videos are lengthy and untrimmed with sparse segments of interest. 
% define the problem
The task of temporal activity detection in untrimmed videos aims to localize the temporal boundary of actions and classify the action categories. 
% supervision level
Temporal activity detection task has been investigated in full and limited supervision settings depending on the availability of action annotations. 
% this literature
This paper provides an extensive overview of deep learning-based algorithms to tackle temporal action detection in untrimmed videos with different supervision levels including fully-supervised, weakly-supervised, unsupervised, self-supervised, and semi-supervised. In addition, this paper also reviews advances in spatio-temporal action detection where actions are localized in both temporal and spatial dimensions. 
Moreover, the commonly used action detection benchmark datasets and evaluation metrics are described, and the performance of the state-of-the-art methods are compared.
Finally, real-world applications of temporal action detection in untrimmed videos and a set of future directions are discussed.

\end{abstract}

% Note that keywords are not normally used for peerreview papers.
\begin{IEEEkeywords}
Action Understanding, Temporal Action Detection, Untrimmed Videos, Deep Learning, Full and Limited Supervision.  
\end{IEEEkeywords}}

% make the title area
\maketitle

% To allow for easy dual compilation without having to reenter the
% abstract/keywords data, the \IEEEtitleabstractindextext text will
% not be used in maketitle, but will appear (i.e., to be "transported")
% here as \IEEEdisplaynontitleabstractindextext when compsoc mode
% is not selected <OR> if conference mode is selected - because compsoc
% conference papers position the abstract like regular (non-compsoc)
% papers do!
\IEEEdisplaynontitleabstractindextext
% \IEEEdisplaynontitleabstractindextext has no effect when using
% compsoc under a non-conference mode.

% For peer review papers, you can put extra information on the cover
% page as needed:
% \ifCLASSOPTIONpeerreview
% \begin{center} \bfseries EDICS Category: 3-BBND \end{center}
% \fi
%
% For peerreview papers, this IEEEtran command inserts a page break and
% creates the second title. It will be ignored for other modes.
\IEEEpeerreviewmaketitle

\ifCLASSOPTIONcompsoc
\IEEEraisesectionheading{\section{Introduction }\label{sec:introduction}}
\else
\section{Introduction}
\label{sec:introduction}
\fi

This paper provides a comprehensive overview of temporal action detection. This task aims to detect the start and end of action instances in long untrimmed videos and predict the action categories. Temporal action detection is crucial for many video analysis applications such as sports analysis, autonomous driving, anomaly detection in surveillance cameras, understanding  instructional videos, etc. Learning with limited supervision is a scheme where annotations of actions are unavailable or only partially available during the training phase. Because annotation of long untrimmed videos is very time-consuming, designing action detection methods with limited supervision has been very popular. This survey reviews temporal action detection methods with full and limited supervision signals.

% {\color{red} brief summary, high-level motivation guidance }

\begin{figure}[t]
\centering
\includegraphics[scale=0.35]{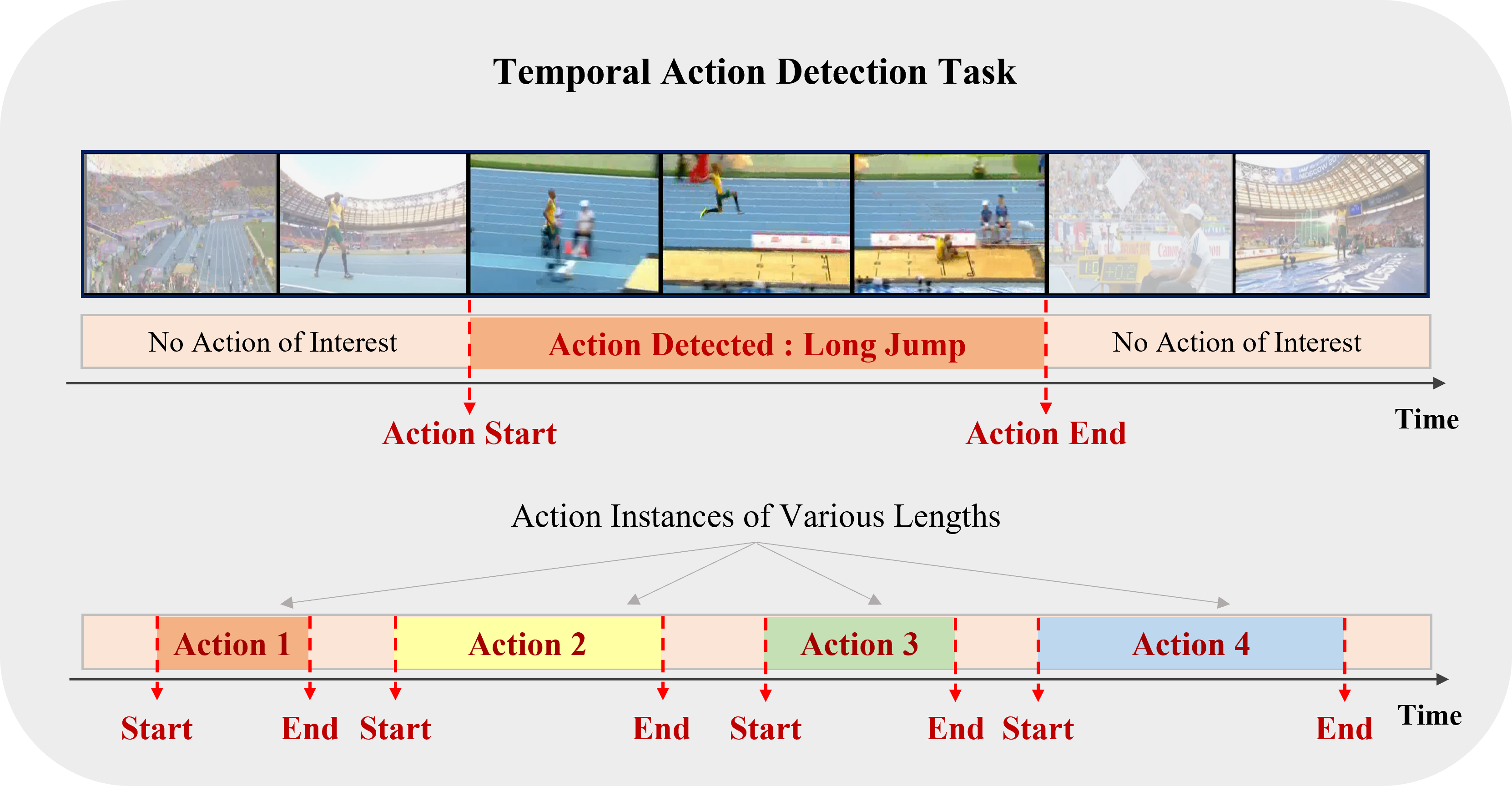}
\caption{Temporal action detection aims to localize action instances in time and recognize their categories. The first row demonstrates an example of action ``long jump'' detected in an untrimmed video from THUMOS14 dataset \cite{jiang2014thumos}. The second row is an example of an untrimmed video including several action instances of interest with various lengths.}
\label{fig:tasks}
\end{figure}

\subsection{Motivation}

% \textit{Video Understanding}. 
Social networks and digital cameras have led to substantial video and media content produced by individuals each day. Hence, video understanding and analysis continues to be one of the essential research subjects in computer vision. While deep learning has accomplished remarkable performance in many computer vision tasks, video understanding is still far from ideal. 
Action understanding, notably, as a vital element of video analysis, facilitates the advancement of numerous real-world applications. For instance,
collaborative robots need to recognize how the human partner completes the job to cope with the variations in the task \cite{rea2019human}. Sport analysis systems must comprehend game actions to report commentaries of live activities \cite{cioppa2020context}. Autonomous driving cars demand an understanding of operations performed by the surrounding cars and pedestrians \cite{rasouli2019autonomous}.

In this paper, we define \textit{trimmed videos} as pre-segmented video clips that each contains only one action instance. In other words, the \textit{context} of the action, i.e., moments before or after the action are not included in the video. Therefore, action detection in trimmed videos only need to classify the action categories without the need to detect starting and ending timestamps.  Recognizing actions in trimmed videos has many applications in video surveillance, robotics, medical diagnosis \cite{herath2017going}, and has achieved excellent performance in recent years \cite{feichtenhofer2019slowfast, ghadiyaram2019large,duan2020omni}.
% , obtaining nearly $80\%$ accuracy 
%  on benchmark datasets \cite{kay2017kinetics}. 
However, the majority of videos \textit{in the wild}, i.e., recorded in unconstrained environments, are naturally untrimmed. \textit{Untrimmed videos} are lengthy unsegmented videos that may include several action instances, the moments before or after each action, and the transition from one action to another. The action instances in one video can belong to several action classes and have different duration. 

% Even though recognizing actions in trimmed videos has achieved excellent performance in recent years, it is not a realistic task in common video analysis applications. Humans perceive the world continuously, and the majority of real-world videos are untrimmed. 

% The first row shows several instances of action "long jump" with different duration performed by different people in the video. The orange boxes are \textit{temporal backgrounds} that consist of moments before the jump (preparing), after the jump (finishing) and diverse activities between the jump actions such as crowd cheering in the stadium. The second row shows several frames sampled from the moments before, within, and after the action ``long jump''.

% of jumping sampled from 
% within the annotated boundaries as well as frames before the start and after the end. 

\textit{Temporal activity detection} in untrimmed videos aims to localize the action instances in time and recognize their categories. This task is considerably more complicated than action recognition which merely seeks to classify the categories of trimmed video clips. Fig. \ref{fig:tasks} shows an example of temporal activity detection in an untrimmed video recorded in a stadium. The first row demonstrates the detection of action "long jump" in temporal domain where the start and end time of the action are localized. The goal is to only detect \textit{the actions of interest}, i.e., actions that belong to a predefined set of action classes. The temporal intervals of other activities that do not belong to this set of actions are called \textit{temporal background}. For example, the segments right before or right after action "long jump" may belong to other diverse activities such as crowd cheering in the stadium. In some cases, the frames right before or right after an action are visually very similar to the start or end of the action which makes the localization of action intervals very challenging. Another challenge (as shown in the second row of Fig. \ref{fig:tasks}) is that action instances may occur at any time of the video and have various duration, lasting from less than a second to several minutes \cite{idrees2017thumos}.

\begin{figure}[!t]
    \centering
    \includegraphics[scale=0.35]{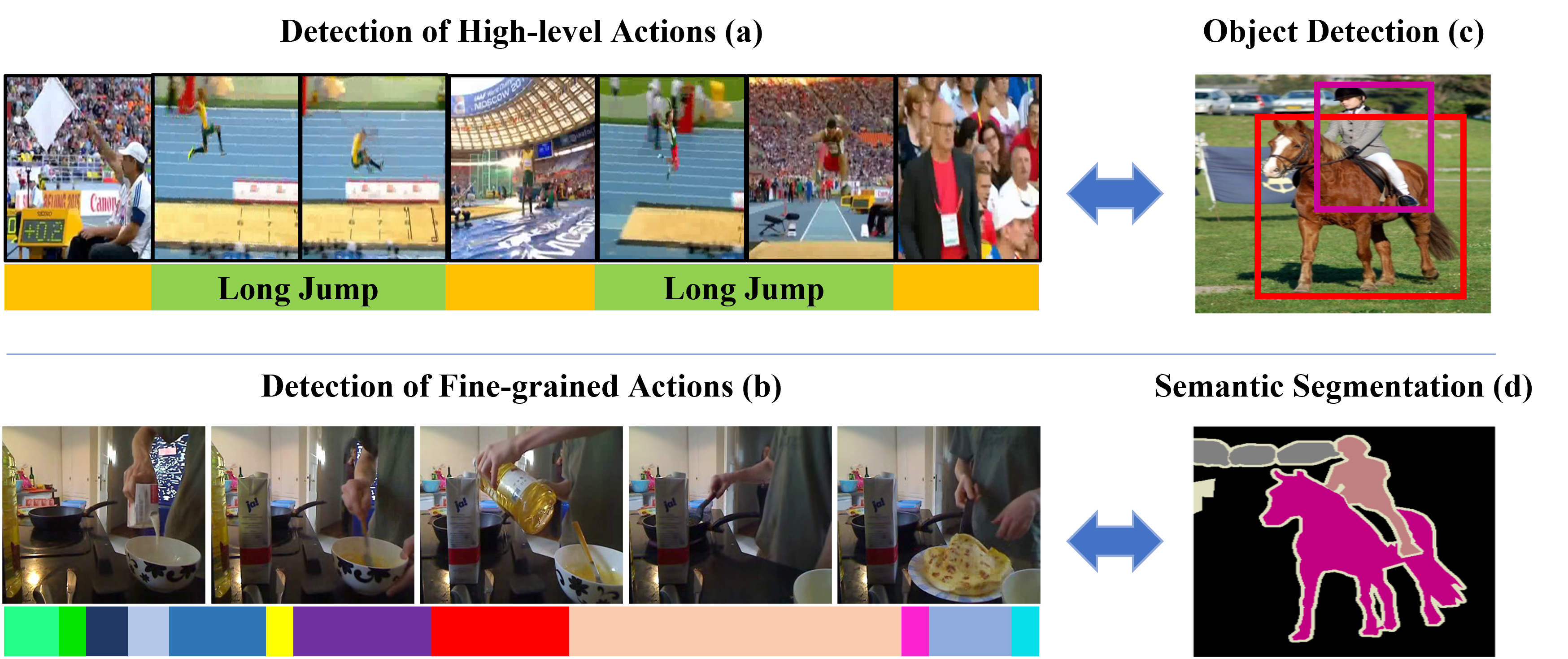}
        \caption{Task Relations: (a) Temporal detection of action ``Long Jump'' on THUMOS14 \cite{jiang2014thumos}. (b) Temporal detection (segmentation) of fine-grained actions shown by different colors in a ``making pancakes'' video on Breakfast \cite{kuehne2014language}. 
         (c) and (d) Results from \cite{long2015fully} on PASCAL \cite{everingham2011pascal}.}
    \label{fig:problems_relations}
\end{figure}

Temporal action detection mainly targets activities of high-level semantics and videos with a sparse set of actions (e.g., actions only cover $30\%$ of the frames in \cite{gorban2015thumos}). However, in some cases, the goal is to predict action labels at every frame of the video. In such cases, the task is referred to as \textit{temporal action segmentation} which targets the fine-grained actions and videos with dense occurrence of actions ($93\%$ of the frames in \cite{kuehne2014language}). One can convert between a given segmentation and a set of detected instances in the temporal domain by simply adding or removing temporal background segments \cite{lea2017temporal}. Temporal action detection similar to object detection belongs to the family of detection problems. Both of these problems aim to localize the instances of interest, i.e., action intervals in temporal domain versus object bounding boxes in spatial domain, Fig. \ref{fig:problems_relations} (a and c). When targeting fine-grained actions, temporal action detection (segmentation) is similar to semantic segmentation as both aim to classify every single instance, i.e., frames in temporal domain versus pixels in spatial domain, Fig. \ref{fig:problems_relations} (b and d). As a result, many techniques for temporal action detection and segmentation are inspired by the advancements in object detection and semantic segmentation \cite{chao2018rethinking,shou2017cdc,gao2019video}.

Action detection has drawn much attention in recent years and has broad applications in video analysis tasks. As surveillance cameras are increasingly deployed in many places, the demand for anomaly detection has also surged. Anomalous events such as robbery or accidents occur less frequently compared with normal activities and it can be very time-consuming to detect such events by humans. Therefore, automatic detection of suspicious events has a great advantage. By growing popularity of social media many people follow online tutorials and \textit{instructional videos} to learn how to perform a task such as ``changing the car tire'' properly for the first time. The instructional videos are usually untrimmed and include several steps of the main task, e.g., ``jack up the car'' and ``put on the tire'' for changing the tire. Automatic segmentation of these videos to the main action steps can facilitate and optimize the learning process. Another application is in sport video analysis to localize the salient actions and highlights of a game and analyze the strategies of specific teams. Furthermore, action detection has a critical role in self-driving cars to analyze the behavior of pedestrians, cyclists, and other surrounding vehicles to make safe autonomous decisions.

% Action detection has recently received significant attention because of its applications in various video analysis tasks such as understanding instructional videos, anomaly detection in surveillance Videos, action spotting in sports, and detection in self-driving cars. 

% Because action instances can happen at any moment of an untrimmed video, designing activity detection algorithms is required to find these instances. 
% lengthy with sparse segments of interests. 
% An untrimmed video may include several actions of interest with different duration that can occur at any moment of the video.

\subsection{Taxonomy}
\label{Taxonomy}

% This paper provides an extensive review of temporal action detection with different supervision levels.
% We begin by introducing the tasks here.

To the best of our knowledge, this is the first comprehensive survey describing deep learning based algorithms for activity detection in untrimmed videos with different supervision levels. We describe the fully-supervised methods in Section \ref{Fully-Supervised Action Detection} and methods with limited supervision (weakly-supervised, unsupervised, self-supervised, and semi-supervised) in Section \ref{Action Detection with Limited Supervision}. Section \ref{Datasets and Evaluation} summarizes action detection benchmark datasets, evaluation metrics, and performance comparison between the-state-of-the-art methods. Finally, Section \ref{Discussions} discusses the most common real-world applications of action detection and possible future directions. We provide a brief introduction of the tasks here.

% and the main technical terms to facilitate reading the subsequent sections.

% In this section, we describe the tasks that are studied in this literature (also listed in Table \ref{table:tasks}) and introduce the main technical terms to facilitate reading the subsequent sections.

% \subsubsection{Tasks Descriptions}
% \label{Temporal Action Detection setting}

% Action detection is studied in temporal domain, i.e., localizing action instances in time and in spatio-temporal domain i.e., localizing action instances in time and space. 
% \paragraph{Temporal Action Detection}

\vspace{0.1in}
\textbf{Temporal action detection} aims to find the precise temporal boundary and label of action instances in untrimmed videos. Depending on annotation availability in train set, temporal action detection can be studied in the following settings (also listed in Table \ref{table:tasks}).  
\vspace{0.1in}

% \begin{itemize}[leftmargin=0.25in]
\begin{itemize}

\item \textbf{Fully supervised action detection:} Temporal boundaries and labels of action instances are available for training.

\vspace{0.1in}
\item \textbf{Weakly-supervised action detection:} Only the video-level labels of action instances are available. The order of action labels can be provided or not.

\vspace{0.1in}
\item \textbf{Unsupervised action detection:} There are no annotations for the action instances. 

\vspace{0.1in}
\item \textbf{Semi-supervised action detection:} The data is split to a small set $S_1$ and a large set $S_2$. The videos in $S_1$ are fully annotated (as in fully-supervised) while the videos in $S_2$ are either not annotated (unsupervised) or only annotated with video-level labels (as in weakly-supervised).

\vspace{0.1in}
\item \textbf{Self-supervised action detection:} A pretext task is defined to extract information from the data in an unsupervised setting by leveraging its structure. Then, this information is used to improve the performance for temporal action detection (downstream task) which can be supervised, unsupervised, or semi-supervised.   

\vspace{0.1in}
\item \textbf{Action detection with limited supervision:} Limited supervision is the opposite of full supervision where the annotations are unavailable or partially available. In this paper, we define limited supervision to include weakly-supervised, unsupervised, self-supervised, and semi-supervised settings as they are defined above.

% \vspace{0.1in}
% To the best of our knowledge, this is the first comprehensive survey describing deep learning based algorithms for activity detection in untrimmed videos with different supervision levels. We describe the fully-supervised methods in Section \ref{Fully-Supervised Action Detection} and methods with limited supervision (weakly-supervised, unsupervised, self-supervised, and semi-supervised) in Section \ref{Action Detection with Limited Supervision}. Section \ref{Datasets and Evaluation} summarizes action detection benchmark datasets, evaluation metrics, and performance comparison between the-state-of-the-art methods. Finally, Section \ref{Discussions} discusses the most common real-world applications of action detection and possible future directions.

\begin{table}[!htb]
\centering
\caption{Main categories of temporal action detection task with different supervision levels in training set. ``\cmark" indicates ``available"; ``\xmark" is for ``unavailable", and $\ast$ is ``partially available".
%The tasks are categorized based on annotation availability of action temporal boundaries and labels during training. %The full description of tasks is provided in section \ref{Temporal Action Detection setting}.
}
\resizebox{\columnwidth}{!}{%
\begin{tabular}{c | c | c  }
 \hline
Supervision Level   & Action Temporal Boundaries &  Action Labels\\
\hline
\hline
Fully-supervised & \cmark  & \cmark \\
Weakly-supervised & \xmark & \cmark  \\
Unsupervised & \xmark & \xmark \\
Semi-supervised & $\ast$ & $\ast$ \\
Self-supervised & \cmark $\ast$ \xmark  & \cmark $\ast$ \xmark \\
\hline
\end{tabular}%
}
\label{table:tasks}
\end{table}

% \begin{equation}
% \forall {\Psi}_i \in D_1, \hspace{0.1in} \exists \ \{ (t^i_{s,n},t^i_{e,n},l^i_n) \} _{n=1}^{N_i} 
% \end{equation}
% \begin{equation}
% \forall {\Psi}_j \in D_2,  \hspace{0.1in} \exists \  \{l^j_n\} _{n=1}^{N_j}  
% \end{equation}

% \vspace{0.1in}
% \item \textbf{Self-supervised action detection:}  
% There are no annotations for the action instances. Pseudo labels are automatically generated for a pretext task to improve the performance of temporal action detection.  %SSTDA\cite{chen2020action} is a self-Supervised temporal domain adaptation method for action segmentation.
\end{itemize}

\section{\text{Temporal Action Detection Methods}}
\label{Temporal-Action-Detection}

We begin this section by introducing important technical terms in Section \ref{Technical Terms}. Given an input video, \textit{video feature encoding} is necessary to extract representative visual features of the video (discussed in Section \ref{Video-Feature-Encoding}). 
Action detection methods with full supervision are described in Section \ref{Fully-Supervised Action Detection} and action detection methods with limited supervision are reviewed in Section \ref{Action Detection with Limited Supervision}.

\subsection{Term Definition}
\label{Technical Terms}

To facilitate reading subsequent sections, we define common terms, scores, and loss functions here.

\vspace{0.1in}

\begin{definition}
\label{Temporal Proposals}
\normalfont \textbf{Temporal action detection.} This task aims to find the precise temporal boundaries and categories of action instances in untrimmed videos. Annotation of an input video is denoted by ${\Psi}_g$ and includes a set of action instances as the following:

\begin{equation}
{\Psi}_g = \{ {\varphi}_n =  (t_{s,n},t_{e,n},l_n) \}_{n=1}^{N}, \label{tad-formula}
\end{equation}

where $N$ is the number of action instances, and ${\varphi}_n$ is the $n$-th action instance. The start time, end time, and label of ${\varphi}_n$ are denoted by $t_{s,n}$, $t_{e,n}$, and $l_n$, respectively. Label $l_n$ belongs to set $\{1,\cdots, C\}$, where $C$ is the number of action classes of interest in the whole dataset. The annotation ${\Psi}_g$ can be fully, partially, or not available for the videos of the training set.

\end{definition}

\begin{definition}
\label{Temporal Proposals}
\normalfont \textbf{Temporal proposals.} The temporal regions of input video that are likely to contain an action are called temporal proposals. Each temporal proposal $P_n$ is an interval identified with a starting time $t_{s,n}$, an ending time $t_{e,n}$, and a confidence score $c_n$. The confidence score is the predicted probability that the interval contains an action. Proposal $P_n$ can be formulated as:

\begin{equation}
P_n = (t_{s,n},t_{e,n}, c_n).
\end{equation}

\end{definition}

\begin{definition}
\label{Temporal IoU}
\normalfont \textbf{Temporal IoU (tIoU).} This is the ratio of temporal intersection over union between two temporal intervals. It is often measured between a predicted proposal (interval $I_p$) and its closest ground-truth action instance (interval $I_g$), formulated as:
\begin{equation}
    tIoU (I_p, I_g) = \frac{I_p \cap I_g}{I_p \cup I_g}.
\end{equation}

\end{definition}

\begin{definition}
\label{pos-neg-proposals}
\normalfont \textbf{Temporal proposal labeling.} Label of proposal $I_p$ is determined by ground-truth action instance $I_g$ that has the maximum tIoU with $I_p$. Let's denote the class label of $I_g$ with $c$. Then, depending on a predefined threshold $\sigma$, the proposal is declared as \textit{positive (true positive)} with label $c$ if $tIoU  \geq \sigma$. Otherwise, it is \textit{negative (or false positive)}. Also, if a ground-truth action instance is matched with several proposals, only the proposal with the highest confidence score is accepted as true positive and the others are declared as false positives. 

%Moreover, ground-truth action instances with no matching proposals are considered as \textit{false negative}. 
\end{definition}

% \begin{definition}
% \label{pos-neg-proposals}
% \normalfont \textbf{Positive and negative proposals.}
% To assign labels to proposals, the closest ground-truth instance with maximum tIoU is considered for each proposal. If the $i$-th proposal has a tIoU greater than a given threshold with a ground-truth instance of label $c$, then its labeled with class $c$, i.e., $l_i = c$. A positive proposal is a proposal that is labeled with an action instance, i.e., $l_i >0$. On the other hand, a negative proposal is labeled with background, i.e., $l_i =0$. During each mini-batch, the number of positive proposals is denoted by $N_{\text{pos}}$. 
% \end{definition}

% \begin{definition}
% \label{}
% \normalfont \textbf{}
% \end{definition}

\begin{definition}
\label{Precision and recall}
\normalfont \textbf{Precision and recall for proposal generation}. Precision is the ratio of true positive proposals to the total number of predicted proposals. Precision must be high to avoid producing exhaustively many irrelevant proposals. Recall is the ratio of true positive proposals to the total number of ground-truth action instances. Recall must be high to avoid missing ground-truth instances.
\end{definition}

% \begin{definition}
% \label{Temporal Proposal Evaluation}
% \normalfont \textbf{Temporal Proposal Evaluation}. The quality of temporal proposals can be measured by considering their temporal IoU (definition \ref{Temporal IoU}) with ground-truth action instances. The temporal proposals are expected to have high precision and recall (definition \ref{Precision and recall}), and reliable confidence scores, and the proposal generation process must be computationally efficient. 
% \end{definition}

\begin{definition}
\label{actionness-probability}
\normalfont \textbf{Actionness score.} Actionness score at a temporal position is the probability of occurrence of an action instance at that time. This score is often denoted by $a_t \in [0,1]$ for time $t$.   
% Actionness score of a proposal is the probability that it contains an action instance. High actionness scores suggest the existence of action instances while lower actionness scores indicate the presence of temporal background.
\end{definition}

\begin{definition}
\label{start-end-probability}
\normalfont \textbf{Startness and endness scores.} Startness score (endness score) at a temporal position is the probability of start (end) of an action instance at that time. 
\end{definition}

% The probability of capturing a complete action in a proposal. 
\begin{definition}
\label{action-completeness-score}
\normalfont \textbf{Action completeness score.} The maximum tIoU between a candidate proposal and ground truth action instances is called action completeness of that proposal. It was shown by \cite{zhao2017temporal} that incomplete proposals that have low tIoU with ground-truth intervals could have high classification scores. Therefore, action completeness must be considered to evaluate and rank the predicted proposals. 
\end{definition}

\begin{definition}
\label{action-classification-score}
\normalfont \textbf{Action classification score.} Generated temporal proposals are fed to action classifiers to produce a probability distribution over all action classes. This can be represented by vector $(p^{1},\cdots,p^{C})$ where $p^{i}$ is the probability of action class $i$, and $C$ is the number of classes. For a fair comparison, researchers utilize classifiers from earlier work SCNN-classifier \cite{shou2016temporal}, UntrimmedNet \cite{shou2016temporal}, \cite{xiong2016cuhk}, etc. They uniformly sample a constant number of frames from the video segment and feed it to ConvNets such as C3D \cite{tran2015learning}, two stream CNNs \cite{simonyan2014two} or temporal segment networks \cite{wang2016temporal}. In some cases, the recognition scores of sampled frames are aggregated with the Top-k pooling or weighted sum to yield the final prediction.

% Temporal action detection task has two main steps. First, generating temporal proposals (def \ref{Temporal Proposals}), i.e., temporal intervals of the video that are likely to include an action. Second, feeding the detected proposals to the state-of-the-art action classifiers for classification. For fair comparison, researchers often use the following classifiers for evaluation.  

% Shou \textit{et al.} in SCNN-classifier \cite{shou2016temporal} uniformly sample 16 frames from each video and feed it to 3D ConvNets (C3D \cite{tran2015learning}) for action classification. Wang \textit{et al.} in UntrimmedNet \cite{wang2017untrimmednets} sample a single frame (or 5 frame stacking of optical flow) every 30 frames and apply the two stream CNNs \cite{simonyan2014two} or temporal segment networks \cite{wang2016temporal} for classification. The recognition scores of sampled frames are aggregated with top-k pooling (k set to 20) or weighted sum to yield the final video-level prediction.% Xiong \textit{et al.} in \cite{xiong2016cuhk} top-k pooling and attention weighted pool-ing.  

\end{definition}

\subsection{Video Feature Encoding}
\label{Video-Feature-Encoding}

Untrimmed videos are often lengthy and can be as long as several minutes, and thus it is difficult to directly input the entire video to a visual encoder for feature extraction due to the limits of computational resources. For instance, popular video feature extractors such as 3D-CNNs can only operate on short clips spanning about 4 seconds. A common strategy for video representation is to partition the video into equally sized temporal intervals called \textit{snippets}, and then apply a pre-trained visual encoder over each snippet. Formally, given input video $X$ with $l$ frames, a sequence $S$ of snippets with regular duration $\sigma$ is generated where

% $S= \{s_n\}_{n=1}^{l_s}$ and $l_s = \frac{l}{\sigma}$. 

\begin{equation}
 S= \{s_n\}_{n=1}^{l_s} \hspace{0.1in} , \hspace{0.1in}  l_s = \frac{l}{\sigma},    
\end{equation}

and $s_n$ is the n-th snippet. Then, each snippet is fed to a pre-trained visual encoder such as two-stream \cite{simonyan2014two}, C3D \cite{tran2015learning}, or I3D \cite{carreira2017quo} for feature extraction. In two-stream network \cite{simonyan2014two}, snippet $s_n$ which is centered at $t_n$-th frame of the video, has an RGB frame $x_{t_n}$, and a stacked optical flow $o_{t_n}$ derived around the center frame. The RGB frame $x_{t_n}$ is fed to spatial network ResNet \cite{he2016deep}, extracting feature vector $f_{S, n}$. The optical flow $o_{t_n}$ is fed to temporal network BN-Inception \cite{ioffe2015batch}, extracting feature $f_{T, n}$. The spatial and temporal features, $f_{S,n}$ and $f_{T,n}$, are concatenated to represent the visual feature $f_n$ for snippet $s_n$. 

% \begin{equation}
%     s_n = (x_{t_n},o_{t_n})   \hspace{0.1in} , \hspace{0.1in}  f_n = (f_{S,n},f_{T,n})
% \end{equation}

Similarly, in I3D \cite{carreira2017quo}, a stack of RGB and optical flow frames from each snippet $s_n$ are fed to I3D network, extracting spatial and temporal feature vectors $f_{S, n}$ and $f_{T, n}$ which are concatenated to create feature $f_n$. In C3D \cite{tran2015learning}, the frames of each snippet $s_n$ are directly fed to a 3D-CNN architecture to capture spatio-temporal information, and extracting feature vector $f_n$.

\subsection{Action Detection with Full Supervision}
\label{Fully-Supervised Action Detection}

In fully-supervised action detection, the annotation (${\Psi}_g$ in Eq. \eqref{tad-formula}) of temporal boundaries and labels of action instances are provided for each video of training set. During inference, the goal is to find the temporal boundaries of action instances and predict their labels. A main step in action detection is \textit{temporal proposal generation} to identify the temporal intervals of the video that are likely to include action instances. Fully-supervised temporal proposal generation methods can be categorized to \textit{anchor-based} and \textit{anchor-free}. Anchor-based methods generate action proposals by assigning dense and multi-scale intervals with pre-defined lengths at each temporal position of the video (Section \ref{sec:anchor-based-methods}). Anchor-free methods often predict action boundary confidence or actionness scores at temporal positions of the video, and employ a bottom-up grouping strategy to match pairs of start and end (Section \ref{sec:anchor-free-methods}). There are also several methods that combine the advantages of anchor-free and anchor-based proposal generation methods (Section \ref{Anchor-based and Anchor-free Combination}). After generating the proposals, rich features must be extracted from the proposals to evaluate the quality of proposals. Section \ref{Temporal Proposal Evaluation} reviews common loss functions that are used during training for proposal evaluation. Section \ref{Dependencies-section} discusses modeling long-range dependencies to capture the relation between video segments in untrimmed videos to improve action localization. Finally, Section \ref{Spatio-temporal-Action-Detection} summarizes spatio-temporal action detection methods.

% \begin{figure}[!ht]
%     \centering
%     \includegraphics[scale=0.4]{figs/anchor-based-free.png}
%         \caption{Anchor-based vs anchor-free methods. Anchor-based methods assign multi-scale intervals with pre-defined lengths while anchor-free methods predict action boundary confidence at each temporal position.}
%     \label{fig:anchor-based-vs-free}
% \end{figure}

\subsubsection{\text{Anchor-based Proposal Generation and Evaluation}}
\label{sec:anchor-based-methods}

Anchor-based methods, also known as top-down methods, generate temporal proposals by assigning dense and multi-scale intervals with pre-defined lengths to uniformly distributed temporal locations in the input video. Formally, given a video with $T$ frames, $\frac{T}{\sigma}$ temporal positions, known as \textit{anchors}, are uniformly sampled from every $\sigma$ frames. Then, several temporal windows with different duration $\{d_1,d_2, \cdots, d_n\}$ are centered around each anchor as initial temporal proposals. The proposal lengths ($d_i$ s) must have a wide range to align with action instances of various lengths that can last from less than a second to several minutes in untrimmed videos \cite{idrees2017thumos}. Then visual encoders and convolution layers are applied on the temporal proposals for feature extraction and features are used to evaluate the quality of temporal proposals and adjust their boundaries (Section \ref{Temporal Proposal Evaluation}). 

\begin{figure}[!ht]
    \centering
    \includegraphics[scale=0.6]{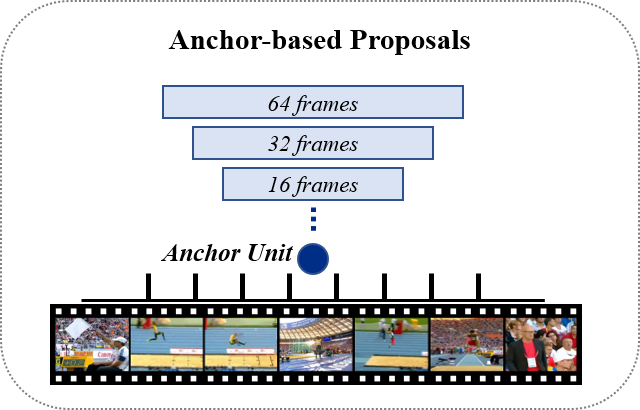}
        \caption{Anchor-based methods assign multi-scale intervals with pre-defined lengths at uniformly distributed temporal positions.}
    \label{fig:anchor-based}
\end{figure}

% and are supervised by loss functions described in section \ref{Temporal Proposal Evaluation}.

\paragraph{Feature Extraction of Multi-scale Proposals }

% \begin{table*}[]
% \caption{Summary of fully-supervised anchor-based techniques for proposal feature extraction.}
% \centering
% \resizebox{2\columnwidth}{!}{%
% \begin{tabular}{|l|l|l|}
% \hline
%  Category & Methods  & Contribution \\
% \hline
 
% Feature Concatenation & \cite{shou2016temporal,gao2017turn,gao2017cascaded} & Simple concatenation to extract fixed-size features. \\

% 3D RoI pooling & \cite{xu2017r,chen2020afnet,li2020graph} & RoI Pooling for fixed-size feature extraction. \\

% Multi-tower Network &  \cite{chao2018rethinking,gong2020scale} &  Receptive fields with different sizes for multi-scale proposals.\\

% TFPN & \cite{lin2017single,zhang2018s3d} & Aligning receptive field to proposal span in a unified pyramid.\\

% U-shaped TFPN &  \cite{li2019deep,liu2019multi,liuprogressive,gao2020accurate,wang2020multi} &  Connecting high-level and low-level features in the pyramid.\\

% \hline
% \end{tabular}
% }
% \label{tab:methods-summary}
% \end{table*}

As mentioned earlier, temporal proposals have very diverse time spans to align with action instances. However, fixed-size features must be extracted from each proposal to be fed to fully connected layers for proposal evaluation (action classification and regression). Here, we review different strategies to extract fixed-size features from proposals with different lengths.  

% \vspace{0.1in}
\textbf{Sampling and Feature Concatenation}: Shou \textit{et al.} in SCNN \cite{shou2016temporal} uniformly sampled a fixed number of frames from each proposal and fed them to a visual encoder for feature extraction. This is not computationally efficient because there are many overlapping proposals and overlapping segments are processed multiple times. To address this problem, Gao \textit{et al.} in Turn-Tap \cite{gao2017turn} and CBR \cite{gao2017cascaded} decomposed the video into non-overlapping equal-length units and extracted the features of each unit only once. Different numbers of consecutive units are grouped together at each anchor unit to generate multi-scale proposals. To obtain the proposal features, the features of all units are concatenated. Using this approach, the proposal features are computed from unit-level features, which are calculated only once. However, concatenation of features within each proposal or sampling frames do not lead to rich feature extraction. 

% \vspace{0.1in}
\textbf{3D RoI Pooling}: This approach extracts fixed size features from multi-scale proposals using 3D RoI pooling. Specifically, an input feature volume of size $l \times h \times w$ ( $l$ for temporal dimension, $h$ for height and $w$ for width dimensions) is divided into  $l_s \times h_s \times w_s$ sub-volumes (where $l_s, h_s,$ and $w_s$ are fixed), and max pooling is performed inside each sub-volume. Therefore, proposals of various lengths generate output volume features of the same size, which is $d \times l_s \times h_s \times w_s$, where $d$ is the channel dimension. The idea of 3D RoI pooling for action detection is an extension of the 2D RoI pooling for object detection in Faster R-CNN \cite{ren2015faster}. This idea was first introduced in R-C3D\cite{xu2017r} and used in other frameworks such as AGCN \cite{li2020graph} and AFNet \cite{chen2020afnet}. The limitation of this approach is that the multi-scale proposals at each location share the same receptive field, which may be too small or too large for some anchor scales.

% \begin{figure}[!h]
%     \centering
%     \includegraphics[scale=0.5]{figs/sliding-window-based.png}
%     \caption{Sliding window-based methods with anchor units and feature concatenation \cite{gao2017turn}, \cite{gao2017cascaded}.}
%     \label{fig:my_label}
% \end{figure}

% Formally, an input video $X$ with $l$ frames is represented as $X = \{t_i\}_{i=1}^{l}$. The video is divided into $l_s = \frac{l}{\sigma}$ units, where $\sigma$ is the number of frames in a unit. 
% A unit position $u_i$ ($1 \leq i \leq l_s$) is also known as \textit{anchor unit}, see figure \ref{fig:anchor-based-methods}. Consecutive units are grouped together at each anchor unit to generate proposals. A proposal $P_{j,n}   = \{u_i\}_{i=j}^{j+n}$ consisting of $n$ units, starts at the beginning of unit $u_{j}$ and ends right before unit $u_{j+n}$, including $\sigma n$ frames. To obtain the feature of proposal $P_{j,n}$, the features of all units $\{u_i\}_{i=j}^{j+n}$ are concatenated. 

\paragraph{\text{Receptive Field Alignment with Proposal Span}}

To address the variation of action duration, multi-scale anchors are assigned to each temporal location of the video. Before receptive field alignment, multi-scale anchors at any position share the same receptive field size. This is problematic because if the receptive field is too small or too large with respect to the anchor size, the extracted feature may not contain sufficient information or include too much irrelevant information. Here, we review the strategies to align the receptive field size with proposal span.

%\vspace{0.1in}
\textbf{Multi-tower Network}: TAL-Net \cite{chao2018rethinking} proposed a multi-tower network, compose of several temporal convNets, each one responsible for a certain anchor-size. Then, the receptive field of each anchor segment was aligned with its temporal span using dilated temporal convolutions. This idea was also used in TSA-Net \cite{gong2020scale}. However, assigning pre-defined temporal intervals limits the accuracy of generated proposals. 

%\vspace{0.1in}
\textbf{Temporal Feature Pyramid Network}: In a temporal feature pyramid network (TFPN), the predictions are yielded from multiple resolution feature maps. This idea was first introduced in Single Shot Detector (SSD) \cite{liu2016ssd} for object detection, and then extended to temporal domain for action detection in SSAD \cite{lin2017single} and $\text{S}^3\text{D}$ \cite{zhang2018s3d}. They proposed an end-to-end network where the lower-level feature maps with higher resolution and smaller receptive field are responsible to detect short action instances while the top layers with lower resolution and larger receptive field, detect long action instances. For each feature map cell, several anchor segments with multiple scales are considered around the center that are fed to convolutional layers for evaluation. The limitation of this approach is that lower layers in the pyramid are unaware of high-level semantic information, and top layers lack enough details, so they all fail to localize the actions accurately. 

\textbf{U-shaped Temporal Feature Pyramid Network}: In order to mitigate the problem with regular TFPNs, a U-shaped TFPN architecture was designed to connect high-level and low-level features. This idea was first introduced in Unet \cite{ronneberger2015u}, FPN \cite{lin2017feature}, and DSSD \cite{fu2017dssd} for object detection and then was generalized to temporal domain in MGG \cite{liu2019multi}, PBRNet \cite{liuprogressive}, RapNet \cite{gao2020accurate}, C-TCN \cite{li2019deep}, and MLTPN \cite{wang2020multi}. The video representation features are extracted using off-the-shelf feature extractors. Then temporal convolution and max pooling layers are applied to reduce the temporal dimension and increase the receptive field size. This is followed by temporal deconvolution layers for upscaling. Then, high-level features are combined with corresponding low-level features with lateral connections between the convolutional and deconvolutional layers. U-shaped TFPNs have drawn much attention recently and achieved state-of-the art results for temporal action detection task.

\subsubsection{\text{Anchor-free Proposal Generation and Evaluation}}
\label{sec:anchor-free-methods}

Anchor-free methods employ a bottom-up grouping strategy for proposal generation based on predicted boundary probability or actionness scores at temporal positions of the video. Anchor-free methods are capable to generate proposals with precise boundary and flexible duration because the proposal lengths are not predefined.  

%In this section, we review and analyze the state-of-the-art anchor-free methods.

% The advantage of this strategy is to generate proposals with more precise boundaries and flexible duration.

% predict a set of probability scores, such as actionness (definition \ref{actionness-probability}), at each temporal position of the video. Then, initial proposals are generated by matching or grouping temporal positions with higher scores (or other relevant conditions). 

% Anchor-free proposal generation methods, also known as bottom-up methods, are per-frame action labeling schemes, classifying every frame into the background or action classes. 

% The main target of anchor-free methods is to capture the local information to generate more precise boundaries and flexible duration.

% SSN \cite{zhao2017temporal} groups continuous high-score regions as the proposal by actionness scores. TSN \cite{wang2016temporal} is applied to extract actionness probabilities for each snippet of the video. Then a multi-scale temporal actionness grouping (TAG) method is proposed to generate action proposals using actionness probabilities. The idea is to find the continuous temporal regions with high actionness scores to serve as proposals. TAG method is based on a classic watershed algorithm \cite{roerdink2000watershed} and is applied to the 1D signal formed by actionness values. 

% \subsubsection{Common Proposal Generation Methods}

% \vspace{0.1in}
% $\bullet$ \textbf{Temporal Actionness Grouping}. 

\paragraph{\text{Proposal Generation with Actionness Scores}}

Zhao \textit{et al.} in SSN \cite{zhao2017temporal} proposed to identify continuous temporal regions with high actionness scores (def \ref{actionness-probability}) as proposals (known as TAG proposals). Continuous temporal regions are grouped using a classic watershed algorithm \cite{roerdink2000watershed} applied on the 1D signal formed by complemented actionness values. The proposals are fed to a temporal pyramid for feature extraction and proposal evaluation. The feature extraction process is too simple to capture rich features.

\paragraph{Proposal Generation with Boundary Scores}

These methods predict three probability signals for \textit{actionness} (def \ref{actionness-probability}), \textit{startness} and \textit{endness} scores (def \ref{start-end-probability}). They generate temporal proposals by matching the temporal positions that are likely to be the \textit{start} or \textit{end} of an action (peak of startness and endness signals). 
In BSN \cite{lin2018bsn} proposal features are constructed by concatenation of a fixed number of points, sampled from the actionness scores (def \ref{actionness-probability}) by linear interpolation. BSN ignores the global information for actions with blurred boundaries, causing unreliable confidence scores. Also, proposal features are too weak to capture enough temporal context, and the feature construction and confidence evaluation are performed for each proposal separately, which is inefficient. BMN \cite{lin2019bmn} explores the global context for simultaneously evaluating all proposals end-to-end. They construct a feature map by aggregating the features of all proposals together. The feature map is fed to convolution layers to simultaneously evaluate all proposals. The advantage of this approach is to extract rich feature and temporal context for each proposal and exploit the context of adjacent proposals. Also, proposal evaluation is very fast during inference. However, they use the same method as BSN \cite{lin2018bsn} to generate boundary probabilities (start and end) which ignores the global information for actions with blurred boundaries. DBG \cite{lin2020fast} simultaneously evaluates all proposals to explore global context and extract rich features similar to BMN \cite{lin2019bmn}. Moreover, instead of only exploiting the local information to predict boundary probabilities (probability of start and end), DBG proposed to employ global proposal-level features.

\begin{figure}[!ht]
    \centering
    \includegraphics[scale=0.6]{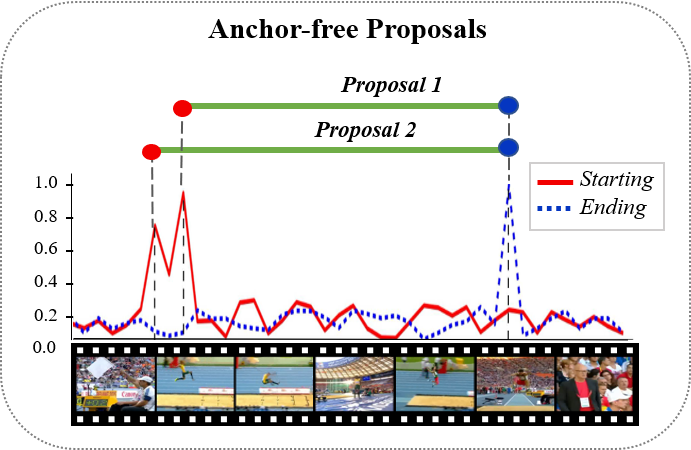}
        \caption{Anchor-free proposal generation with boundary matching. These methods predict action boundary probabilities at uniformly distributed temporal positions and match the start and end points with high probabilities as proposals.}
    \label{fig:anchor-based-vs-free}
\end{figure}

In order to model the relations between the boundary and action content of temporal proposals, BC-GNN \cite{bai2020boundary} proposed a graph neural network where boundaries and content of proposals are taken as the nodes and edges of the graph, and their features are updated through graph operations. Then the updated edges and nodes are used to predict boundary probabilities and content confidence score to generate proposals. A2Net \cite{yang2020revisiting} and AFSD \cite{lin2021learning} visit the anchor-free mechanism, where the network predicts the distance to the temporal boundaries for each temporal location in the feature sequence. AFSD \cite{lin2021learning} also proposes a novel boundary refinement strategy for precise temporal localization.

\subsubsection{Anchor-based and Anchor-free Combination}
\label{Anchor-based and Anchor-free Combination}

Anchor-based methods consider segments of various lengths as initial proposals at regularly distributed temporal positions of the video. However, because the segment sizes are designed beforehand, they cannot accurately predict the temporal boundary of actions. Also, because the duration of action instances varies from seconds to minutes, covering all ground-truth instances with anchor-based methods is computationally expensive. Anchor-free methods predict action boundary confidence or actionness score at all temporal positions of the video, and employ a bottom-up grouping strategy to match pairs of start and end. Anchor-free methods are capable to generate proposals with precise boundaries and flexible duration. However, in some cases they only exploit local context to extract the boundary information. Therefore, they are sensitive to noise, likely to produce incomplete proposals, and fail to yield robust detection results.

Several methods such as CTAP \cite{gao2018ctap}, MGG \cite{liu2019multi}, PBRNet \cite{liuprogressive}, RapNet \cite{gao2020accurate} balance the advantages and disadvantages between anchor-based and anchor-free approaches for proposal generation. CTAP \cite{gao2018ctap} designed a complementary filter applied on the initial proposals to generate the probabilities of proposal detection by anchor-free TAG \cite{zhao2017temporal} (defined in \ref{sec:anchor-free-methods}). The original use of complementary filtering is to estimate a signal given two noisy measurements, where one of them is mostly high-frequency (maybe precise but not stable) similar to TAG proposals and the other one is mostly low-frequency (stable but not precise) similar to sliding-window proposals. Also, several temporal feature pyramid networks (TFPN, defined in \ref{sec:anchor-based-methods}) such as MGG~\cite{liu2019multi}, PBRNet ~\cite{liuprogressive}, and RapNet ~\cite{gao2020accurate} generate coarse segment proposals of various length with TFPN (anchor-based), and simultaneously predict fine-level frame actionness (anchor-free). The advantage of this idea is to adjust the segment boundary of proposals with frame actionness information during the inference.

\subsubsection{\text{Common Loss Functions for Proposal Evaluation}}
\label{Temporal Proposal Evaluation}

After generating the temporal proposals, rich features are extracted from the proposals to evaluate their quality. Several convolutional layers are applied on the features to predict actionness score (def \ref{actionness-probability}), completeness score (def \ref{action-completeness-score}), classification score (def \ref{action-classification-score}), and to adjust the temporal boundary of the proposals. Here, we review common loss functions that are used during training to supervise these predicted scores and evaluate the quality of proposals.

\begin{definition}
\label{actionness-loss}
\normalfont \textbf{Actionness loss.} This is a binary cross-entropy loss that classifies the temporal proposals as action or background. Given $N$ proposals, this loss is defined as:

\begin{equation}
    L_{\text{act}} = - \frac{1}{N}
    \sum_{i=1}^{N} b_i \log (a_i) + (1-b_i) \log (1-a_i),
\end{equation}

where $a_i$ is the predicted actionness score (def \ref{actionness-probability}), and $b_i \in \{0,1\}$ is a binary ground-truth label for the $i$-th proposal. If the proposal is positive (def \ref{pos-neg-proposals}), then $b_i=1$. Otherwise, $b_i=0$. 
\end{definition}

\begin{definition}
\label{action-completeness}
\normalfont \textbf{Action completeness loss.} Given $N$ proposals, the completeness loss is defined as:

\begin{equation}
    L_{\text{com}} = \frac{1}{N_{\text{pos}}}
    \sum_{i=1}^{N} d(c_i , g_i) \cdot [l_i > 0],
\end{equation}

where $c_i$ is the predicted action completeness score (def \ref{action-completeness-score}) for $i$-th proposal, and $g_i$ is the ground-truth action completeness score. $d$ is a distance metric which is often $L_2$ or smooth $L_1$ loss. $l_i$ is the label of the $i$-th proposal and condition $[l_i >0]$ implies that action completeness is only considered for positive proposals (def \ref{pos-neg-proposals}). $N_{\text{pos}}$ is the number of positive proposals during each mini-batch. 
\end{definition}

\begin{definition}
\label{overlap-loss}
\normalfont \textbf{Action overlap loss.} This is another variation of action completeness loss which rewards the proposals with higher temporal overlap with ground truths and is defined as the following:

\begin{equation}
    \mathcal{L}_{overlap} = \frac{1}{N_{\text{pos}}} \sum_{i}  \frac{1}{2} \cdot \Big( \frac{(p^{l_i}_i)^2}{(g_i)^{\alpha}}-1 \Big) \cdot [l_i >0 ],
\end{equation}

where $p_i$ is the classification probability vector over action labels for the $i$-th proposal, and $p^{l_i}_i$ is the probability of action class $l_i$. Other notations are the same as in $L_{\text{com}}$ (def \ref{action-completeness}) and $\alpha$ is a hyper-parameter. 
\end{definition}

\begin{definition}
\label{action-classification}
\normalfont \textbf{Action classification loss.} This is the classification (cross-entropy) loss and the probability distribution is over all action classes as well as temporal background as the following:

\begin{equation}
    L_{\text{cls}} = - \frac{1}{N}
    \sum_{i=1}^{N} \log (p^{l_i}_i),
\end{equation}

where $l_i \in \{0,1,\cdots,C\}$ is the label of $i$-th proposal, and $p^{l_i}_i$ is the probability of class $l_i$.
\end{definition}

\begin{definition}
\label{regression-loss}
\normalfont \textbf{Action regression loss.} To adjust the temporal boundary of proposals, the start and end offset of proposals are predicted and supervised by a regression loss as the following:

\begin{equation}
    L_{\text{reg}} = 
    \frac{1}{N_{\text{pos}}} \sum_{i=1}^{N} |(o_{s,i} -o^{\star}_{s,i} ) + (o_{e,i} - o^{\star}_{e,i})| \cdot [l_i > 0],
\end{equation}

where term $o_{s,i}$ is the difference between the start coordinate of $i$-th proposal and the start coordinate of the closest ground truth action instance. The term $o^{\star}_{s,i}$ is the predicted offset. Similarly, $o_{e,i}$ and $o^{\star}_{e,i}$ are the ground-truth and predicted offset for end coordinate of the $i$-th proposal. The condition $[l_i >0]$ implies that boundary adjustment is only considered for positive proposals (def \ref{pos-neg-proposals}).

\end{definition}

\subsubsection{Modeling Long-range Dependencies}

\label{Dependencies-section}

As mentioned earlier, untrimmed videos are often lengthy and must be partitioned into shorter clips for feature extraction. Processing these shorter clips independently can lead to loss of temporal or semantic dependencies between video segments. Therefore, several tools such as recurrent neural networks, graph convolutions, attention mechanism and transformers are used to capture these dependencies. The advantage of modeling dependencies is to refine the temporal boundary of proposals or predict their action category or action completeness given the information from other neighboring proposals. 

\paragraph{Recurrent Neural Networks} 

RNNs are used for sequence modeling and are capable of capturing long-term dependencies in videos. Buch \textit{et al.} in Sst \cite{buch2017sst} and SS-TAD \cite{buch2019end} used RNNs for action detection. They partition the video into non-overlapping equal-length segments and feed each segment to a visual encoder for feature extraction. At time $t$, visual feature $f_t$ and the hidden state of the previous time step ($h_{t-1}$) are fed to a Gated Recurrent Unit (GRU)-based architecture to produce hidden state $h_t$. This hidden state is then fed to fully connected layers to evaluate multi-scale proposals at time $t$ by producing actionness scores (def \ref{actionness-probability}). In an earlier work, Yuan \textit{et al.} in PSDF \cite{yuan2016temporal} captured the motion information over multiple resolutions and utilized RNNs to improve inter-frame consistency. Yeung \textit{et al.} learn decision policies for an RNN-based agent \cite{yeung2016end}, and later proposed an LSTM model to process multiple input frames with temporal attention mechanism \cite{yeung2018every}. LSTMs are also used in other frameworks such as \cite{escorcia2016daps}, \cite{singh2016multi}, \cite{ma2016learning} to evaluate temporal proposals. The advantage of using RNNs is that hidden state at time $t$ encodes the information from previous time steps which is useful to capture temporal dependencies. However, RNNs are not capable to encode very long videos as the hidden vector gets saturated after some time steps. 

\begin{figure}[!h]
    \centering
    \includegraphics[scale=0.5]{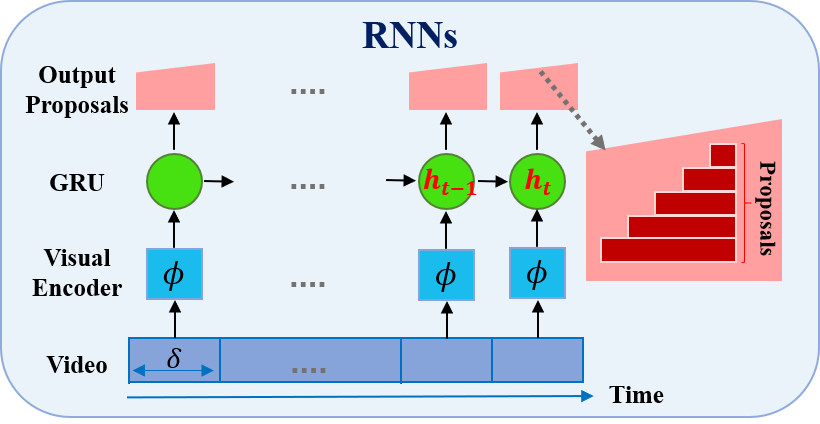}
    \caption{Capturing temporal dependencies in untrimmed videos with RNNs. Hidden state at time $t$, $h_t$, encodes the information from previous time steps. This picture is regenerated from \cite{buch2017sst}.}
    \label{fig:RNN}
\end{figure}

% \begin{figure*}[!t]
% \centering
% \includegraphics[scale=0.6]{figs/long-range-modeling.png}
% \caption{Modeling proposal-proposal relations with graph convolutional networks (GCNs), where the nodes are the proposals and the edges model the relations between proposals. The feature of proposal $p_3$ is influenced by the features of proposals $p_1,p_2$, and $p_4$. Image is reproduced from PGCN \cite{zeng2019graph}.}
% \label{fig:pgcn-pic}
% \end{figure*}

% Recurrent neural networks, graph convolutions and transformers are three main approaches to model the dependencies in untrimmed videos. 
% However, partitioning the video into shorter segments leads to loss of temporal and semantic dependencies. Here, we discuss strategies developed to capture the temporal dependencies. 

\paragraph{Graph Models} 

% \textbf{Proposal-proposal Relation}: 
A full action often consists of several sub-actions that may independently be detected in several overlapping proposals. Based on this observation, Zeng \textit{et al.} in PGCN \cite{zeng2019graph} captured proposal-proposal relations by applying graph-convolution networks (GCNs). They constructed a graph where the nodes are the proposals. The edges connect highly overlapped proposals as well as disjoint but nearby proposals to provide contextual information. The edge weights model the relation between the proposals by measuring cosine similarity of their features. Through graph convolutions feature of each proposal gets updated by aggregating the information from other proposals. The updated features are then used to predict action categories, completeness, and refining the boundaries.

\begin{figure}[!h]
\centering
\includegraphics[scale=0.5]{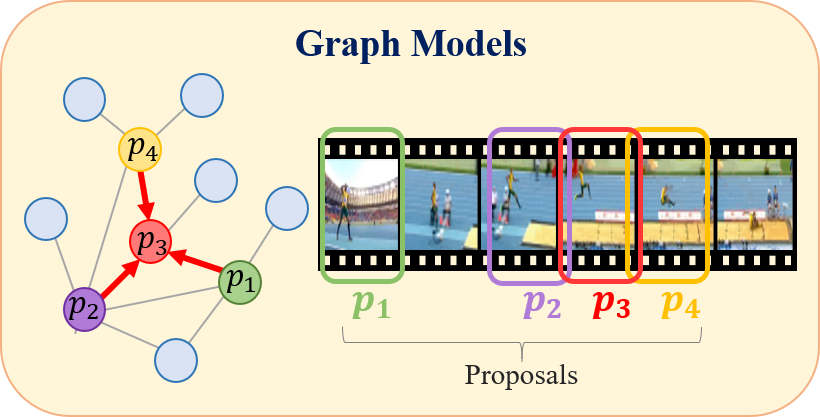}
\caption{Modeling proposal-proposal relations with graph convolutional networks (GCNs), where the nodes are the proposals and the edges model the relations between proposals. The feature of proposal $p_3$ is influenced by the features of proposals $p_1,p_2$, and $p_4$. Image is reproduced from PGCN \cite{zeng2019graph}.}
\label{fig:pgcn-pic}
\end{figure}

Li \textit{et al.} in AGCN \cite{li2020graph} proposed an attention based GCN to model the inter and intra dependencies of the proposals. Intra attention learns the long-range dependencies among pixels inside each action proposal and inter attention learns the adaptive dependencies among the proposals to adjust the imprecise boundary. Bai \textit{et al.} in BC-GNN \cite{bai2020boundary} proposed a graph neural network to model the relations between the boundary and action content of temporal proposals. 

Xu \textit{et al.} proposed G-TAD \cite{xu2020g} to capture the relations between different snippets of input video. They constructed a graph where the nodes are temporal segments of the video and the edges are either temporal or semantic. The temporal edges are pre-defined according to the snippets’ temporal order but the semantic edges are dynamically updated between the nodes according to their feature distance. Temporal and semantic context of the snippets are aggregated using graph convolutions. All possible pairs of start and end with duration within a specific range are considered to generate the proposals. To evaluate each proposal, the temporal and semantic features of the corresponding sub-graph are extracted. Chang \textit{et al.} in ATAG \cite{chang2021augmented} also designed an adaptive GCN similar to G-TAD \cite{xu2020g} to capture local temporal context where the graph nodes are the snippets and the edges model the relation between snippets. The temporal context is then captured through graph convolutions where the feature of each snippet is influenced and updated by the features of other snippets. VSGN \cite{zhao2020video} builds a graph on video snippets similar to G-TAD \cite{xu2020g}, but also exploits correlations between cross-scale snippets. They propose a cross-scale graph pyramid network which aggregates features from cross scales, and progressively enhances the features of original and magnified scales at multiple network levels.

% Graph convolutions aggregate the information among proposals and the features of each proposal (node) gets updated by the features of its neighboring nodes. The proposal features are extracted by I3D \cite{carreira2017quo} and fed to two GCNs to predict the action categories, refine the temporal boundaries, and find the action completeness.

% Therefore, aggregating the features of overlapping proposals is useful to predict the action completeness of each proposal and refine the temporal boundaries.  Recent methods, model the proposal-proposal relations or temporal segment relations with graphs and apply graph-convolution networks (GCNs) to aggregate the features of adjacent nodes. 

% \begin{figure}[!t]
% \centering
% \includegraphics[scale=0.5]{figs/pgcn-pic.png}
% \caption{Modeling proposal-proposal relations with GCNs, where the nodes are the proposals and the edges are between overlapped or nearby proposals. The feature of $p_3$ is influenced by the features of $p_1,p_2$, and $p_4$.
% Image is reproduced from PGCN \cite{zeng2019graph}.}
% \label{fig:pgcn-pic}
% \end{figure}

\paragraph{Transformers}

Some action instances in a video have \textit{non-sequential dependencies}, meaning that they are related but are separated by other events in the video. Also, some action instances may have overlaps in their temporal extents. Based on these observations, Nawhal \textit{et al.} in AGT \cite{nawhal2021activity} proposed an encoder decoder transformer to capture non-linear temporal structure by reasoning over videos as nonsequential entities. Their encoder generates a context graph where the nodes are initially video level features and the interactions among nodes are modeled as learnable edge weights. Also, positional information for each node is provided using learnable positional encodings. Their decoder learns the interactions between context graph (latent representation of the input video) and graph structured query embeddings (latent representations of the action queries). 

Tan \textit{et al.} in RTD-Net \cite{tan2021relaxed} proposed a relaxed transformer to directly generate action proposals without the need to human prior knowledge for careful design of anchor placement or boundary matching mechanisms. The transformer encoder models long-range temporal context and captures inter-proposal relationships from a global view to precisely localize action instances. They also argued that the snippet features in a video change at a very slow speed and direct employment of self-attention in transformers can lead to over-smoothing. Therefore, they customized the encoder with a boundary-attentive architecture to enhance the discrimination capability of action boundary. %Moreover, they relax the strict criteria of assigning a single proposal to each ground-truth action instance because of ambiguity of temporal boundaries and annotation noises. Relaxing this criteria is useful for convergence of transformer during training. 
Chang \textit{et al.} in ATAG \cite{chang2021augmented} designed an augmented transformer to mine long-range temporal context for noisy action instance localization. The snippet-level features generated by transformer are used to classify the snippets to action or background under supervision of a binary classification loss. Throughout this process the transformer learns to capture long-term dependencies at snippet level.

\subsubsection{Spatio-temporal Action Detection}
\label{Spatio-temporal-Action-Detection}

\begin{figure}[!ht]
\centering
\includegraphics[scale=0.35]{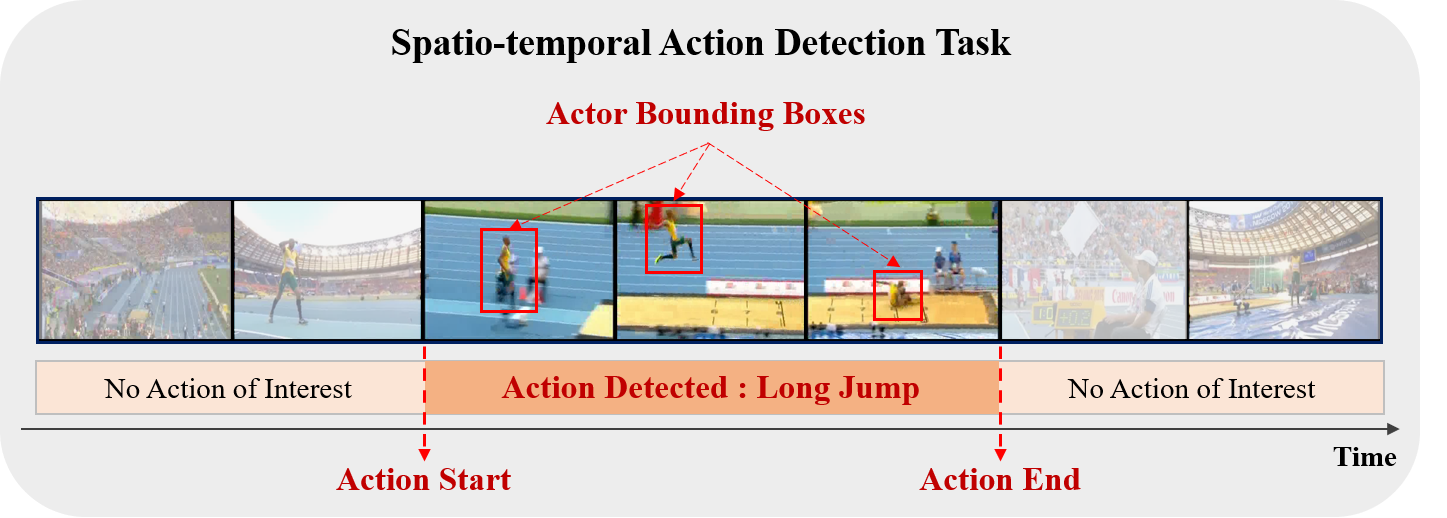}
\caption{Spatio-temporal activity detection task: action "long jump" is localized in time and space. Other than temporal interval of the action, bounding box of the person performing the action is detected in each frame. }
\label{fig:spatio-temporal}
\end{figure}

Spatio-temporal action detection aims to localize action instances in both space and time, and recognize the action labels. In the fully-supervised setting of this task, the temporal boundary of action instances at the video-level, the spatial bounding box of actions at the frame-level, and action labels are provided during training and must be detected during inference. Fig. \ref{fig:spatio-temporal} shows an example of this task. The start and end of action ``long jump'' are detected in temporal domain. Also, bounding box of the actor performing the action is detected in each frame in spatial domain.

% To evaluate spatio-temporal action detection methods, two metrics are frequently used. First, \textit{frame-AP} measures the area under the precision-recall curve of the detection for each frame. A detection is correct if the intersection-over-union with the ground truth at that frame is greater than a threshold and the action label is correctly predicted. Second, \textit{video-AP} measures the area under the precision-recall curve of the action tubes predictions. An action tube is a sequence of bounding boxes of action. A tube is correct if the mean per frame intersection-over-union with the ground truth across the frames of the video is greater than a threshold and the action label is correctly predicted.

% \paragraph{object proposals extended}
\paragraph{Frame-level Action Detection}

Early methods \cite{laptev2007retrieving,cao2010cross} were based on extensions of the sliding window scheme, requiring strong assumptions such as a cuboid shape, i.e., a fixed spatial extent of the actor across frames. Later, advancements in object detection inspired frame-level action detection methods to recognize human action classes at the frame level. In the first stage action proposals are produced by a region proposal algorithm or densely sampled anchors, and in the second stage the proposals are used for action classification and localization refinement. Hundreds of action proposals are extracted per video given low-level cues, such as super-voxels \cite{jain2014action,oneata2014spatio} or dense trajectories \cite{chen2015action,van2015apt,puscas2015unsupervised}, and then proposals are classified to localize actions. 

After detecting the action regions in the frames, some methods \cite{gkioxari2015finding, hou2017tube, peng2016multi, singh2017online, saha2016deep, weinzaepfel2015learning, yang2017spatio, ye2019discovering} use optical flow to capture motion cues. They employ linking algorithms to connect the frame-level bounding boxes into spatio-temporal action tubes. Gkioxari \textit{et al.} \cite{gkioxari2015finding} used dynamic programming approach to link the resulting per-frame detection. The cost function of the dynamic programming is based on detection scores of the boxes and overlap between detection of consecutive frames. Weinzaepfel et al. \cite{weinzaepfel2015learning} replaced the linking algorithm by a tracking-by-detection method. Then, two-stream Faster R-CNN was introduced by \cite{peng2016multi,saha2016deep}. Saha \textit{et al.} \cite{saha2016deep} fuse the scores of both streams based on overlap between the appearance and the motion. Peng \textit{et al.} \cite{peng2016multi} combine proposals extracted from the two streams and then classify and regress them with fused RGB and multi-frame optical flow features. They also use multiple regions inside each action proposal and then link the detection across a video based on spatial overlap and classification score.

Another group \cite{li2018videolstm,wang2016actionness,yu2015fast} rely on an actionness measure, i.e., a pixel-wise probability of containing any action. To estimate actionness, they use low-level cues such as optical flow \cite{yu2015fast}, CNNs with a two-stream architecture \cite{wang2016actionness} or RNNs \cite{wang2016actionness}. They extract action tubes by thresholding the actionness scores \cite{li2018videolstm} or by using a maximum set coverage \cite{yu2015fast}. The output is a rough localization of the action as it is based on noisy pixel-level maps.

The main disadvantage of these methods is that the temporal property of videos is not fully exploited as the detection is performed on each frame independently. Effective temporal modeling is crucial as a number of actions are only identifiable when temporal context information is available.

\paragraph{Clip-level Action Detection}

As mentioned earlier, temporal modeling is necessary for accurate action localization. Here, we discuss methods that exploit temporal information by performing action detection at the clip (i.e., a short video snippet) level.

Kalogeiton \textit{et al.} \cite{kalogeiton2017action} proposed action tubelet detector (ACT-detector) that takes as input a sequence of frames and outputs action categories and regressed tubelets, i.e., sequences of bounding boxes with associated scores. The tubelets are then linked to construct action tubes (sequence of bounding boxes of action). Gu et al. \cite{gu2018ava} further demonstrate the importance of temporal information by using longer clips and taking advantage of I3D pre-trained on the large-scale video dataset \cite{carreira2017quo}. In order to generate action proposals, they extend 2D region proposals to 3D by replicating them over time, assuming that the spatial extent is fixed within a clip. However, this assumption would be violated for the action tubes with large spatial displacement over time, in particular when the clip is long or involves rapid movement of actors or camera. Thus, using long cuboids directly as action proposals is not optimal, since they introduce extra noise for action classification.

Yang \textit{et al.} \cite{yang2019step} perform action detection at clip level, and then linked them to build action tubes across the video. They employ multi-step optimization process to progressively refine the initial proposals. Other methods \cite{feichtenhofer2019slowfast}, \cite{wu2019long} exploited human proposals coming from pretrained image detectors and replicated them in time to build straight spatio-temporal tubes.

\paragraph{Modeling Spatio-temporal Dependencies}

Understanding human actions usually requires understanding the people and objects around them. Therefore, state-of-the-art methods model the relation between actors and the contextual information such as other people and other objects. Some methods used the graph-structured networks \cite{sun2018actor, zhang2019structured} and attention mechanism \cite{girdhar2019video,wu2019long,ulutan2020actor} to aggregate the contextual information from other people and objects in the video.

Wu \textit{et al.} \cite{wu2019long} provided long-term supportive information that enables video models to better understand the present. The designed a long-term feature bank and a feature bank operator FBO that computes interactions between the short-term and long-term features. They integrate information over a long temporal support, lasting minutes or even the whole video. Girdhar et al. \cite{girdhar2019video} proposed a transformer-style architecture to weight actors with features from the context around him. Tomei \textit{et al.} \cite{tomei2019stage} employed self-attention to encode people and object relationships in a graph structure, and use the spatio-temporal distance between proposals. Ji \textit{et al.} proposed Action Genome \cite{ji2020action} to model action-object interaction, by decomposing actions into spatio-temporal scene graphs. Ulutan et al. \cite{ulutan2020actor} suggested combining actor features with every spatio-temporal region in the scene to produce attention maps between the actor and the context. Pan \textit{et al.} \cite{pan2020actor} proposed a relational reasoning module to capture the relation between the two actors based on their respective relations with the context. Tomei \textit{et al.} \cite{tomei2021video} proposed a graph-based framework to learn high-level interactions between people and objects, in both space and time. Spatio-temporal relationships are learned through self-attention on a multi-layer graph structure which can connect entities from consecutive clips, thus considering long-range spatial and temporal dependencies.

% \cite{hou2017tube}, \cite{li2018recurrent} use adaptive proposals for action detection. However, these methods require an offline linking process to generate the proposals. 

% \paragraph{weakly-supervised}

% Finally, weakly-supervised approaches have also been proposed (Escorcia et al., 2020).

% \clearpage

\subsection{\text{Action Detection with Limited Supervision}}
\label{Action Detection with Limited Supervision}

Fully supervised action detection requires the full annotation of temporal boundaries and action labels for all action instances in training videos, which is very time-consuming and costly. To eliminate the need for exhaustive annotations in the training phase, in recent years, researchers have explored the design of efficient models that require limited ground-truth annotations. We discuss weakly-supervised methods in Section \ref{limited-WS}, and other learning methods with limited supervision (unsupervised, semi-supervised, and self-supervised) are described in Section \ref{limited-others}. 

\subsubsection{Weakly-supervised Action Detection}
\label{limited-WS}

Weakly-supervised learning scheme requires coarse-grained or noisy labels during the training phase. Following the work of \cite{sun2015temporal}, weakly-supervised action detection in common settings requires only the video-level labels of actions during training while the temporal boundaries of action instances are not needed. During testing both labels and temporal boundaries of actions are predicted. In the following parts of this section, weakly-supervised action detection refers to this setting. There are also other weak signals utilized for action detection such as order of actions \cite{bojanowski2014weakly}, \cite{huang2016connectionist}, \cite{richard2017weakly}, \cite{kuehne2017weakly}, frequency of action labels \cite{narayan20193c}, and total number of events in each video \cite{schroeter2019weakly}. A common strategy in weakly-supervised action detection is to use attention mechanism to focus on discriminative snippets and combine salient snippet-level features into a video-level feature. The attention scores are used to localize the action regions and eliminate irrelevant background frames. There are two main strategies to extract attention signals from videos. First, \textit{class-specific attention} approaches where attention scores are generated from class activation sequences (def \ref{T-CAM}) for each action class (Section \ref{class-specific Attention}). Second, \textit{class-agnostic attention} approaches where attention scores are class-agnostic and are extracted from raw data (Section \ref{Class-agnostic Attention}). We discuss these two attention strategies in this section.

\paragraph{Term Definition}

To facilitate reading this section, we provide the definition of frequently used terminologies.

\begin{definition}
\label{T-CAM}
\normalfont \textbf{Temporal class activation maps (T-CAM).} For a given video, T-CAM is a matrix denoted by $A$ which represent the possibility of activities at each temporal position. Matrix $A$ has $n_c$ rows which is the total number of action classes, and $T$ columns which is the number of temporal positions in the video. Value of cell $A[c,t]$ is the activation of class $c$ at temporal position $t$. Formally $A$ is calculated by: 

\begin{equation}
    A = W X \oplus  b, 
\end{equation}

where $X \in {\rm I\!R}^{d \times T}$ is a video-level feature matrix, and $d$ is the feature dimension. Also, $W \in {\rm I\!R}^{n_c \times d}$ and $b \in {\rm I\!R}^{n_c}$, are learnable parameters and $\oplus$ is the addition with broadcasting operator.

\end{definition}

\begin{definition}
\label{class-specific attention score}
\normalfont \textbf{Class-specific attention scores.} In a given video, class-specific attention score is the occurrence probability of action class $c$ at temporal position $t$, denoted by $a[c,t]$. Formally, $a[c,t]$ is computed by normalizing the activation of class $c$ over temporal dimension:

\begin{equation}
a[c,t] = \frac{\text{exp}(A[c,t])}{\sum_{t=1}^{T}\text{exp}(A[c,t])},
\end{equation}

where $A$ is the T-CAM (def \ref{T-CAM}), and $T$ is the number of temporal positions. Therefore, row $a_c$ is the probability distribution of occurrence of class $c$ over video length. 
%A high attention value $a[c,t]$ indicates high probability of action class $c$ occurring at time $t$. 

\end{definition}

\begin{definition}
\label{Class-agnostic attention score}
\normalfont \textbf{Class-agnostic attention score.} In a given video, class-agnostic attention score, denoted by $\lambda_t$, is the occurrence probability of any action of interest at temporal position $t$, regardless of the action class. The attention vector for all temporal positions of the video is denoted by $\lambda$.
\end{definition}

\begin{definition}
\label{FG-BG}
\normalfont \textbf{Attention-based aggregated features.} The video-level foreground and background features are generated using temporal pooling of embedded features weighted by attention scores. Class-specific features are defined based on class-specific attention scores $a_c$ (def \ref{class-specific attention score}) for each class $c$ while class-agnostic features are defined based on class-agnostic attention vector $\lambda$ (def \ref{Class-agnostic attention score}). Aggregated foreground feature is most influenced by feature vectors with high attention that represent actions while background feature is impacted by features with low attention. $T$ is the video length and $X$ is the video feature matrix. These features are formulated as the following: 

\begin{table}[!h]
    \centering
    \resizebox{0.9\columnwidth}{!}{%
    \begin{tabular}{l  l  l }
          & Foreground: & Background: \\
          & & \\
         Class-specific: &  $f_c = X a_c  $  &  $b_c  = \frac{1}{T-1} X  ( \mathbb{1}-a_c), $\\
          & \\
         Class-agnostic: & $f = \frac{1}{T} X \lambda $ & $b  = \frac{1}{T} X ( \mathbb{1}-\lambda). $ \\
    \end{tabular}%
    }
\end{table}

\end{definition}

\paragraph{\text{Class-specific Attention for Action Localization}}
\label{class-specific Attention}

Class-specific attention module computes the attention weight $a[c,t]$ (def \ref{class-specific attention score}) for all action classes $c$ and all temporal positions $t$ in each video. The attention scores attend to the portions of the video where an activity of a certain category occurs. Therefore, video segments with attention scores higher than a threshold are localized as action parts. Class-specific attention module is used in \cite{wang2017untrimmednets}, \cite{paul2018w}, \cite{narayan20193c}, \cite{islam2020weakly} to localize the temporal boundary of action instances.

\vspace{0.1in}
\textbf{Class-specific attention learning with MIL}: In general scheme of MIL (multi-instance learning), training instances are arranged in sets, called bags, and a label is provided for the entire bag \cite{carbonneau2018multiple}. In the context of weakly-supervised temporal action detection, each video is treated as a bag of action instances and the video-level action labels are provided. In order to compute the loss for each bag (video in this task), each video should be represented using a single confidence score per category. The confidence score for each category is computed as the average of top $k$ activation scores over the temporal dimension for that category. In a given video, suppose set $\{t^c_1,t^c_2,\cdots,t^c_k\}$ are $k$ temporal positions with highest activation scores for class $c$. Then, the video-level class-wise confidence score $s^c$ for class $c$ is defined as:

\begin{equation}
    s^c = \frac{1}{k} \sum_{l=1}^{k} A[c,t^c_l],
\end{equation}

where $A[c,t^c_l]$ is the activation (def \ref{T-CAM}) of class $c$ at temporal position $t^c_l$. Then, probability mass function (PMF) of action classes is computed by applying softmax function on $s^c$ scores over class dimension:

\begin{equation}
  p^c = \frac{\exp{(s^c)}}{\sum_{c=1}^{n_c} \exp{(s^c)}},  
\end{equation}

where $n_c$ is the number of action classes. MIL loss is a cross-entropy loss applied over all videos and all action classes. For video $i$ and action class $c$, $p^c_i$ is the class-wise probability score, and $y^c_i$ is a normalized ground-truth binary label. MIL is defined as:

\begin{equation}
   L_{MIL} = \frac{1}{n} \sum_{i=1}^{n} \sum_{c=1}^{n_c} -y^c_i \log(p^c_i),
\end{equation}

where $n$ is the total number of videos. MIL loss supervises class-wise probability scores which are computed based on activation scores $A[c,t]$. Therefore, MIL learns activation scores and T-CAM (def \ref{T-CAM}) for each video and is used in W-TALC \cite{paul2018w}, Action Graphs \cite{rashid2020action}, UNet \cite{wang2017untrimmednets}, and Actionbytes \cite{jain2020actionbytes}.

\vspace{0.1in}
\textbf{Class-specific attention learning with CASL}: The CASL (co-activity similarity loss) was initially introduced in W-TALC\cite{paul2018w} and then inspired others Deep Metric \cite{islam2020weakly}, Action Graphs \cite{rashid2020action}, WOAD \cite{gao2020woad}, Actionbytes \cite{jain2020actionbytes}. The main idea is that for a pair of videos including the same action classes, the foreground features in both videos should be more similar than the foreground feature in one video and the background feature in the other video. For a pair of videos with indices $m$ and $n$ that include action class $c$, the foreground features are denoted by $f^m_c$, $f^n_c$ and the background features by $b^m_c$, $b^n_c$ (def \ref{FG-BG}). Then CASL is defined based on ranking hinge loss as the following:

\begin{equation}
\begin{split}
L^{mn}_c  & =  \frac{1}{2} \{ \max\big(0,d(f^m_c , f^n_c)-d(f^m_c , b^n_c)+\delta\big) \\ 
& + \max\big(0,d(f^m_c , f^n_c)-d(b^m_c , f^n_c)+\delta\big) \},
\end{split}
\label{co-activity-similarity-loss-equation}
\end{equation}

where $d$ is a metric (e.g., cosine similarity) to measure the degree of similarity between two feature vectors and $\delta$ is a margin parameter. The average of $L^{mn}_c$ is computed over all video pairs that include action class $c$. This loss trains class-specific attention scores $a_c$ as foreground and background features $f_c$ and $b_c$ are defined based on $a_c$ (def \ref{FG-BG}).

Islam \textit{et al.} in Deep Metric \cite{islam2020weakly} replaced metric $d$ with a class-specific metric $D_c$ defined for each class $c$. Rashid \textit{et al.} in Action Graphs \cite{rashid2020action} applied a GCN to transform each temporal segment’s feature representation to a weighted average of its neighbors. Then updated features are used in CASL for localization. The advantage of this GCN is to model temporal dependencies, cluster the semantically-similar time segments, and pushing dissimilar segments apart.

\vspace{0.1in}
\textbf{Class-specific attention learning with center loss}: The center loss which was first introduced in \cite{wen2016discriminative}, learns the class-specific centers and penalizes the distance between the features and their class centers. Narayan \textit{et al.} in 3C-Net \cite{narayan20193c} employed center loss to enhance the feature discriminability and reduce the intra-class variations. For each video $i$ and each action class $c$, center loss computes the distance (L2 norm) between class-specific foreground feature $f^i_c$ (def \ref{FG-BG}) and cluster center feature $z_c$ as the following:

\begin{equation}
\mathcal{L}_{center} = \frac{1}{N} \sum_{i} \sum_{c : y^i(c) =1} \norm{f^i_c - z_c}^2_2, 
\end{equation}

where cluster center feature $z_c$ is updated during training. Here, $N$ is the total number of videos, and condition $y^i(c) =1$ checks if action class $c$ occurs in video $i$.

\paragraph{\text{Class-agnostic Attention for Action Localization}}
\label{Class-agnostic Attention}

Class-agnostic attention module computes attention vector $\lambda$ (def \ref{Class-agnostic attention score}) directly from raw data, by applying fully connected and ReLU layers over video features, followed by a sigmoid function to scale attention weights to $[0,1]$. Learning class-agnostic attention weights is used in many methods such as RPN \cite{huang2020relational}, BG modeling \cite{nguyen2019weakly}, AutoLoc\cite{shou2018autoloc}, CleanNet \cite{liu2019weakly}, DGAM \cite{shi2020weakly},  STPN \cite{nguyen2018weakly} , BaSNet \cite{lee2020background} ,  MAAN \cite{yuan2019marginalized}, and CMCS \cite{liu2019completeness}.

\vspace{0.1in}
\textbf{Class-agnostic attention learning with cross-entropy}: The video-level class-agnostic foreground and background features $f$ and $b$ (def \ref{FG-BG}) are fed to a classification module, and supervised with a cross entropy loss:

\begin{equation}
    p_{fg}[c] = \frac{\exp{(w_c \cdot f)}}{\sum_{i=0}^{C} \exp{(w_i \cdot f)}}, \mathcal{L}_{fg} = -\log(p_{fg}[y]),  
\end{equation}

where $w_c$ s are the weights of the classification module, $C$ is the number of action classes in the entire dataset, and $y$ is the label of action that happens in the video. Also, label $0$ represents the background class. Similarly, $\mathcal{L}_{bg}$ is defined for $p_{bg}$ which is a softmax applied over multiplication of background feature $b$ and the classification module. This loss trains attention vector $\lambda$ through class-agnostic features $f$ and $b$ (def \ref{FG-BG}), and is used in STPN \cite{nguyen2018weakly}.

\vspace{0.1in}
\textbf{Class-agnostic attention learning with clustering loss}: Nguyen \textit{et al} in background modeling \cite{nguyen2019weakly} propose a method to separate foreground and background using a clustering loss by penalizing the discriminative capacity of background features. Class-agnostic foreground and background features $f$ and $b$ (def \ref{FG-BG}) are encouraged to be distinct using a clustering loss:

\begin{equation}
z_f = \frac{\exp(u f)}{\exp(u f)+\exp(v f)} \ , \ z_b = \frac{\exp(v b)}{\exp(u b)+\exp(v b)},
\end{equation}

\begin{equation}
\mathcal{L}_{cluster} = - \log{z_f}- \log{z_b},    
\end{equation}

where $u , v \in {\rm I\!R}^{d}$ are trainable parameters. Attention $\lambda$ is trained by separating class-agnostic features $f$ and $b$ (def \ref{FG-BG}).

\vspace{0.1in}
\textbf{Class-agnostic attention learning with prototypes:} Prototypical network which was introduced in \cite{snell2017prototypical} for classification task, represents each class as a prototype and matches each instance with a prototype with highest similarity. During training, the semantically-related prototypes are pushed closer than unrelated prototypes. Huang \textit{et al.} in RPN \cite{huang2020relational} proposed a prototype learning scheme for action localization. For temporal position $t$ and action class $c$, the similarity score $s_{t,c}$ between feature $x_t$ and prototype $p_c$ is computed and similarity vector $s_t$ consists of $s_{t,c}$ for all classes. Then the similarity vector $s_t$ is fused with attention score $\lambda_t$ into a video-level score $\hat{s}$:

\begin{equation}
s_{t,c} = - \norm{x_t - p_c}^2_2  \  \ ,  \  \ \hat{s} = \sum_{t=1}^{T} \lambda_t s_t. 
\end{equation}

Score $\hat{s}$ is supervised by a classification loss with respect to the video-level labels, training attention scores $\lambda_t$. 

\vspace{0.1in}
\textbf{Class-agnostic attention learning with CVAE}: DGAM \cite{shi2020weakly} aims to separate actions from context frames by imposing different attentions on different features using a generative model, conditional VAE (CVAE) \cite{sohn2015learning}. Formally, the objective of DGAM is:

\begin{equation}
\max_{\lambda \in [0,1]}   \underbrace{\log p(y|X,\lambda)}_{\text{term 1}} + \underbrace{\log p(X|\lambda)}_{\text{term 2}},
\end{equation}

where $X$ denotes the features, $y$ is the video-level label, and $\lambda$ is the attention signal. Term 1 encourages high discriminative capability of the foreground feature $f$ and punishes any discriminative capability of the background feature $b$. Term 2 is approximated by a generative model which forces the feature representation $X$ to be accurately reconstructed from the attention $\lambda$ using CVAE. By maximizing this conditional probability with respect to the attention, the frame-wise attention is optimized by imposing different attentions on different features, leading to separation of action and context frames. 

% \vspace{0.1in}
% \textbf{Background Suppression with Attention}: Lee \textit{et al.} \cite{lee2020background} proposed a two-branch weight-sharing architecture with an asymmetrical training strategy, named as base branch and suppression branch. Both branches extract the class-specific attentions (def \ref{class-specific attention score}) and are supervised with MIL loss (def \ref{MIL loss}) but background class is considered as positive in base branch and negative in suppression branch. To resolve their contrasting objectives, the suppression branch starts with a filtering module (following the class-agnostic attention module) which learns to suppress the activation scores from background and becomes free from the interference of background frames.

\paragraph{Direct Action Proposal Generation}
Many methods \cite{nguyen2018weakly}, \cite{paul2018w}, \cite{singh2017hide}, \cite{wang2017untrimmednets} localize the actions by applying thresholds on attention scores. The disadvantage of thresholding is that the snippets are treated independently and their temporal relations are neglected. Also, thresholding may not be robust to noises in class activation maps. Shou \textit{et al.} \cite{shou2018autoloc} in AutoLoc directly predict the temporal boundary of each action instance. A localization branch is designed to directly predict the action boundaries (inner boundaries). The outer boundaries are also obtained by inflating the inner boundaries. Knowing that a video includes action class $c$, an outer-inner-contrastive (OIC) loss is applied on the activation scores of action $c$. The OIC loss computes the average activation in the outer area minus the average activation in the inner area to encourage high activations inside and penalize high activations outside because a complete action clip should look different from its neighbours. Liu \textit{et al.} \cite{liu2019weakly} proposed CleanNet to exploit temporal contrast for action localization. A contrast score is generated by summing up action, starting and ending scores for each action proposal. The action localization is trained by maximizing the average contrast score of the proposals, which penalizes fragmented short proposals and promotes completeness and continuity in action proposals.

\paragraph{Action Completeness Modeling}

Previous methods used random hiding and iterative removal to enforce action completeness. Singh \textit{et al.} in Hide-and-seek \cite{singh2017hide} force the model to see different parts of the video by randomly masking different regions of the videos in each training epoch. However, randomly hiding frames does not always guarantee the discovery of new parts and also disrupts the training process. Zhong \textit{et al.} in Step-by-step erasion \cite{zhong2018step} trained a series of classifiers iteratively to find complementary parts, by erasing the predictions of predecessor classifiers from input videos. The major draw-back with this approach is the extra time cost and computational expense to train multiple classifiers. Zeng \textit{et al.} \cite{zeng2019breaking} propose an iterative-winners-out strategy that selects the most discriminative action instances in each training iteration and hide them in the next iteration. Liu \textit{et al.} in CMCS \cite{liu2019completeness} proposed to enforce multiple branches in parallel to discover complementary pieces of an action. Each branch generates a different class activation map (def \ref{T-CAM}). A diversity loss (introduced in \cite{lin2017structured}) is imposed on class activation maps, which computes cosine similarities between pairs of branches and all action categories. Minimizing the diversity loss, encourages the branches to produce activations on different action parts.

\subsubsection{\text{Unsupervised, Semi-supervised, and Self-supervised}}
\label{limited-others}

Although weakly-supervised action detection has been extensively studied in recent years, there are fewer articles addressing action detection task in unsupervised, semi-supervised, or self-supervised setting that are briefly reviewed here.

\paragraph{Unsupervised Action Detection} 

Unsupervised learning does not need any human-annotated labels during training. Seneret \textit{et al.} \cite{sener2018unsupervised} introduced an iterative approach which alternates between discriminative learning of the appearance of sub-activities from visual features and generative modeling of the temporal structure of sub-activities. Kukleva \textit{et al.} \cite{kukleva2019unsupervised} proposed a combination of temporal encoding (generated using a frame time stamp prediction network) and a Viterbi decoding for consistent frame-to-cluster assignment. Gong \textit{et al.} in ACL \cite{gong2020learning} used only the total count of unique actions that appear in the video set as supervisory signal. They propose a two-step clustering and localization iterative procedure. The clustering step provides noisy pseudo-labels for the localization step, and the localization step provides temporal co-attention models to improve the clustering performance.

\paragraph{Self-supervised Action Detection} 

Self-supervised learning refers to training with pseudo labels where pseudo labels are automatically generated for a pre-defined pretext task without involving any human annotations. Chen \textit{et al.} in SSTDA \cite{chen2020action} proposed self-supervised temporal domain adaptation method to address the spatio-temporal variations (different people performing the tasks in different styles) in action segmentation. They designed two self-supervised auxiliary tasks, binary and sequential domain prediction, to jointly align local and global embedded feature spaces across domains. The binary domain prediction task predicts a single domain for each frame-level feature, and the sequential domain prediction task predicts the permutation of domains for an untrimmed video, both trained by adversarial training with a gradient reversal layer (GRL) \cite{ganin2015unsupervised,ganin2016domain}. Jain \textit{et al.} in Actionbytes \cite{jain2020actionbytes} only use short trimmed videos during the training and train an action localization network with cluster assignments as pseudo-labels to segments a long untrimmed videos into interpretable fragments (called ActionBytes). They adopt a self-supervised iterative approach for training boundary-aware models from short videos by decomposing a trimmed video into ActionBytes and generate pseudo-labels to train a CNN to localize ActionBytes within videos.

\begin{table*}[t]
\centering
\caption{The benchmark datasets for temporal and spatio-temporal action detection.}
\begin{tabular}{l | c c c c c c  }
\hline
Dataset	 &	Activities Types	&    \mypound Videos	&	 \makecell{ \mypound Action \\ Categories}	&  \makecell{Avg Video \\  Length (Sec)}	&	\makecell{ \mypound Action Instances \\ (avg per video)} & \makecell{Multi-label \\ (\mypound labels per frame)} \\
% \makecell{Object \\ Annotation} \\
\hline
THUMOS \cite{jiang2014thumos}	 &	Sports	&	413	&	20	&	212	& 15.5 \cmmnt{(1.1 classes)} & No \cmmnt{(0.3)} \\	 %6,363	& \\
\hline
MultiTHUMOS	\cite{yeung2018every}   &	Sports	&	413	&	65	&	212	&	97 \cmmnt{(10.5 classes)} \cmmnt{38,690} & Yes \cmmnt{(1.5)} \\
\hline
ActivityNet \cite{caba2015activitynet}  &	Human Activities 	&	19,994	&	200	&	115	&	1.54 \cmmnt{30,791}	& No \\
\hline
HACS Segment \cite{zhao2019hacs}	 &	Human Activities 	&	50K	&	200	&	156	&	2.8 \cmmnt{139k}	& No \\
\hline
Charades \cite{sigurdsson2016hollywood}	 &	Daily Activities	&	9,848	&	157	&	30	&	6.75 \cmmnt{66,500}	& Yes \\ 
\cmmnt{& 41,104 labels, 46  classes \\}
\hline
Breakfast  \cite{kuehne2014language}   	&	Cooking 	& 1712	\cmmnt{1,989}	&	48	&	162	& 6 \cmmnt{8,456}	& No \\
\hline
50Salads \cite{stein2013combining}    & Cooking 	&	50	&	17	&	384	&	20	& No\\
\hline
MPII  cooking 2	\cite{rohrbach2016recognizing}   &	Cooking 	&	273	&	59	&	356	&   51.6 \cmmnt{14,105}	& No\\
\hline
COIN \cite{tang2019coin}	  &	Daily Activities	&	11,827	&	180	&	142	&	3.9 \cmmnt{46,354}	& No \\
\hline
Ava	\cite{gu2018ava}  &	Movies	&	437	&	80	&	900	&	3361.5 \cmmnt{1590000/473} & Yes \\
\hline
\end{tabular}
\label{tab:datasets}
\end{table*}

%Uses a small amount of labeled data in combination with a large amount of unlabeled data. 
\paragraph{Semi-supervised Action Detection}

In Semi-supervised setting, a small number of videos are fully annotated with the temporal boundary of actions and class labels while a large number of videos are either unlabeled or include only video-level labels. Ji \textit{et al.} \cite{ji2019learning} employ a fully supervised framework, known as BSN \cite{lin2018bsn}, to exploit the small set of labeled data. They encode the input video into a feature sequence and apply sequential perturbations (time warping and time masking \cite{tarvainen2017mean}) on it. Then, the student proposal model takes this perturbed sequence as the input but the teacher model predicts directly on the original feature sequence. In the end, the student model is jointly optimized with a supervised loss applied to labeled videos and a consistency loss to all videos.

\section{Datasets and Evaluation}
\label{Datasets and Evaluation}

In this section, we describe the datasets collected for action detection and the evaluation metrics for this task.

\subsection{Datasets}

%at each time step there can be more than one action label.

Gaidon \textit{et al.} \cite{gaidon2011actom,gaidon2013temporal} introduced the problem of temporally localizing the actions in untrimmed videos, focusing on limited actions such as ``drinking and smoking'' \cite{laptev2007retrieving} and ``open door and sitdown'' \cite{duchenne2009automatic}. Later, researchers worked on building the following datasets that include large number of untrimmed videos with multiple action categories and complex background information. Some of these datasets target activities of high-level semantics (such as sports) while others include fine-grained activities (such as cooking). The details are summarized in Table \ref{tab:datasets}.

% \textbf{Temporal Action Detection}: Targets activities of high-level semantics, and aims to produce a sparse set of action intervals, where an interval is defined by a start time, end time, and a class label. Several large-scale datasets such as THUMOS \cite{gorban2015thumos,jiang2014thumos} and Activitynet \cite{caba2015activitynet} are designed for this task with long untrimmed videos. The action instances are sparsely distributed in these datasets and the average number of action instances per video is $1.5$ in Activitynet. Also, a large portion of each video is labeled as background ($70\%$ of the frames in THUMOS dataset). 

% Fine-grained datasets such as Breakfast \cite{kuehne2014language}, 50Salads \cite{stein2013combining}, GTEA \cite{fathi2011learning} are built for action segmentation task by densely annotating all frames. The temporal background for this task can be as low as  $7\%$ (e.g., in Breakfast dataset \cite{kuehne2014language}). 

\vspace{0.1in}
$\bullet$ THUMOS14 \cite{jiang2014thumos} is the most widely used dataset for temporal action localization. There are $220$ and $213$ videos for training and testing with temporal annotations in $20$ classes. Action instances are rather sparsely distributed through the videos and about $70\%$ of all frames are labeled as background. The number of action instances per video on average is $15.5$ (and $1.1$ for distinct action instances). Also, maximum number of distinct actions per video is 3.

\vspace{0.1in}
$\bullet$ MultiTHUMOS \cite{yeung2018every} has the same set of videos as in THUMOS14 \cite{jiang2014thumos}, but it extends the latter from $20$ action classes with $0.3$ labels per frame to $65$ classes with $1.5$ labels per frame. Also, the average number of distinct action classes in a video is $10.5$ (compared to $1.1$ in THUMOS14), making it a more challenging multi-label dataset. Also, maximum number of distinct actions per video is 25. 
% \cite{piergiovanni2019temporal}, Actionbytes have results

\vspace{0.1in}
$\bullet$ ActivityNet \cite{caba2015activitynet} has two versions, v1.2  and v1.3. The former contains $9,682$ videos in $100$ classes, while the latter, which is a superset of v1.2 and was used in the ActivityNet Challenge 2016, contains $19,994$ videos in $200$ classes. In each version, the dataset is divided into three disjoint subsets, training, validation, and testing, by 2:1:1.  
%\cite{ghanem2017activitynet}.

\vspace{0.1in}
$\bullet$ HACS \cite{zhao2019hacs} includes $504K$ untrimmed videos retrieved from YouTube where each one is strictly shorter than $4$ minutes. HACS clips consists of $1.5M$ annotated clips of 2-second duration and HACS Segments contains $139K$ action segments densely annotated in $50K$ untrimmed videos spanning $200$ action categories.

\vspace{0.1in}
$\bullet$ CHARADES \cite{sigurdsson2016hollywood} consists of $9,848$ videos recorded by Amazon Mechanical Turk users based on provided scripts. This dataset contains videos with multiple actions and involves daily life activities from $157$ classes of $267$ people from three continents. Over $15\%$ of the videos have more than one person. 
% (typical examples include pairs of activities like ‘holding a phone’ and ‘playing with a phone’ or ‘holding a towel’ and ‘tidying up a towel’) 
% Standard evaluation process from \cite{sigurdsson2017asynchronous}. R-C3D\cite{xu2017r}, \cite{ghosh2020stacked} have results on charades\\

% \vspace{0.1in}
% $\bullet$ MEXaction2 dataset \cite{stoian2015fast,stoian2015scalable} has two action categories: "HorseRiding" and "BullChargeCape". This dataset is consisted of three subsets: YouTube clips, UCF101 Horse Rid-ing clips and INA videos. YouTube and UCF101 Horse Riding clips are trimmed and used for training set, whereas INA videos are untrimmed with approximately 77 hours in total and are divided into training, validation and testing set. Regarding to temporal annotated action instances, there are 1336 instances in training set, 310 instances in validation set and 329 instances in testing set.
% % used in SSAD , SCNN \\

\vspace{0.1in} 
$\bullet$  Breakfast \cite{kuehne2014language} includes $1712$ videos for breakfast preparation activities performed by $52$ subjects. The videos were recorded in $18$ different kitchens and belong to $10$ different types of breakfast activities (such as fried egg or coffee) which consist of $48$ different fine-grained actions. Each video contains $6$ action instances on average and only $7\%$ of the frames are background. 

% average frame accuracy (MoF)metric over the predefined train/test splits
% used in \cite{chen2020actionmixed},\cite{farha2019ms}, \cite{fayyaz2020sct}

\vspace{0.1in}
$\bullet$ 50Salads \cite{stein2013combining} contains $50$ videos for salad preparation activities performed by $25$ subjects and with $17$ distinct action classes. On average, each video contains $20$ action instances and is $6.4$ minutes long. 
% used in \cite{chen2020actionmixed},\cite{farha2019ms}

\vspace{0.1in}
$\bullet$ MPII Cooking 2 \cite{rohrbach2016recognizing} consists of $273$ videos with about $2.8$ million frames. There are $59$ action classes and about $29\%$ of the frames are background. The dataset provides a fixed split into a train and test set, separating $220$ videos for training.
% used in \cite{fayyaz2020sct}. 

\vspace{0.1in}
$\bullet$ COIN dataset \cite{tang2019coin},\cite{tang2020comprehensive} contains $180$ tasks and $11,827$ videos and $46,354$ annotated segments. The videos are collected from YouTube in $12$ domains (e.g., vehicles, gadgets, etc.) related to daily activities.
% used in ActBert, \\

% \vspace{0.1in}
% $\bullet$ Hollywood Extended \cite{bojanowski2014weakly} is collected from 69 Hollywood movies. It comprises a total of 937 video clips which have been annotated with 16 different action classes. On average, there are 5.9 action instances per video. With 61\%, the background ratio is similarly high as in THUMOS14. The overall number of frames is 780,000. 
% % % We report the Jaccard index (intersection over union) metric over the predeﬁned train/test splits following [28].
% % used in \cite{fayyaz2020sct}. 

% \vspace{0.1in}
% $\bullet$ GTEA (Georgia Tech Egocentric Activities) \cite{fathi2011learning} has 28 videos of 7 kitchen activities such as preparing coffee or cheese sandwich, performed by 4 subjects. All the videos were recorded by a camera that is mounted on the actors head. The frames of the videos are annotated with 11 distinct action classes including background, and on average each video has 20 action instances (not necessarily distinct). 
% % used in \cite{lei2018temporal},\cite{farha2019ms}\\
% % There are totally 11 action classes including back-ground. On average, each video is around one minute long with 20 action instances. 
% %  Each video contains about 1800 RGB frames, showing a sequence of 20 actions , including the background action. 

\vspace{0.1in}
$\bullet$ AVA \cite{gu2018ava} is designed for spatio-temporal action detection and consists of $437$ videos where each video is a $15$ minute segment taken from a movie. Each person appearing in a test video must be detected in each frame and the multi-label actions of the detected person must be predicted correctly. The action label space contains $80$ atomic action classes but often the results are reported on the most frequent $60$ classes.

% \vspace{0.1in}
% $\bullet$ Action Genome \cite{ji2020action} is designed for spatio-temporal action detection and built upon Charades dataset \cite{sigurdsson2016hollywood} and includes daily life activities from $157$ classes.
% Action Genome provides a decomposition of actions into spatio-temporal scene graphs. It contains $10K$ videos with $0.4M$ objects and $1.7M$ visual relationships annotated. There are $476K$ object bounding boxes and $234K$ video frames. The average length of videos is $30$ secs.

\subsection{Evaluation Metrics}

Here, we discuss the metrics designed to evaluate the performance of proposal generation, temporal action detection, and spatio-temporal action detection. 

\vspace{0.1in}
\textbf{Temporal Action Proposal Generation}. For this task, Average Recall (AR)  with multiple IoU thresholds is usually used as evaluation metrics. 
Most methods use IoU thresholds set $[0.5$ : $0.05$ : $0.95]$ in ActivityNet-1.3 \cite{caba2015activitynet} and $[0.5:0.05:1.0]$ in THUMOS14 \cite{jiang2014thumos}.
To evaluate the relation between recall and proposals number, most methods evaluate AR with Average Number of proposals (AN) on both datasets, which is denoted as AR@AN. On ActivityNet-1.3, area under the AR vs. AN curve (AUC) is also used as metrics, where AN varies from $0$ to $100$.

\vspace{0.1in}
\textbf{Temporal Action Detection}. For this task, mean Average Precision (mAP) is used as evaluation metric, where Average Precision (AP) is calculated on each action class, respectively. On ActivityNet-1.3 \cite{caba2015activitynet}, mAP with IoU thresholds $\{0.5, 0.75, 0.95\}$ and average mAP with IoU thresholds set $[0.5 : 0.05 : 0.95]$ are often used. On THUMOS14\cite{jiang2014thumos}, mAP with IoU thresholds $\{0.3, 0.4, 0.5, 0.6, 0.7\}$ is used.

\vspace{0.1in}
\textbf{Spatio-temporal Action Detection}. Two metrics are frequently used for this task. First, \textit{frame-AP} measures the area under the precision-recall curve of the detections for each frame. A detection is correct if the intersection-over-union with the ground truth at that frame is greater than a threshold and the action label is correctly predicted. Second, \textit{video-AP} measures the area under the precision-recall curve of the action tubes predictions. A tube is correct if the mean per frame intersection-over-union with the ground truth across the frames of the video is greater than a threshold and the action label is correctly predicted.

\subsection{Performance Analysis}
\label{Performance Analysis}

Action detection results of the state-of-the-art methods on THUMOS14 \cite{jiang2014thumos} and ActivityNet \cite{caba2015activitynet} dataset are compared by mAP (\%) in Tables \ref{performance-comp_thumos} and \ref{performance-comp_anet} respectively. The methods are categorized to fully-supervised, weakly-supervised, semi-supervised, self-supervised and US (unsupervised). We also summarize the advantageous and limitations of fully-supervised methods and methods with limited supervision in Tables \ref{tab:methods-summary} and \ref{tab:LS-methods-summary}.

\begin{table}[!htb]
\centering
\caption{Action detection results of the-state-of-the-art on testing set of THUMOS-14, measured by mAP (\%) at tIoU thresholds.}
\resizebox{0.98\columnwidth}{!}{%
\begin{tabular}{|c |l |  c c c c c| }
 \hline
Supervision & Method   & 0.3	&0.4	&0.5	&0.6	&0.7\\
\hline
\multicolumn{1}{|c|}{\multirow{32}{*}{\makecell{Fully\\supervised}}} & Yeung \textit{et al.} \cite{yeung2016end}  &  36.0 &   26.4 &   17.1 & - & - \\
\multicolumn{1}{|c|}{}& SMS \cite{yuan2017temporal}  &  36.5  &  27.8  &  17.8 & - & - \\
\multicolumn{1}{|c|}{}& SCNN \cite{shou2016temporal}    & 36.3	&28.7	&19	& - 	& -  \\
\multicolumn{1}{|c|}{}& Sst \cite{buch2017sst}  & - 	& - 	&23.0	& - 	& -  \\
\multicolumn{1}{|c|}{}& CDC \cite{shou2017cdc}   & 40.1	&29.4	&23.3	&13.1	&7.9\\
\multicolumn{1}{|c|}{}& SSAD \cite{lin2017single}   &43	&35	&24.6	& - 	& - \\
\multicolumn{1}{|c|}{}& TCN \cite{dai2017temporal}   & - & 33.3 &   25.6  &  15.9 & 9.0 \\
\multicolumn{1}{|c|}{}& TURN TAP \cite{gao2017turn}   &44.1	&34.9	&25.6	& - 	& - \\
\multicolumn{1}{|c|}{}& R-C3D \cite{xu2017r}   &44.8	&35.6	&28.9	& - 	& - \\
\multicolumn{1}{|c|}{}& SS-TAD \cite{buch2019end}   &45.7	& - 	&29.2	& - 	&9.6 \\
\multicolumn{1}{|c|}{}& SSN \cite{zhao2017temporal}   & 51.9	&41.0	&29.8	& - 	& - \\
\multicolumn{1}{|c|}{}& CTAP \cite{gao2018ctap}  	& - 	& - 	&29.9	& - 	& -  \\
\multicolumn{1}{|c|}{}& CBR	\cite{gao2017cascaded}   & 50.1	&41.3	&31.0	&19.1	&9.9 \\ 
\multicolumn{1}{|c|}{}& S3D\cite{zhang2018s3d}   & 47.9 & 41.2 & 32.6 &   23.3  &  14.3 \\
\multicolumn{1}{|c|}{}&  DBS \cite{gao2019video} &  50.6  & 43.1 & 34.3  & 24.4 &  14.7 \\
\multicolumn{1}{|c|}{}& BSN \cite{lin2018bsn}   	& 53.5	&45.0	&36.9	&28.4	&20.0  \\
\multicolumn{1}{|c|}{}& MGG \cite{liu2019multi}   	& 53.9	&46.8	&37.4	&29.5	&21.3\\
\multicolumn{1}{|c|}{}& AGCN \cite{li2020graph}  &  57.1 & 51.6 & 38.6 & 28.9 & 17.0 \\
\multicolumn{1}{|c|}{}& GTAN \cite{long2019gaussian}   & 57.8  &  47.2 & 38.8 & - & - \\
\multicolumn{1}{|c|}{}& BMN \cite{lin2019bmn}  	& 56.0	&47.4	&38.8	&29.7	&20.5 \\
\multicolumn{1}{|c|}{}& SRG\cite{eun2019srg}   & 54.5 & 46.9 & 39.1 & 31.4 & 22.2 \\
\multicolumn{1}{|c|}{}& DBG	\cite{lin2020fast}    & 57.8	&49.4	&39.8	&30.2	&21.7 \\
\multicolumn{1}{|c|}{}& G-TAD \cite{xu2020g}    &54.5	&47.6	&40.2	&30.8	&23.4 \\
\multicolumn{1}{|c|}{}& BC-GNN \cite{bai2020boundary}   & 57.1 & 49.1 & 40.4 & 31.2 & 23.1 \\
\multicolumn{1}{|c|}{}& BSN++ \cite{su2020bsn++}  &  59.9  &  49.5 & 41.3 & 31.9  &  22.8 \\
\multicolumn{1}{|c|}{}& TAL-Net \cite{chao2018rethinking}   &53.2	&48.5	&42.8	&33.8 & 20.8\\
\multicolumn{1}{|c|}{}& TSA-Net \cite{gong2020scale}   & 55.8 & 52.0 & 44.1 & 33.0 & 21.8 \\
\multicolumn{1}{|c|}{}& BU \cite{zhaobottom}    & 53.9 & 50.7 & 45.4 & 38.0 & 28.5 \\
\multicolumn{1}{|c|}{}& A2Net \cite{yang2020revisiting}  &  58.6  &54.1  &  45.5 &  32.5 &  17.2 \\
\multicolumn{1}{|c|}{}&  ATAG \cite{chang2021augmented} & 62.0  & 53.1  &  47.3  & 38.0 & 28.0  \\
\multicolumn{1}{|c|}{}& Lianli \textit{et al.} \cite{gao2020play}  & 66.4 & 58.4 & 48.8 & 36.7 & 25.5 \\
\multicolumn{1}{|c|}{}& PGCN \cite{zeng2019graph}   & 63.6	&57.8	&49.1	& - & -  \\
\multicolumn{1}{|c|}{}&  TadTR \cite{liu2021end} &  62.4  & 57.4 & 49.2  & 37.8  & 26.3  \\
\multicolumn{1}{|c|}{}& AFNet \cite{chen2020afnet}  & 63.4 & 58.5 & 49.5 & 36.9 & 23.5 \\
\multicolumn{1}{|c|}{}& AGT \cite{nawhal2021activity}   &  65.0 & 58.1 & 50.2 &  & \\
\multicolumn{1}{|c|}{}& PBRNet \cite{liu2020progressive}  & 58.5	&54.6	&51.3	&41.8	&29.5\\
\multicolumn{1}{|c|}{}&  RTD-Net\cite{tan2021relaxed} &  68.3  & 62.3 & 51.9  & 38.8  & 23.7  \\
\multicolumn{1}{|c|}{}&C-TCN \cite{li2019deep}   &  68.0 & 62.3 & 52.1 & - & - \\
\multicolumn{1}{|c|}{}& VSGN \cite{zhao2020video} &  66.7  &  60.4  & 52.4 &  41.0 & 30.4   \\
\multicolumn{1}{|c|}{}& MLTPN \cite{wang2020multi}  &  66.0  & 62.6  & 53.3  & 37.0  & 21.2 \\
\multicolumn{1}{|c|}{}& TSP \cite{alwassel2020tsp}  &  69.1  & 63.3  & 53.5 &  40.4  & 26.0 \\
\multicolumn{1}{|c|}{}& DaoTAD \cite{wang2021rgb}  &  62.8 &  59.5 &  53.8 & 43.6  & 30.1 \\
\multicolumn{1}{|c|}{}&  AFSD \cite{lin2021learning} &  67.3  &62.4  & 55.5  & 43.7  & 31.1  \\
\multicolumn{1}{|c|}{}& SP-TAD \cite{wu2021towards}  & 69.2   & 63.3 & 55.9  & 45.7 &  33.4  \\
\multicolumn{1}{|c|}{}& Liu \textit{et al.}\cite{liu2021multi}  &  68.9  & 64.0 & 56.9  & 46.3  & 31.0 \\

\hline
\multicolumn{1}{|c|}{\multirow{25}{*}{\makecell{Weakly\\supervised}}} & Hide-Seek \cite{singh2017hide} 	&19.5	&12.7	&6.8	& -	& -	\\
\multicolumn{1}{|c|}{}& UNet \cite{wang2017untrimmednets} &28.2	&21.1	&13.7	& -	& -	\\
\multicolumn{1}{|c|}{}& Step-by-step \cite{zhong2018step}	&31.1	&22.5	&15.9	& -	& -	\\
\multicolumn{1}{|c|}{}& STPN \cite{nguyen2018weakly}  	&35.5	&25.8	&16.9	&9.9	&4.3 \\
\multicolumn{1}{|c|}{}& MAAN \cite{yuan2019marginalized} 	&41.1	&30.6	&20.3	&12	&6.9 \\
\multicolumn{1}{|c|}{}& AutoLoc \cite{shou2018autoloc}  & 35.8	&29	&21.2	&13.4	&5.8	\\
\multicolumn{1}{|c|}{}& W-TALC \cite{paul2018w}  	&40.1	&31.1	&22.8	& -	&7.6  \\
\multicolumn{1}{|c|}{}& STAR \cite{xu2019segregated} &	48.7&34.7& 23 &	 -	& - \\
\multicolumn{1}{|c|}{}& CMCS \cite{liu2019completeness}  &41.2	&32.1	&23.1	&15	&7\\
\multicolumn{1}{|c|}{}& AdapNet \cite{zhang2020adapnet} & 41.09  & 31.61 & 23.65 & 14.53 & 7.75 \\
\multicolumn{1}{|c|}{}& Cleannet \cite{liu2019weakly} &37	&30.9	&23.9	&13.9	&7.1	\\
\multicolumn{1}{|c|}{}& TSM  \cite{yu2019temporal} 	&39.5	& 31.9	&24.5 & 13.8	&7.1	\\
\multicolumn{1}{|c|}{}& 3C-Net \cite{narayan20193c}  &40.9	&32.3	&24.6	& -	&7.7\\
\multicolumn{1}{|c|}{}& Shen \textit{et al} \cite{shen2020weakly}  & 44 & 34.4 & 25.5 & 15.2 & 7.2 \\
\multicolumn{1}{|c|}{}& Action Graphs \cite{rashid2020action} &47.3& 36.4&26.1& -	&	 -\\
\multicolumn{1}{|c|}{}& BG modeling \cite{nguyen2019weakly} &46.6	&37.5&26.8&17.6	&9	\\
\multicolumn{1}{|c|}{}& BaSNet	\cite{lee2020background} &44.6	&36	&27	&18.6	&10.4 \\
\multicolumn{1}{|c|}{}& RPN	\cite{huang2020relational}	&48.2	&37.2	&27.9	&16.7	&8.1 \\
\multicolumn{1}{|c|}{}&  TSCN \cite{zhai2020two} &    47.8 & 37.7 & 28.7  & 19.4  & 10.2 \\
\multicolumn{1}{|c|}{}& DGAM \cite{shi2020weakly} &46.8	&38.2	&28.8	&19.8	&11.4	\\
\multicolumn{1}{|c|}{}& ECM \cite{yang2020equivalent} &  46.5 & 38.2 &  29.1 &  19.5 &  10.9 \\
\multicolumn{1}{|c|}{}& Deep Metric \cite{islam2020weakly}	&46.8& - &29.6 & -	&	9.7 \\
\multicolumn{1}{|c|}{}& A2CL-PT \cite{min2020adversarial} & 48.1 & 39.0 & 30.1 & 19.2 & 10.6 \\
\multicolumn{1}{|c|}{}& EM-MIL \cite{luo2020weakly} & 45.5 & 36.8 & 30.5 & 22.7 & 16.4 \\
\multicolumn{1}{|c|}{}& Lee \textit{et al} \cite{lee2020backgrounduncertain} & 46.9 & 39.2 & 30.7 & 20.8 & 12.5 \\
\multicolumn{1}{|c|}{}& ASL \cite{ma2021weakly}  &  51.8  & - & 31.1   & - & 11.4 \\

\multicolumn{1}{|c|}{}& Huang \textit{et al} \cite{huang2021modeling} & 49.1  & 40.0 & 31.4   & 18.8 & 10.6 \\

\multicolumn{1}{|c|}{}& Ding \textit{et al} \cite{ding2020weakly} & 48.2 & 39.7 & 31.6 & 22.0 & 13.8 \\
\multicolumn{1}{|c|}{}& CoLA \cite{zhang2021cola}  &  51.5 & 41.9 & 32.2 & 22.0 & 13.1 \\
\multicolumn{1}{|c|}{}& Acsnet \cite{liu2021acsnet}  &  51.4  & 42.7 & 32.4  &  22.0 & 11.7 \\
\multicolumn{1}{|c|}{}&  Lee \textit{et al.} \cite{lee2020weakly} &  52.3  & 43.4 & 33.7  & 22.9 & 12.1 \\
\multicolumn{1}{|c|}{}& ACM-Net \cite{qu2021acm} & 55.0   & 44.6 & 34.6  &  21.8  & 10.8 \\
\multicolumn{1}{|c|}{}&  D2-Net \cite{narayan2020d2} & 52.3  &  43.4  & 36.0 & - & - \\
\hline

\multicolumn{1}{|c|}{\multirow{2}{*}{\makecell{Semi\\supervised}}}&  TTC-Loc \cite{lin2019towards}  & 52.8   & 44.4  & 35.9   &  24.7  &   13.8\\

\multicolumn{1}{|c|}{}& Ji \textit{et al} \cite{ji2019learning} \hspace{0.27in} &  53.4  &  45.2  &  37.2  &  29.5   &  20.5 \\

% \multicolumn{1}{|c|}{}&  &  &  &   &   &   \\

\hline
\multicolumn{1}{|c|}{\multirow{2}{*}{\makecell{Self\\supervised}}}& Actionbytes \cite{jain2020actionbytes} \hspace{0.27in} &  43.0  & 35.8 & 29.0 & - &  9.5\\

\multicolumn{1}{|c|}{}& Gong \textit{et al.} \cite{gongself} & 50.8 & 42.2 & 32.9 & 21.0 & 10.1\\

\hline
\multicolumn{1}{|c|}{\multirow{1}{*}{US}}& ACL \cite{gong2020learning} & 39.6 &  32.9  & 25.0  & 16.7 &  8.9 \\
\hline 
\end{tabular}}
\label{performance-comp_thumos}
\end{table}

\begin{table}[h]
\centering
\caption{Action detection results of the-state-of-the-art on validation set of ActivityNet (V is the version), measured by mAP (\%) at tIoU thresholds. $\star$ indicates utilization of weaker feature extractor (UNet \cite{wang2017untrimmednets}).}
%The methods are categorized to Full (fully-supervised),  Weak (weakly-supervised), Semi (semi-supervised), SLS (self-supervised) and US (unsupervised). 
\resizebox{0.98\columnwidth}{!}{%
\begin{tabular}{|c |l | c c c c c |}
 \hline
Supervision & Method & V & 0.5 & 0.75 & 0.95 & Average \\
\hline
\multicolumn{1}{|c|}{\multirow{17}{*}{\makecell{Fully\\supervised}}}& R-C3D \cite{xu2017r} & 1.3 & 26.8 & - & - & 12.7 \\
\multicolumn{1}{|c|}{}& AFNet \cite{chen2020afnet} & 1.3 & 36.1 & 17.8 & 5.2 & 18.6 \\
\multicolumn{1}{|c|}{}& TAL-Net \cite{chao2018rethinking} & 1.3 & 38.23 &  18.30 & 1.30 & 20.22 \\
\multicolumn{1}{|c|}{}& TCN \cite{dai2017temporal} & 1.3  & 37.49  & 23.47  &  4.47  & 23.58 \\
\multicolumn{1}{|c|}{}& CDC \cite{shou2017cdc} & 1.3   & 45.3  & 26.0 & 0.2 & 23.8 \\
\multicolumn{1}{|c|}{}& SSN \cite{zhao2017temporal} & 1.3   & 39.12 & 23.48  & 5.49 & 23.98  \\
\multicolumn{1}{|c|}{}&  DBS \cite{gao2019video} & 1.3 & 43.2 & 25.8 & 6.1 & 26.1\\ 
\multicolumn{1}{|c|}{}& A2Net \cite{yang2020revisiting}  & 1.3  & 43.55 &  28.69 & 3.7 & 27.75 \\
\multicolumn{1}{|c|}{}& MLTPN \cite{wang2020multi} & 1.3  &  44.86 &  28.96 & 4.30  & 28.27   \\
\multicolumn{1}{|c|}{}& SRG\cite{eun2019srg} & 1.3  & 46.53 & 29.98 & 4.83 & 29.72 \\
\multicolumn{1}{|c|}{}& BSN \cite{lin2018bsn} & 1.3  & 46.45 & 29.96  & 8.02 & 30.03 \\
\multicolumn{1}{|c|}{}& BU \cite{zhaobottom} & 1.3   & 43.47&  33.91 & 9.21 & 30.12\\
\multicolumn{1}{|c|}{}& AGCN \cite{li2020graph}  &  1.3  & - & - & - & 30.4 \\
\multicolumn{1}{|c|}{}&  RTD-Net\cite{tan2021relaxed} & 1.3  & 47.21 & 30.68  & 8.61  &   30.83\\
\multicolumn{1}{|c|}{}& Lianli \textit{et al.} \cite{gao2020play} & 1.3  & 47.01  & 30.52 & 8.21&  30.88 \\
\multicolumn{1}{|c|}{}&C-TCN \cite{li2019deep} & 1.3  & 47.6  & 31.9 & 6.2 & 31.1 \\
\multicolumn{1}{|c|}{}& PGCN \cite{zeng2019graph} & 1.3 & 48.26 &  33.16 & 3.27 &  31.11 \\
\multicolumn{1}{|c|}{}&  TadTR \cite{liu2021end} & 1.3 & 49.08   & 32.58  & 8.49  & 32.27 \\
\multicolumn{1}{|c|}{}& SP-TAD \cite{wu2021towards}  &  1.3 & 50.06  & 32.92  & 8.44  & 32.99 \\
\multicolumn{1}{|c|}{}& BMN \cite{lin2019bmn} & 1.3  & 50.07 &  34.78 & 8.29 & 33.85 \\
\multicolumn{1}{|c|}{}& Liu \textit{et al.}\cite{liu2021multi}  & 1.3 & 50.02 & 34.97 & 6.57 & 33.99\\
\multicolumn{1}{|c|}{}& G-TAD \cite{xu2020g}& 1.3  & 50.36  & 34.60&  9.02&  34.09 \\
\multicolumn{1}{|c|}{}& BC-GNN \cite{bai2020boundary} & 1.3 & 50.56 &  34.75 & 9.37 & 34.26 \\
\multicolumn{1}{|c|}{}& GTAN \cite{long2019gaussian} & 1.3  & 52.61  & 34.14 & 8.91 & 34.31 \\
\multicolumn{1}{|c|}{}&  AFSD \cite{lin2021learning} & 1.3  & 52.4 &  35.3 & 6.5  & 34.4 \\
\multicolumn{1}{|c|}{}&  ATAG \cite{chang2021augmented} & 1.3  & 50.92  & 35.35  & 9.71   & 34.68 \\
\multicolumn{1}{|c|}{}& BSN++ \cite{su2020bsn++}  &  1.3 & 51.27 & 35.70  & 8.33  & 34.88 \\
\multicolumn{1}{|c|}{}& PBRNet \cite{liu2020progressive} & 1.3 & 53.96 & 34.97&  8.98 & 35.01 \\
\multicolumn{1}{|c|}{}& VSGN \cite{zhao2020video} &  1.3  &  52.38  & 36.01  & 8.37   &  35.07   \\
\multicolumn{1}{|c|}{}& TSP \cite{alwassel2020tsp} & 1.3  &  51.26  & 37.12  &  9.29 & 35.81   \\

\hline
\multicolumn{1}{|c|}{\multirow{17}{*}{\makecell{Weakly\\supervised \\ (V=1.2)}}} & UNet$^\star$ \cite{wang2017untrimmednets} & 1.2 & 7.4 & 3.2  & 0.7 & 3.6 \\
\multicolumn{1}{|c|}{}& Step-by-step \cite{zhong2018step} & 1.2 & 27.3 & 14.7 & 2.9 & 15.6 \\
\multicolumn{1}{|c|}{}& {AutoLoc}$^\star$ \cite{shou2018autoloc}  & 1.2  & 27.3 &  15.1 & 3.3  & 16.0  \\
\multicolumn{1}{|c|}{}& TSM \cite{yu2019temporal}  & 1.2 & 28.3  & 17.0  & 3.5 & 17.1 \\
\multicolumn{1}{|c|}{}& Action Graphs \cite{rashid2020action} & 1.2 & 29.4 & - & - &-   \\
\multicolumn{1}{|c|}{}& W-TALC \cite{paul2018w}  & 1.2 & 37.0 & & & 18.0  \\ 
\multicolumn{1}{|c|}{}& EM-MIL \cite{luo2020weakly}  & 1.2 & 37.4 & - & - & 20.3 \\
\multicolumn{1}{|c|}{}& Cleannet \cite{liu2019weakly}  & 1.2 & 37.1 & 20.3  & 5.0 & 21.6 \\
\multicolumn{1}{|c|}{}& 3C-Net \cite{narayan20193c}  & 1.2 & 37.2 &  - & - &  21.7\\
\multicolumn{1}{|c|}{}& Deep Metric \cite{islam2020weakly}  & 1.2 & 35.2 & - &- &-  \\
\multicolumn{1}{|c|}{}& CMCS \cite{liu2019completeness}  & 1.2 &  36.8 &  22.0 &  5.6 &  22.4\\ 
\multicolumn{1}{|c|}{}& Shen \textit{et al} \cite{shen2020weakly}  & 1.2 & 36.9 & 23.1 & 3.4 & 22.8  \\
\multicolumn{1}{|c|}{}& RPN \cite{huang2020relational}  & 1.2	& 37.6  &23.9 & 5.4  & 23.3 \\ %U 37.0 21.1 5.2 22.0
\multicolumn{1}{|c|}{}&  TSCN \cite{zhai2020two} & 1.2  & 37.6 &  23.7 & 5.7 &  23.6 \\
\multicolumn{1}{|c|}{}& BaSNet \cite{lee2020background} & 1.2 & 38.5  & 24.2 & 5.6 & 24.3 \\
\multicolumn{1}{|c|}{}& DGAM \cite{shi2020weakly}  & 1.2 & 41.0 & 23.5  & 5.3 & 24.4 \\
\multicolumn{1}{|c|}{}& Acsnet \cite{liu2021acsnet}  & 1.2  & 41.0 &  23.5 & 5.3  & 24.4   \\
\multicolumn{1}{|c|}{}& ECM \cite{yang2020equivalent}  & 1.2 & 41.0 & 24.9 & 6.5 & 25.5 \\
\multicolumn{1}{|c|}{}& ASL \cite{ma2021weakly}  &  1.2 & 40.2 &  - & - &  25.8 \\
\multicolumn{1}{|c|}{}&  Lee \textit{et al.} \cite{lee2020weakly} &   1.2 & 41.2 &  25.6 &  6.0 & 25.9 \\
\multicolumn{1}{|c|}{}&  D2-Net \cite{narayan2020d2} &  1.2 &  42.3 & 25.5  & 5.8  & 26.0 \\
\multicolumn{1}{|c|}{}& CoLA \cite{zhang2021cola} &  1.2 & 42.7 & 25.7 & 5.8 & 26.1 \\
\multicolumn{1}{|c|}{}& Ding \textit{et al} \cite{ding2020weakly} & 1.2 & 41.7 & 26.7 & 6.3 & 26.4 \\

\hline
\multicolumn{1}{|c|}{\multirow{7}{*}{\makecell{Weakly\\supervised \\ (V=1.3)}}}& STPN \cite{nguyen2018weakly}  & 1.3 & 29.3 & 16.9&  2.6& - \\
\multicolumn{1}{|c|}{}& STAR \cite{xu2019segregated}  & 1.3 & 31.1  & 18.8 & 4.7& - \\
\multicolumn{1}{|c|}{}& AdapNet \cite{zhang2020adapnet} & 1.3 & 33.61 & 18.75 & 3.40 & 21.97 \\
\multicolumn{1}{|c|}{}& MAAN \cite{yuan2019marginalized}  & 1.3 & 33.7 & 21.9 & 5.5 & - \\
\multicolumn{1}{|c|}{}& BG modeling \cite{nguyen2019weakly}  & 1.3 & 36.4 & 19.2 & 2.9& - \\
\multicolumn{1}{|c|}{}& A2CL-PT \cite{min2020adversarial}  & 1.3 & 36.8 & 22.0 & 5.2  & 22.5 \\
\multicolumn{1}{|c|}{}& Huang \textit{et al} \cite{huang2021modeling} & 1.3  & 36.5 & 22.8   & 6.0 & 22.9 \\
\multicolumn{1}{|c|}{}& ACM-Net \cite{qu2021acm} & 1.3 & 40.1 & 24.2  & 6.2  & 24.6 \\

\hline

\multicolumn{1}{|c|}{\multirow{1}{*}{\makecell{Semi}}}&  TTC-Loc \cite{lin2019towards}  & 1.2   & 40.6   & 3.6    &  5.3  &   24.5 \\

\hline
\multicolumn{1}{|c|}{\multirow{2}{*}{\makecell{Self\\supervised}}}& Actionbytes \cite{jain2020actionbytes}  \hspace{0.27in} & 1.2 & 39.4 & - & -& - \\
\multicolumn{1}{|c|}{}& Gong \textit{et al.} \cite{gongself} & 1.2 & 45.5 & 27.3 & 5.4 & 27.6 \\

\hline
\multicolumn{1}{|c|}{\multirow{1}{*}{\makecell{US}}}& ACL \cite{gong2020learning}  & 1.2 &35.2 & 21.4  &3.1  & 21.1 \\
\hline 
\end{tabular}}
\label{performance-comp_anet}
\end{table}

%---------------------- point weakly thumos-------------------
% \multicolumn{1}{c|}{}&  SF-Net \cite{ma2020sf} &  53.2  &40.7  & 29.3  & 18.4  & 9.6 \\
% \multicolumn{1}{c|}{}& PTAL \cite{ju2020point}  & 58.2   & 47.1 & 35.9  & 23.0 & 12.8  \\ 

\subsubsection{Fully-supervised Methods}

\textbf{Proposal Generation.} Anchor-free methods such as SSN \cite{zhao2017temporal}, BSN \cite{lin2018bsn}, BMN \cite{lin2019bmn}, DBG \cite{lin2020fast}, BC-GNN \cite{bai2020boundary}, BU \cite{zhaobottom}, and BSN++ \cite{su2020bsn++}, A2Net \cite{yang2020revisiting} and AFSD \cite{lin2021learning} achieved superior results compared with anchor-based methods such as Yeung \textit{et al.} \cite{yeung2016end}, SMS \cite{yuan2017temporal}, TCN \cite{dai2017temporal}, SCNN \cite{shou2016temporal}, TURN TAP \cite{gao2017turn}, CBR \cite{gao2017cascaded}, and CDC \cite{shou2017cdc}. This is because anchor-free methods generate temporal action proposals with more flexibility and precise temporal boundaries. Some methods such as CTAP\cite{gao2018ctap}, MGG\cite{liu2019multi}, PBRNet \cite{liuprogressive}, SRG\cite{eun2019srg}, and RapNet \cite{gao2020accurate} combine advantageous of anchor-based and anchor-free methods and attained a higher results. 

\textbf{Proposal Feature Extraction}. R-C3D \cite{xu2017r} and AFNet \cite{chen2020afnet} employ 3D RoI pooling for feature extraction and obtained low results on ActivityNet due to lack of receptive field alignment with proposal span. TAL-Net \cite{chao2018rethinking}, TSA-Net \cite{gong2020scale} employ a multi-tower network and achieve a higher performance compared with 3D RoI pooling methods. The methods of SSAD \cite{lin2017single}, S3D\cite{zhang2018s3d},  MGG \cite{liu2019multi}, PBRNet \cite{liu2020progressive}, MLTPN \cite{wang2020multi}, C-TCN \cite{li2019deep}, RapNet \cite{gao2020accurate}, SP-TAD \cite{wu2021towards}, and DaoTAD \cite{wang2021rgb} employ temporal feature pyramid to extract features from actions with different duration and achieved superior performance. 

%PBRNet \cite{liu2020progressive} and C-TCN \cite{li2019deep} further aggregate the course and fine-grained features extracted by temporal feature pyramid and achieve superior performance. 

\textbf{Modeling Long-term Dependencies.} Sst \cite{buch2017sst} and SS-TAD \cite{buch2019end} which are RNN-based methods achieve relatively lower results as they can not generate flexible proposals. PGCN \cite{zeng2019graph}, G-TAD \cite{xu2020g}, BC-GNN \cite{bai2020boundary}, AGCN \cite{li2020graph}, ATAG \cite{chang2021augmented}, and VSGN \cite{zhao2020video} are graph models that capture dependencies between proposals or video
segments. Among them VSGN \cite{zhao2020video} achieved the best performance by exploiting correlations between cross-scale snippets (original and magnified) and aggregating their features with a graph pyramid network. AGT \cite{nawhal2021activity}, RTD-Net \cite{tan2021relaxed}, ATAG \cite{chang2021augmented}, and TadTR \cite{liu2021end} use transformers to model long-range dependencies. Among them RTD-Net \cite{tan2021relaxed} achieved the best results (on THUMOS14) by customizing the encoder with a boundary-attentive architecture to enhance the discrimination capability of action boundary.

There are also two state-of-the-art (SOTA) methods that do not belong to the mentioned categories of methods. TSP \cite{alwassel2020tsp} proposed a novel supervised pretraining paradigm for clip features, and improved the performance of SOTA using features trained with the proposed pretraining strategy. Liu \textit{et al.} in \cite{liu2021multi} leverages temporal aggregation to improve the feature discriminative power of each snippet and enhance the feature coherence within a single instance.

\begin{table*}[h]
\caption{Summary of fully-supervised methods for temporal action detection. $(+)$ and $(-)$ denote the advantages and disadvantages.}
\centering
\resizebox{2\columnwidth}{!}{%
\begin{tabular}{|c|c|l|l|}
\hline
Objective &  Category &   Methods & Advantages and Limitations \\
\hline

& & & \\[-7pt]

\multicolumn{1}{|c|}{\multirow{9}{*}{\makecell{Proposal \\ Generation}}} &  
\multicolumn{1}{c|}{\multirow{2}{*}{ Anchor-based }} &  
\multicolumn{1}{c|}{\multirow{2}{*}{\makecell[c]{
SCNN \cite{shou2016temporal}, CBR\cite{gao2017cascaded}, \\ Turn-Tap\cite{gao2017turn}, CDC \cite{shou2017cdc}
}}}  & 

\makecell[l]{+ Efficiently generate multiple-scales proposals, use global  info \\ \hspace{4pt} of all anchors to generate reliable confidence scores.} \\

\multicolumn{1}{|c|}{} & \multicolumn{1}{c|}{} & \multicolumn{1}{c|}{} & 

\makecell[l]{- Proposals are not temporally flexible and precise.}   \\

\cline{2-4}
& & & \\[-7pt]

%CTAP \cite{gao2018ctap},\\ 

\multicolumn{1}{|c|}{}&  \multicolumn{1}{c|}{\multirow{5}{*}{ Anchor-free}} &  
\multicolumn{1}{c|}{\multirow{5}{*}{\makecell[c]{TAG \cite{zhao2017temporal}, BSN \cite{lin2018bsn},\\ BMN \cite{lin2019bmn}, DBG \cite{lin2020fast} \\
BC-GNN \cite{bai2020boundary}, BU \cite{zhaobottom} \\
A2Net \cite{yang2020revisiting}, AFSD \cite{lin2021learning}\\
BSN++ \cite{su2020bsn++}}}} &  \makecell[l]{+ Generate proposals with flexible duration.}

\\

\multicolumn{1}{|c|}{} & \multicolumn{1}{c|}{}  & \multicolumn{1}{c|}{} &  
\makecell[l]{+ Global context for proposal evaluation ( in BMN, DBG).}  \\

\multicolumn{1}{|c|}{} & \multicolumn{1}{c|}{}  & \multicolumn{1}{c|}{} &    \makecell[l]{+ Global context for proposal generation (in DBG).} \\

\multicolumn{1}{|c|}{}&  \multicolumn{1}{c|}{} &  \multicolumn{1}{c|}{}   &  
\makecell[l]{- Proposal evaluation is not efficient in some cases.}
\\

\multicolumn{1}{|c|}{}&  \multicolumn{1}{c|}{} &  \multicolumn{1}{c|}{}   &  
\makecell[l]{- Distorting the information of short actions due to down-scaling.}
\\

\cline{2-4}
& & & \\[-7pt]

\multicolumn{1}{|c|}{}&  \multicolumn{1}{c|}{\multirow{2}{*}{\makecell{Anchor-based +Anchor-free}}} &  

\multicolumn{1}{c|}{\multirow{2}{*}{\makecell[c]{
CTAP\cite{gao2018ctap}, MGG\cite{liu2019multi}\\ 
PBRNet \cite{liuprogressive}, RapNet \cite{gao2020accurate} 
}}} &  \makecell[l]{+ Combining advantageous of anchor-based and anchor-free.}\\

\multicolumn{1}{|c|}{} & \multicolumn{1}{c|}{}  &  \multicolumn{1}{c|}{}  & - Not modeling long-range dependencies. \\

\hline 
& & & \\[-7pt]

\multicolumn{1}{|c|}{\multirow{11}{*}{\makecell{Proposal \\ Feature \\ Extraction}}} &  \multicolumn{1}{c|}{\multirow{3}{*}{3D RoI pooling}} & 

\multicolumn{1}{c|}{\multirow{3}{*}{R-C3D \cite{xu2017r}, AFNet \cite{chen2020afnet} }} & 
\makecell[l]{+ Fast feature extraction from multi-scale proposals.} \\

\multicolumn{1}{|c|}{} & \multicolumn{1}{c|}{} & \multicolumn{1}{c|}{}  &
\makecell[l]{- Proposal features may include insufficient or irrelevant info\\ \hspace{3pt} because of receptive field misalignment.}\\

\cline{2-4}
& & & \\[-7pt]

\multicolumn{1}{|c|}{} &  \multicolumn{1}{c|}{\multirow{2}{*}{Multi-tower Network }} & \multicolumn{1}{c|}{\multirow{2}{*}{\makecell[c]{TAL-Net \cite{chao2018rethinking}, TSA-Net \cite{gong2020scale}}}}&
\makecell[l]{+ Alignment of receptive field to proposal span to extract rich\\ \hspace{4pt} features from proposals.} \\

\multicolumn{1}{|c|}{} & \multicolumn{1}{c|}{}   & \multicolumn{1}{c|}{}  &  \makecell[l]{- Pre-defined temporal intervals limit the accuracy of proposals.} \\

\cline{2-4}
& & & \\[-7pt]

\multicolumn{1}{|c|}{}  & \multicolumn{1}{c|}{\multirow{6}{*}{TFPN}} & 

\multicolumn{1}{c|}{\multirow{6}{*}{\makecell[c]{SSAD \cite{lin2017single}, S3D\cite{zhang2018s3d} \\ MGG \cite{liu2019multi}, C-TCN \cite{li2019deep} \\ MLTPN \cite{wang2020multi}, PBRNet \cite{liuprogressive}\\ 
A2Net \cite{yang2020revisiting}, AFSD \cite{lin2021learning}\\
RapNet \cite{gao2020accurate}, SP-TAD \cite{wu2021towards} \\
DaoTAD \cite{wang2021rgb} 
}}}&

+ Feature pyramids to detect different scales of actions.\\ 

\multicolumn{1}{|c|}{}  & \multicolumn{1}{c|}{} & \multicolumn{1}{c|}{} &    \makecell[l]{+ Re-fine the proposal boundaries from coarse to fine (in MGG, \\ \hspace{4pt} PBRNet, and RapNet).}\\ 

\multicolumn{1}{|c|}{}  & \multicolumn{1}{c|}{} &  \multicolumn{1}{c|}{} &   \makecell[l]{+ Combination with anchor-free pipeline for flexible and precise \\ \hspace{4pt} proposal generation (A2Net, AFSD).}  \\

\multicolumn{1}{|c|}{}  & \multicolumn{1}{c|}{} &  \multicolumn{1}{c|}{} &   \makecell[l]{- No modeling of temporal dependencies in most cases.}  \\

\hline
& & & \\[-7pt]

\multicolumn{1}{|c|}{\multirow{9}{*}{\makecell{Modeling\\ Long-term\\ Dependencies}}}  &  \multicolumn{1}{c|}{\multirow{2}{*}{RNNs}}  & \multicolumn{1}{c|}{\multirow{2}{*}{Sst \cite{buch2017sst}, SS-TAD \cite{buch2019end}}} & + Modeling long-term dependencies for proposal generation. \\

\multicolumn{1}{|c|}{} & \multicolumn{1}{c|}{} & \multicolumn{1}{c|}{} & 
- Proposals are not flexible and precise. \\

\cline{2-4}
& & & \\[-7pt]

\multicolumn{1}{|c|}{} & \multicolumn{1}{c|}{\multirow{3}{*}{  
Graphs}} & 

\multicolumn{1}{c|}{\multirow{3}{*}{\makecell[c]{
PGCN \cite{zeng2019graph}, G-TAD \cite{xu2020g},\\ BC-GNN \cite{bai2020boundary}, AGCN \cite{li2020graph} \\
ATAG \cite{chang2021augmented} , VSGN \cite{zhao2020video}
}}} &    
\makecell[l]{+ Modeling temporal dependencies between proposals or video \\ 
\hspace{4pt} segments for proposal generation and refinement.} \\

% \multicolumn{1}{|c|}{}  & \multicolumn{1}{c|}{}  & \multicolumn{1}{c|}{} &   \multicolumn{1}{c|}{} \\

\multicolumn{1}{|c|}{}  & \multicolumn{1}{c|}{}  & \multicolumn{1}{c|}{} &   \makecell[l]{- Proposal generation is inefficient or temporal dependencies are\\ \hspace{4pt} used only for proposal refinement.}\\

\cline{2-4}
& & & \\[-7pt]

\multicolumn{1}{|c|}{} & \multicolumn{1}{c|}{\multirow{2}{*}{Transformer}}   & \multicolumn{1}{c|}{\multirow{2}{*}{
\makecell{AGT \cite{nawhal2021activity}, RTD-Net \cite{tan2021relaxed} \\
ATAG \cite{chang2021augmented}, TadTR \cite{liu2021end}}}}   &

% \multicolumn{1}{l|}{\multirow{3}{*}{

\makecell[l]{+ Modeling non-linear temporal structure and inter-proposal \\\hspace{4pt} relationships for proposal generation.} \\ 

% }}\\

\multicolumn{1}{|c|}{} & \multicolumn{1}{|c|}{}  &  \multicolumn{1}{|c|}{} &  - High parametric complexity. \\

\hline
\end{tabular}
}
\label{tab:methods-summary}
\end{table*}

\subsubsection{Methods with Limited Supervision}

\textbf{Action Localization with Class-specific Attention.} UNet \cite{wang2017untrimmednets} is supervised with MIL loss which is not strong enough to predict accurate attention scores. The methods of W-TALC \cite{paul2018w}, Action Graphs \cite{rashid2020action}, and Deep Metric \cite{islam2020weakly} all target action-action separation by employing a co-activity similarity loss. 3C-Net \cite{narayan20193c} applied center loss on video-level aggregated features to enhance feature discriminability. Deep Metric \cite{islam2020weakly} outperforms W-TALC \cite{paul2018w}, Action Graphs \cite{rashid2020action}, and 3C-Net \cite{narayan20193c} by defining a class-specific metric for each action category.

\textbf{Action Localization with Class-agnostic Attention.} STPN \cite{nguyen2018weakly} proposed to learn attention through class-agnostic features but has a low performance as cross entropy loss alone does not train accurate attention signals. BG modeling \cite{nguyen2019weakly} used a clustering loss to separate action from background. BG modeling \cite{nguyen2019weakly} and BaSNet \cite{lee2020background} force all background frames to belong to one specific class which is not desirable as they do not share any common semantics. RPN \cite{huang2020relational}, and Huang \textit{et al.} \cite{huang2021modeling} increase inter-class separateness by pushing action (or sub-action) features to their prototypes. Huang \textit{et al.} \cite{huang2021modeling} outperforms RPN \cite{huang2020relational} by modeling the relations between sub-actions of each action. DGAM \cite{shi2020weakly} addressed the action-context confusion through imposing different attentions on different features with a generative model. EM-MIL \cite{luo2020weakly} employed Expectation-Maximization to  capture complete action instances and outperformed DGAM \cite{shi2020weakly} on THUMOS14 dataset.

\textbf{Direct Action Localization.} AutoLoc \cite{shou2018autoloc}, and CleanNet \cite{liu2019weakly} regress the intervals of action instances for proposal generation, instead of performing hard thresholding. They obtained a lower performance compared with most recent methods as they do not model action completeness or address action-context confusion.

\textbf{Action Completeness Modeling.} The methods of CMCS \cite{liu2019completeness}, Hide-and-Seek \cite{singh2017hide}, and Step-by-step \cite{zhong2018step} target the action completeness and CMCS \cite{liu2019completeness} achieves a superior performance. This is because Hide-and-Seek \cite{singh2017hide} and Step-by-step \cite{zhong2018step} do not guarantee the discovery of new parts by randomly hiding or removing different video regions. In contrary, CMCS \cite{liu2019completeness} employs a diversity loss to enforce the model to discover complementary action parts.

ACL \cite{gong2020learning} is an unsupervised method and only uses the total count of unique actions that appear in the video, but it still achieves a comparable performance with respect to some of the weakly-supervised methods such as 3C-Net \cite{narayan20193c}. Gong \textit{et al.} \cite{gongself} is a self-supervised method that attained the state-of-the-art results on ActivityNet-1.2 among methods with limited supervision, confirming the advantageous of self-supervised learning. The recent state-of-the-art weakly supervised methods such as D2-Net \cite{narayan2020d2} achieved comparable performance to the semi-supervised methods of Ji \textit{et al} \cite{ji2019learning} and TTC-Loc \cite{lin2019towards}. This is interesting specially because D2-Net \cite{narayan2020d2} does not use temporal annotation of actions at all while Ji \textit{et al} \cite{ji2019learning} and TTC-Loc \cite{lin2019towards} use temporal annotations at least for a small percentage of videos in the dataset.

\begin{table*}[!h]
\caption{Summary of methods with limited supervision for temporal action detection. $(+)$ and $(-)$ denote the advantages and disadvantages.} %"fg" and "bg" stand for foreground, and background, respectively. 
\centering
\resizebox{2\columnwidth}{!}{%
\begin{tabular}{|c|c|l|l|}
\hline
  Objective & Category  & Method & Advantages and Limitations \\
\hline

& & &  \\[-7pt]

 \multicolumn{1}{c|}{\multirow{9}{*}{ \makecell{Localization with \\ Class-specific \\ Attention} }} & \multicolumn{1}{|c|}{\multirow{3}{*}{MIL Loss}}&

\multicolumn{1}{c|}{\multirow{3}{*}{\makecell[c]{ UNet \cite{wang2017untrimmednets}, W-TALC \cite{paul2018w} \\ 
Action Graphs \cite{rashid2020action}, BaSNet \cite{lee2020background}\\ 3C-Net \cite{narayan20193c}, Actionbytes \cite{jain2020actionbytes}}}} 
& + Learns temporal class activation maps. \\

\multicolumn{1}{|c|}{} & \multicolumn{1}{c|}{} & \multicolumn{1}{|c|}{}   & - MIL loss alone does not predict accurate attention scores. \\

\multicolumn{1}{|c|}{} & \multicolumn{1}{c|}{} & \multicolumn{1}{|c|}{}& - Only supervising temporal positions with highest activation scores.\\

\cline{2-4}
& & &  \\[-7pt]

\multicolumn{1}{|c|}{} & \multicolumn{1}{|c|}{\multirow{3}{*}{\makecell{Co-activity Similarity \\ Loss (CASL)}}} &
\multicolumn{1}{c|}{\multirow{3}{*}{\makecell[c]{ W-TALC\cite{paul2018w}, \\ Action Graphs \cite{rashid2020action}\\ DM\cite{islam2020weakly}, Actionbytes \cite{jain2020actionbytes}}}}

& \makecell[l]{+ Action-background separation and reducing intra-class variations.} \\

\multicolumn{1}{|c|}{} & \multicolumn{1}{c|}{} & \multicolumn{1}{|c|}{}   & - Action-context confusion is not addressed.\\

\multicolumn{1}{|c|}{} & \multicolumn{1}{c|}{} & \multicolumn{1}{|c|}{}   & - Not modeling action completeness. \\

\cline{2-4}
& & &  \\[-7pt]

\multicolumn{1}{|c|}{} & \multicolumn{1}{c|}{\multirow{2}{*}{Center Loss}}  & 
 \multicolumn{1}{c|}{\multirow{2}{*}{\makecell[c]{ 3C-Net \cite{narayan20193c} }}}
& \makecell[l]{+ Reduce intra-class variations by pushing action features to class centers.} \\

\multicolumn{1}{|c|}{} & \multicolumn{1}{c|}{} & \multicolumn{1}{|c|}{}   & 
\makecell[l]{- Imprecise attention signal, supervises video-level aggregated features.}\\ %(not segment level)

% \\\hspace{4pt} 

\cline{1-4}
& & &  \\[-7pt]

\multicolumn{1}{|c|}{\multirow{8}{*}{\makecell{Localization with \\ Class-agnostic \\ Attention} }} &  \multicolumn{1}{c|}{\multirow{2}{*}{CE Loss}} & 
\multicolumn{1}{c|}{\multirow{2}{*}{\makecell[c]{ STPN \cite{nguyen2018weakly}, RPN \cite{huang2020relational}\\ BG modeling \cite{nguyen2019weakly} }}}
& + Learns attention through class-agnostic features. \\

\multicolumn{1}{|c|}{} & \multicolumn{1}{c|}{} & \multicolumn{1}{|c|}{}   & - CE loss alone does not train accurate attention signals.\\

\cline{2-4}
& & &  \\[-7pt]

\multicolumn{1}{|c|}{} & \multicolumn{1}{c|}{\multirow{2}{*}{Clustering Loss}} 
& \multicolumn{1}{c|}{\multirow{3}{*}{\makecell[c]{ RPN \cite{huang2020relational} \\
BG modeling \cite{nguyen2019weakly} }}}  & + Separating foreground-background features. \\

\multicolumn{1}{|c|}{} & \multicolumn{1}{c|}{} & \multicolumn{1}{|c|}{}   & \makecell[l]{- Force all background frames to belong to one specific class, but they do \\\hspace{4pt} not share any common semantics.} \\

\cline{2-4}
& & &  \\[-7pt]

\multicolumn{1}{|c|}{} & \multicolumn{1}{c|}{\multirow{2}{*}{Prototype Learning}} &  \multicolumn{1}{c|}{\multirow{2}{*}{\makecell[c]{ RPN \cite{huang2020relational}\\ Huang \textit{et al.} \cite{huang2021modeling} }}}
 & \makecell[l]{+ Inter-class separateness by pushing action (or sub-action) features to \\\hspace{4pt} their prototypes.} \\

 % Modeling relation between different actions. 
 
\multicolumn{1}{|c|}{} & \multicolumn{1}{c|}{} & \multicolumn{1}{|c|}{}   & - Action-context confusion is not addressed. \\

\cline{2-4}
& & &  \\[-7pt]

\multicolumn{1}{|c|}{} & \multicolumn{1}{c|}{\multirow{2}{*}{Generative Model}} &
\multicolumn{1}{c|}{\multirow{2}{*}{\makecell[c]{ DGAM \cite{shi2020weakly}\\ EM-MIL \cite{luo2020weakly} }}}&  
\makecell[l]{ + Conditional VAE / Expectation-Maximization to separate actions from \\\hspace{4pt} context frames and capture complete action instances.} \\

\multicolumn{1}{|c|}{} & \multicolumn{1}{c|}{} & \multicolumn{1}{|c|}{}   & - Not modeling temporal dependencies and relation between sub-actions.\\

\hline
& & &  \\[-7pt]

\multicolumn{1}{|c|}{\multirow{2}{*}{\makecell{Direct Localization }}} &  \multicolumn{1}{c|}{\multirow{2}{*}{\makecell{Action-boundary \\ Contrast} }}
&  
\multicolumn{1}{c|}{\multirow{2}{*}{\makecell[c]{ AutoLoc \cite{shou2018autoloc} \\ CleanNet \cite{liu2019weakly} }}}
& \makecell[l]{+ Regress the intervals of action instances for proposal generation, instead of \\\hspace{4pt} performing hard thresholding.}\\

\multicolumn{1}{|c|}{} & \multicolumn{1}{c|}{} & \multicolumn{1}{|c|}{}   & - Not modeling action completeness.\\

\hline
& & &  \\[-7pt]

\multicolumn{1}{|c|}{\multirow{4}{*}{\makecell{ Action \\ Completeness \\ Modeling }}} & \multicolumn{1}{c|}{\multirow{2}{*}{\makecell{Masking} }}
& 
\multicolumn{1}{c|}{\multirow{2}{*}{\makecell[c]{ Hide-and-seek \cite{singh2017hide} \\Step-by-step \cite{zhong2018step} }}}

& \makecell[l]{+ Randomly hiding or removing different video regions to see \\\hspace{4pt} different action parts. }\\

\multicolumn{1}{|c|}{} & \multicolumn{1}{|c|}{} & \multicolumn{1}{|c|}{}  & - Does not guarantee the discovery of new parts.\\

\cline{2-4}
& & &  \\[-7pt]

\multicolumn{1}{|c|}{} & \multicolumn{1}{|c|}{\multirow{2}{*}{\makecell{Diversity Loss}}} &  \multicolumn{1}{c|}{\multirow{2}{*}{\makecell[c]{ CMCS \cite{liu2019completeness} }}} & 
\makecell[l]{+ Enforcing the model to discover complementary pieces of an action.} \\

\multicolumn{1}{|c|}{} & \multicolumn{1}{|c|}{} & \multicolumn{1}{|c|}{}  & - Not modeling the relation between sub-actions. \\

\hline
\end{tabular}
}
\label{tab:LS-methods-summary}
\end{table*}

\section{Discussions}
\label{Discussions}

In this section, we describe the application of temporal action detection in real-world applications and introduce several directions for future work in this domain.

\subsection{Applications }

Temporal action detection has numerous real-world applications as most of the videos in practice are untrimmed with a sparse set of actions. In this section, we describe several applications such as understanding instructional videos, anomaly detection in surveillance videos, action spotting in sports, and detection in self-driving cars. 

\subsubsection{Action Localization in Instructional videos}

With the rising popularity of social media and video sharing sites such as YouTube, people worldwide upload numerous instructional videos in diverse categories. Millions of people watch these tutorials to learn new tasks such as "making pancakes" or "changing a flat tire." Analysis of the instructional videos has drawn more attention in recent years, leading to the proposition of several tasks including step localization and action segmentation \cite{rohrbach2012database}. Based on the psychological studies, it has been shown that simplifying and segmenting the video into smaller steps (sub-actions) is a more effective way to learn a new task \cite{tang2020comprehensive,nadolski2005optimizing}. For example, the task of "making pancakes" can be segmented to action steps such as "add the eggs," "pour the mixture into the pan," "heat a frying pan," and such. Many datasets are designed to study action localization and action anticipation such as EPIC-Kitchen \cite{damen2018scaling} and INRIA Instructional Videos Dataset \cite{alayrac2016unsupervised}. Both of these tasks (step localization and action segmentation) are directly related to action detection. Step localization is the task of localizing the start and endpoints of a series of steps and recognizing their labels while action segmentation is the frame-level labeling.

% , video captioning \cite{das2013thousand}, reference resolution\cite{huang2017unsupervised}, visual grounding \cite{huang2018finding}, procedure segmentation \cite{zhou2018towards}, activity anticipation\cite{abu2018will}, skill determination \cite{doughty2018s}, and procedure planning\cite{chang2019procedure}. 

\subsubsection{Anomaly Detection in Surveillance Videos}

Surveillance cameras are increasingly deployed in public places, monitoring the areas of interest to ensure security. With the stream of data from these video cameras, there has been a rise in video analysis and anomaly detection research. Anomalies are significant deviations of scene entities from normal behavior \cite{chandola2007outlier,chalapathy2019deep}. Fighting, traffic accidents, burglary, and robbery are a few examples of anomalies. Compared to normal activities, anomalous events rarely occur. Therefore, intelligent computer vision algorithms are required to detect anomalous events automatically, to avoid the waste of time and labor. In some methods, anomaly detection models are trained with normal behaviors to learn distributions of normal patterns. These models identify anomalous activities based on dissimilarity to the standard data distributions \cite{li2013anomaly,zhu2012context}. In other cases, normal and anomalous videos are used during training to automatically predict high anomaly scores \cite{sultani2018real,he2018anomaly}. In many real-time applications, the system must detect anomalous events as soon as each video frame arrives, only based on history and the current data; for instance, an intelligent video surveillance application designed to raise the alarm when suspicious activity is detected. To this end, online action detection algorithms are developed to accumulate historical observations and predicted future information to analyze current events \cite{sabokrou2018deep}, \cite{sabokrou2017deep}, \cite{sabokrou2018avid}, \cite{liu2019exploring},\cite{xu2019temporal}.

% \cite{shou2018online}

\subsubsection{Action Spotting in Sports }

Professional analysts utilize sports videos to investigate the strategies in a game, examine new players, and generate meaningful statistics. In order to analyze the videos, they watch many broadcasts to spot the highlights within a game, which is a time-consuming and costly process. Fortunately, automated sports analytic methods developed in the computer vision field can facilitate sports broadcasts understanding. In recent years, many automated methods have been proposed to help localize the salient actions of a game. They produce statistics of events within a game by either analyzing camera shots or semantic information. Human activity localization in sports videos is studied in \cite{bettadapura2016leveraging, heilbron2017scc, felsen2017will, kapela2014real}, salient game actions are identified in \cite{cioppa2018bottom, tsunoda2017football}, automatic game highlights identification and summarization are performed in \cite{cai2019temporal,sanabria2019deep,shukla2018automatic,tsagkatakis2017goal, turchini2019flexible}. Moreover, action spotting, which is the task of temporal localization of human-induced events, has been popular in soccer game broadcasts \cite{cioppa2020context,giancola2018soccernet} and some methods aimed to automatically detect goals, penalties, corner kicks, and card events \cite{huang2006semantic}. Action detection algorithms can inspire many of the tasks mentioned above. 
% TAPOS \cite{shao2020intra} is a new dataset developed on sport videos with manual annotations of sub-actions to study temporal action parsing. 
% focuses on instances of Olympics sport actions

\subsubsection{Action Detection in Autonomous Driving}

With the rapid development and advancement of cars and other vehicles in urban transportation, autonomous driving has attracted more attention in the last decades. The cameras assembled on the self-driving cars capture the real-time stream of videos that need to be processed with online algorithms. The car should be aware of the surrounding environment and spot road users, including pedestrians, cyclists, and other vehicles, to make safe autonomous decisions. Also, it should be able to detect and anticipate road users activities such as moving away, moving towards, crossing the road, and anomalous events in real-time to adjust the speed and handle the situation. Therefore, spatio-temporal action localization algorithms need to be developed to guarantee the safety of self-driving cars \cite{fontana2018action}. Yao \textit{et al.} \cite{yao2020and} proposed a traffic anomaly detection with a when-where-what pipeline to detect, localize, and recognize anomalous events from egocentric videos. To improve the detection and prediction of pedestrian movements, Rasouli \textit{et al.} \cite{rasouli2019autonomous} studied pedestrian behavior depending on various factors, such as demographics of the pedestrians, traffic dynamics, and environmental conditions. Moreover, Mahadevan \textit{et al.} \cite{mahadevan2019av} proposed an immersive VR-based pedestrian mixed traffic simulator to examine pedestrian behavior in street crossing tasks.

% \subsubsection{Action Detection in Sign Language Recognition}

% \subsubsection{Action Detection for Assisting Robots (medical purposes)}
% a robotic platform that must interact with humans in a realistic scenario, 

\subsection{Future work }

Weakly-supervised action localization in untrimmed videos has drawn much research attention by providing only video-level labels during training instead of exhaustive annotation of temporal boundaries in the training phase. Subsequently, knowledge transfer from publicly available trimmed videos is a promising trend to make up for the coarse-grained video-level annotations in weakly-supervised settings. Nevertheless, domain-adaptation schemes must fulfill the domain gap between trimmed and untrimmed videos to transfer robust and reliable knowledge. Only a few methods have explored knowledge transfer from trimmed videos \cite{jain2020actionbytes},  \cite{zhang2020adapnet}, \cite{cao2019action}, \cite{shi2019weakly}, but we expect to see more in the future.

In recent years, zero-shot learning (ZSL) in the visual recognition domain has been emerging as a rising trend as it is challenging to collect a large number of samples for each class during training. ZSL works by transferring the knowledge from the seen classes with sufficiently many instances to generalize the models on unseen classes with no samples during training. The task of zero-shot temporal activity detection (ZSTAD) is introduced in \cite{zhang2020zstad} to generalize the applicability of action detection methods to newly emerging or rare events that are not included in the training set. The task of ZSTAD is highly challenging because each untrimmed video in the testing set possibly contains multiple novel action classes that must be localized and detected. It is worth mentioning that activity detection with few-shot learning has been recently explored in \cite{jain2020actionbytes}, \cite{xu2018similarity}, \cite{xu2020revisiting}, \cite{yang2018one}, \cite{huang2019decoupling}, \cite{zhang2020metal}. The advancement of both zero-shot and few-shot action detection is anticipated in the near future.

% \vspace{0.1in}
% \textbf{Action Step Localization:}  

% Understanding the temporal configuration and semantic dependencies of sub-actions to assist the localization task.  

% \subsubsection{Action Detection with Few-shot Learning}

% \cite{xu2018similarity},\cite{xu2020revisiting}, 
% \cite{zhang2020zstad}, \cite{yang2018one},\cite{mettes2017spatial},
% \cite{huang2019decoupling},

% \subsubsection{Action Detection with Domain Adaptation}

% \cite{chen2020action}

% \subsubsection{Action Detection with Active Learning}
% \cite{bandla2013active}

% \subsubsection{Action Detection with Reinforcement Learning}

% \cite{vaudaux2020actionspotter} , \cite{li2018active} , \cite{huang2018sap}, \cite{wu2019multi},

% \subsubsection{Action Detection with Language Queries}

% \cite{zhu2020actbert},  \cite{richard2016temporal}
% text-to-video retrieval (YouCook2, MSR-VTT), 

\section{Conclusion}

Action detection schemes have expedited the progress in many real-world applications such as instructional video analysis, anomaly detection in surveillance videos, sports analysis, and autonomous driving. 
The advancement of learning methods with limited supervision has facilitated action detection by detachment from costly need to annotate the temporal boundary of actions in long videos. 
This survey has extensively studied recently developed deep learning-based methods for action detection from different aspects including 
fully-supervised schemes, methods with limited supervision, benchmark datasets, performance analysis, applications, and future directions. 
The performance analysis and future directions are summarized to inspire the design of new and efficient methods for action detection that serves the computer vision community.

% \cite{zhou2015interaction},\cite{ni2014multiple},\cite{lai2014video},,\cite{wang2016actionness},\cite{mettes2016spot},\cite{montes2016temporal},,\cite{lin2017temporal},\cite{yuan2017temporal},
% \cite{alwassel2018diagnosing},\cite{qiu2018precise},\cite{heidarivincheh2018action}, ,\cite{yang2018exploring},\cite{alwassel2018action},\cite{},\cite{bai2018contextual},\cite{ma2019cross}, \cite{cao2019action},\cite{dwibedi2019temporal},\cite{zhou2019temporal},\cite{iqbal2019enhancing},  \cite{chen2019relation},,\cite{mac2019learning}, \cite{hu2019cmsn},\cite{tang2019afo},\cite{li2019deep},,\cite{long2019gaussian}, ,,\cite{jin2020regression},\cite{zhang2019learning},,\cite{ma2020sf},\cite{biswas2019hierarchical},\cite{zhai2020action}

\bibliographystyle{IEEEtran}
\bibliography{allcitations}

% Generated by IEEEtran.bst, version: 1.14 (2015/08/26)
\begin{thebibliography}{100}
\providecommand{\url}[1]{#1}
\csname url@samestyle\endcsname
\providecommand{\newblock}{\relax}
\providecommand{\bibinfo}[2]{#2}
\providecommand{\BIBentrySTDinterwordspacing}{\spaceskip=0pt\relax}
\providecommand{\BIBentryALTinterwordstretchfactor}{4}
\providecommand{\BIBentryALTinterwordspacing}{\spaceskip=\fontdimen2\font plus
\BIBentryALTinterwordstretchfactor\fontdimen3\font minus
  \fontdimen4\font\relax}
\providecommand{\BIBforeignlanguage}[2]{{%
\expandafter\ifx\csname l@#1\endcsname\relax
\typeout{** WARNING: IEEEtran.bst: No hyphenation pattern has been}%
\typeout{** loaded for the language `#1'. Using the pattern for}%
\typeout{** the default language instead.}%
\else
\language=\csname l@#1\endcsname
\fi
#2}}
\providecommand{\BIBdecl}{\relax}
\BIBdecl

\bibitem{jiang2014thumos}
Y.-G. Jiang, J.~Liu, A.~R. Zamir, G.~Toderici, I.~Laptev, M.~Shah, and
  R.~Sukthankar, ``Thumos challenge: Action recognition with a large number of
  classes,'' 2014.

\bibitem{rea2019human}
F.~Rea, A.~Vignolo, A.~Sciutti, and N.~Noceti, ``Human motion understanding for
  selecting action timing in collaborative human-robot interaction,''
  \emph{Front. Robot. AI}, vol.~6, p.~58, 2019.

\bibitem{cioppa2020context}
A.~Cioppa, A.~Deliege, S.~Giancola, B.~Ghanem, M.~V. Droogenbroeck, R.~Gade,
  and T.~B. Moeslund, ``A context-aware loss function for action spotting in
  soccer videos,'' in \emph{Proceedings of the IEEE/CVF Conference on Computer
  Vision and Pattern Recognition}, 2020, pp. 13\,126--13\,136.

\bibitem{rasouli2019autonomous}
A.~Rasouli and J.~K. Tsotsos, ``Autonomous vehicles that interact with
  pedestrians: A survey of theory and practice,'' \emph{IEEE Transactions on
  Intelligent Transportation Systems}, vol.~21, no.~3, pp. 900--918, 2019.

\bibitem{herath2017going}
S.~Herath, M.~Harandi, and F.~Porikli, ``Going deeper into action recognition:
  A survey,'' \emph{Image and vision computing}, vol.~60, pp. 4--21, 2017.

\bibitem{feichtenhofer2019slowfast}
C.~Feichtenhofer, H.~Fan, J.~Malik, and K.~He, ``Slowfast networks for video
  recognition,'' in \emph{Proceedings of the IEEE international conference on
  computer vision}, 2019, pp. 6202--6211.

\bibitem{ghadiyaram2019large}
D.~Ghadiyaram, D.~Tran, and D.~Mahajan, ``Large-scale weakly-supervised
  pre-training for video action recognition,'' in \emph{Proceedings of the IEEE
  Conference on Computer Vision and Pattern Recognition}, 2019, pp.
  12\,046--12\,055.

\bibitem{duan2020omni}
H.~Duan, Y.~Zhao, Y.~Xiong, W.~Liu, and D.~Lin, ``Omni-sourced webly-supervised
  learning for video recognition,'' \emph{arXiv preprint arXiv:2003.13042},
  2020.

\bibitem{idrees2017thumos}
H.~Idrees, A.~R. Zamir, Y.-G. Jiang, A.~Gorban, I.~Laptev, R.~Sukthankar, and
  M.~Shah, ``The thumos challenge on action recognition for videos “in the
  wild”,'' \emph{Computer Vision and Image Understanding}, vol. 155, pp.
  1--23, 2017.

\bibitem{gorban2015thumos}
A.~Gorban, H.~Idrees, Y.-G. Jiang, A.~R. Zamir, I.~Laptev, M.~Shah, and
  R.~Sukthankar, ``Thumos challenge: Action recognition with a large number of
  classes,'' 2015.

\bibitem{kuehne2014language}
H.~Kuehne, A.~Arslan, and T.~Serre, ``The language of actions: Recovering the
  syntax and semantics of goal-directed human activities,'' in
  \emph{Proceedings of the IEEE conference on computer vision and pattern
  recognition}, 2014, pp. 780--787.

\bibitem{lea2017temporal}
C.~Lea, M.~D. Flynn, R.~Vidal, A.~Reiter, and G.~D. Hager, ``Temporal
  convolutional networks for action segmentation and detection,'' in
  \emph{proceedings of the IEEE Conference on Computer Vision and Pattern
  Recognition}, 2017, pp. 156--165.

\bibitem{chao2018rethinking}
Y.-W. Chao, S.~Vijayanarasimhan, B.~Seybold, D.~A. Ross, J.~Deng, and
  R.~Sukthankar, ``Rethinking the faster r-cnn architecture for temporal action
  localization,'' in \emph{Proceedings of the IEEE Conference on Computer
  Vision and Pattern Recognition}, 2018, pp. 1130--1139.

\bibitem{shou2017cdc}
Z.~Shou, J.~Chan, A.~Zareian, K.~Miyazawa, and S.-F. Chang, ``Cdc:
  Convolutional-de-convolutional networks for precise temporal action
  localization in untrimmed videos,'' in \emph{Proceedings of the IEEE
  conference on computer vision and pattern recognition}, 2017, pp. 5734--5743.

\bibitem{gao2019video}
Z.~Gao, L.~Wang, Q.~Zhang, Z.~Niu, N.~Zheng, and G.~Hua, ``Video imprint
  segmentation for temporal action detection in untrimmed videos,'' in
  \emph{Proceedings of the AAAI Conference on Artificial Intelligence},
  vol.~33, 2019, pp. 8328--8335.

\bibitem{long2015fully}
J.~Long, E.~Shelhamer, and T.~Darrell, ``Fully convolutional networks for
  semantic segmentation,'' in \emph{Proceedings of the IEEE conference on
  computer vision and pattern recognition}, 2015, pp. 3431--3440.

\bibitem{everingham2011pascal}
M.~Everingham, L.~Van~Gool, C.~Williams, J.~Winn, and A.~Zisserman, ``The
  pascal visual object classes challenge 2012 (voc2012) results (2012),'' in
  \emph{URL http://www. pascal-network.
  org/challenges/VOC/voc2011/workshop/index. html}, 2011.

\bibitem{zhao2017temporal}
Y.~Zhao, Y.~Xiong, L.~Wang, Z.~Wu, X.~Tang, and D.~Lin, ``Temporal action
  detection with structured segment networks,'' in \emph{Proceedings of the
  IEEE International Conference on Computer Vision}, 2017, pp. 2914--2923.

\bibitem{shou2016temporal}
Z.~Shou, D.~Wang, and S.-F. Chang, ``Temporal action localization in untrimmed
  videos via multi-stage cnns,'' in \emph{Proceedings of the IEEE Conference on
  Computer Vision and Pattern Recognition}, 2016, pp. 1049--1058.

\bibitem{xiong2016cuhk}
Y.~Xiong, L.~Wang, Z.~Wang, B.~Zhang, H.~Song, W.~Li, D.~Lin, Y.~Qiao,
  L.~Van~Gool, and X.~Tang, ``Cuhk \& ethz \& siat submission to activitynet
  challenge 2016,'' \emph{arXiv preprint arXiv:1608.00797}, 2016.

\bibitem{tran2015learning}
D.~Tran, L.~Bourdev, R.~Fergus, L.~Torresani, and M.~Paluri, ``Learning
  spatiotemporal features with 3d convolutional networks,'' in
  \emph{Proceedings of the IEEE international conference on computer vision},
  2015, pp. 4489--4497.

\bibitem{simonyan2014two}
K.~Simonyan and A.~Zisserman, ``Two-stream convolutional networks for action
  recognition in videos,'' in \emph{Advances in neural information processing
  systems}, 2014, pp. 568--576.

\bibitem{wang2016temporal}
L.~Wang, Y.~Xiong, Z.~Wang, Y.~Qiao, D.~Lin, X.~Tang, and L.~Van~Gool,
  ``Temporal segment networks: Towards good practices for deep action
  recognition,'' in \emph{European conference on computer vision}.\hskip 1em
  plus 0.5em minus 0.4em\relax Springer, 2016, pp. 20--36.

\bibitem{carreira2017quo}
J.~Carreira and A.~Zisserman, ``Quo vadis, action recognition? a new model and
  the kinetics dataset,'' in \emph{proceedings of the IEEE Conference on
  Computer Vision and Pattern Recognition}, 2017, pp. 6299--6308.

\bibitem{he2016deep}
K.~He, X.~Zhang, S.~Ren, and J.~Sun, ``Deep residual learning for image
  recognition,'' in \emph{Proceedings of the IEEE conference on computer vision
  and pattern recognition}, 2016, pp. 770--778.

\bibitem{ioffe2015batch}
S.~Ioffe and C.~Szegedy, ``Batch normalization: Accelerating deep network
  training by reducing internal covariate shift,'' \emph{arXiv preprint
  arXiv:1502.03167}, 2015.

\bibitem{gao2017turn}
J.~Gao, Z.~Yang, K.~Chen, C.~Sun, and R.~Nevatia, ``Turn tap: Temporal unit
  regression network for temporal action proposals,'' in \emph{Proceedings of
  the IEEE International Conference on Computer Vision}, 2017, pp. 3628--3636.

\bibitem{gao2017cascaded}
J.~Gao, Z.~Yang, and R.~Nevatia, ``Cascaded boundary regression for temporal
  action detection,'' \emph{arXiv preprint arXiv:1705.01180}, 2017.

\bibitem{ren2015faster}
S.~Ren, K.~He, R.~Girshick, and J.~Sun, ``Faster r-cnn: Towards real-time
  object detection with region proposal networks,'' in \emph{Advances in neural
  information processing systems}, 2015, pp. 91--99.

\bibitem{xu2017r}
H.~Xu, A.~Das, and K.~Saenko, ``R-c3d: Region convolutional 3d network for
  temporal activity detection,'' in \emph{Proceedings of the IEEE international
  conference on computer vision}, 2017, pp. 5783--5792.

\bibitem{li2020graph}
J.~Li, X.~Liu, Z.~Zong, W.~Zhao, M.~Zhang, and J.~Song, ``Graph attention based
  proposal 3d convnets for action detection.'' in \emph{AAAI}, 2020, pp.
  4626--4633.

\bibitem{chen2020afnet}
G.~Chen, C.~Zhang, and Y.~Zou, ``Afnet: Temporal locality-aware network with
  dual structure for accurate and fast action detection,'' \emph{IEEE
  Transactions on Multimedia}, 2020.

\bibitem{gong2020scale}
G.~Gong, L.~Zheng, and Y.~Mu, ``Scale matters: Temporal scale aggregation
  network for precise action localization in untrimmed videos,'' in \emph{2020
  IEEE International Conference on Multimedia and Expo (ICME)}.\hskip 1em plus
  0.5em minus 0.4em\relax IEEE, 2020, pp. 1--6.

\bibitem{liu2016ssd}
W.~Liu, D.~Anguelov, D.~Erhan, C.~Szegedy, S.~Reed, C.-Y. Fu, and A.~C. Berg,
  ``Ssd: Single shot multibox detector,'' in \emph{European conference on
  computer vision}.\hskip 1em plus 0.5em minus 0.4em\relax Springer, 2016, pp.
  21--37.

\bibitem{lin2017single}
T.~Lin, X.~Zhao, and Z.~Shou, ``Single shot temporal action detection,'' in
  \emph{Proceedings of the 25th ACM international conference on Multimedia},
  2017, pp. 988--996.

\bibitem{zhang2018s3d}
D.~Zhang, X.~Dai, X.~Wang, and Y.-F. Wang, ``S3d: single shot multi-span
  detector via fully 3d convolutional networks,'' \emph{arXiv preprint
  arXiv:1807.08069}, 2018.

\bibitem{ronneberger2015u}
O.~Ronneberger, P.~Fischer, and T.~Brox, ``U-net: Convolutional networks for
  biomedical image segmentation,'' in \emph{International Conference on Medical
  image computing and computer-assisted intervention}.\hskip 1em plus 0.5em
  minus 0.4em\relax Springer, 2015, pp. 234--241.

\bibitem{lin2017feature}
T.-Y. Lin, P.~Doll{\'a}r, R.~Girshick, K.~He, B.~Hariharan, and S.~Belongie,
  ``Feature pyramid networks for object detection,'' in \emph{Proceedings of
  the IEEE conference on computer vision and pattern recognition}, 2017, pp.
  2117--2125.

\bibitem{fu2017dssd}
C.-Y. Fu, W.~Liu, A.~Ranga, A.~Tyagi, and A.~C. Berg, ``Dssd: Deconvolutional
  single shot detector,'' \emph{arXiv preprint arXiv:1701.06659}, 2017.

\bibitem{liu2019multi}
Y.~Liu, L.~Ma, Y.~Zhang, W.~Liu, and S.-F. Chang, ``Multi-granularity generator
  for temporal action proposal,'' in \emph{Proceedings of the IEEE Conference
  on Computer Vision and Pattern Recognition}, 2019, pp. 3604--3613.

\bibitem{liuprogressive}
Q.~Liu and Z.~Wang, ``Progressive boundary refinement network for temporal
  action detection.''

\bibitem{gao2020accurate}
J.~Gao, Z.~Shi, G.~Wang, J.~Li, Y.~Yuan, S.~Ge, and X.~Zhou, ``Accurate
  temporal action proposal generation with relation-aware pyramid network.'' in
  \emph{AAAI}, 2020, pp. 10\,810--10\,817.

\bibitem{li2019deep}
X.~Li, T.~Lin, X.~Liu, C.~Gan, W.~Zuo, C.~Li, X.~Long, D.~He, F.~Li, and
  S.~Wen, ``Deep concept-wise temporal convolutional networks for action
  localization,'' \emph{arXiv preprint arXiv:1908.09442}, 2019.

\bibitem{wang2020multi}
X.~Wang, C.~Gao, S.~Zhang, and N.~Sang, ``Multi-level temporal pyramid network
  for action detection,'' in \emph{Chinese Conference on Pattern Recognition
  and Computer Vision (PRCV)}.\hskip 1em plus 0.5em minus 0.4em\relax Springer,
  2020, pp. 41--54.

\bibitem{roerdink2000watershed}
J.~B. Roerdink and A.~Meijster, ``The watershed transform: Definitions,
  algorithms and parallelization strategies,'' \emph{Fundamenta informaticae},
  vol.~41, no. 1, 2, pp. 187--228, 2000.

\bibitem{lin2018bsn}
T.~Lin, X.~Zhao, H.~Su, C.~Wang, and M.~Yang, ``Bsn: Boundary sensitive network
  for temporal action proposal generation,'' in \emph{Proceedings of the
  European Conference on Computer Vision (ECCV)}, 2018, pp. 3--19.

\bibitem{lin2019bmn}
T.~Lin, X.~Liu, X.~Li, E.~Ding, and S.~Wen, ``Bmn: Boundary-matching network
  for temporal action proposal generation,'' in \emph{Proceedings of the IEEE
  International Conference on Computer Vision}, 2019, pp. 3889--3898.

\bibitem{lin2020fast}
C.~Lin, J.~Li, Y.~Wang, Y.~Tai, D.~Luo, Z.~Cui, C.~Wang, J.~Li, F.~Huang, and
  R.~Ji, ``Fast learning of temporal action proposal via dense boundary
  generator.'' in \emph{AAAI}, 2020, pp. 11\,499--11\,506.

\bibitem{bai2020boundary}
Y.~Bai, Y.~Wang, Y.~Tong, Y.~Yang, Q.~Liu, and J.~Liu, ``Boundary content graph
  neural network for temporal action proposal generation,'' \emph{arXiv
  preprint arXiv:2008.01432}, 2020.

\bibitem{yang2020revisiting}
L.~Yang, H.~Peng, D.~Zhang, J.~Fu, and J.~Han, ``Revisiting anchor mechanisms
  for temporal action localization,'' \emph{IEEE Transactions on Image
  Processing}, vol.~29, pp. 8535--8548, 2020.

\bibitem{lin2021learning}
C.~Lin, C.~Xu, D.~Luo, Y.~Wang, Y.~Tai, C.~Wang, J.~Li, F.~Huang, and Y.~Fu,
  ``Learning salient boundary feature for anchor-free temporal action
  localization,'' in \emph{Proceedings of the IEEE/CVF Conference on Computer
  Vision and Pattern Recognition}, 2021, pp. 3320--3329.

\bibitem{gao2018ctap}
J.~Gao, K.~Chen, and R.~Nevatia, ``Ctap: Complementary temporal action proposal
  generation,'' in \emph{Proceedings of the European conference on computer
  vision (ECCV)}, 2018, pp. 68--83.

\bibitem{buch2017sst}
S.~Buch, V.~Escorcia, C.~Shen, B.~Ghanem, and J.~Carlos~Niebles, ``Sst:
  Single-stream temporal action proposals,'' in \emph{Proceedings of the IEEE
  conference on Computer Vision and Pattern Recognition}, 2017, pp. 2911--2920.

\bibitem{buch2019end}
S.~Buch, V.~Escorcia, B.~Ghanem, L.~Fei-Fei, and J.~C. Niebles, ``End-to-end,
  single-stream temporal action detection in untrimmed videos,'' 2019.

\bibitem{yuan2016temporal}
J.~Yuan, B.~Ni, X.~Yang, and A.~A. Kassim, ``Temporal action localization with
  pyramid of score distribution features,'' in \emph{Proceedings of the IEEE
  Conference on Computer Vision and Pattern Recognition}, 2016, pp. 3093--3102.

\bibitem{yeung2016end}
S.~Yeung, O.~Russakovsky, G.~Mori, and L.~Fei-Fei, ``End-to-end learning of
  action detection from frame glimpses in videos,'' in \emph{Proceedings of the
  IEEE Conference on Computer Vision and Pattern Recognition}, 2016, pp.
  2678--2687.

\bibitem{yeung2018every}
S.~Yeung, O.~Russakovsky, N.~Jin, M.~Andriluka, G.~Mori, and L.~Fei-Fei,
  ``Every moment counts: Dense detailed labeling of actions in complex
  videos,'' \emph{International Journal of Computer Vision}, vol. 126, no. 2-4,
  pp. 375--389, 2018.

\bibitem{escorcia2016daps}
V.~Escorcia, F.~C. Heilbron, J.~C. Niebles, and B.~Ghanem, ``Daps: Deep action
  proposals for action understanding,'' in \emph{European Conference on
  Computer Vision}.\hskip 1em plus 0.5em minus 0.4em\relax Springer, 2016, pp.
  768--784.

\bibitem{singh2016multi}
B.~Singh, T.~K. Marks, M.~Jones, O.~Tuzel, and M.~Shao, ``A multi-stream
  bi-directional recurrent neural network for fine-grained action detection,''
  in \emph{Proceedings of the IEEE conference on computer vision and pattern
  recognition}, 2016, pp. 1961--1970.

\bibitem{ma2016learning}
S.~Ma, L.~Sigal, and S.~Sclaroff, ``Learning activity progression in lstms for
  activity detection and early detection,'' in \emph{Proceedings of the IEEE
  Conference on Computer Vision and Pattern Recognition}, 2016, pp. 1942--1950.

\bibitem{zeng2019graph}
R.~Zeng, W.~Huang, M.~Tan, Y.~Rong, P.~Zhao, J.~Huang, and C.~Gan, ``Graph
  convolutional networks for temporal action localization,'' in
  \emph{Proceedings of the IEEE International Conference on Computer Vision},
  2019, pp. 7094--7103.

\bibitem{xu2020g}
M.~Xu, C.~Zhao, D.~S. Rojas, A.~Thabet, and B.~Ghanem, ``G-tad: Sub-graph
  localization for temporal action detection,'' in \emph{Proceedings of the
  IEEE/CVF Conference on Computer Vision and Pattern Recognition}, 2020, pp.
  10\,156--10\,165.

\bibitem{chang2021augmented}
S.~Chang, P.~Wang, F.~Wang, H.~Li, and J.~Feng, ``Augmented transformer with
  adaptive graph for temporal action proposal generation,'' \emph{arXiv
  preprint arXiv:2103.16024}, 2021.

\bibitem{zhao2020video}
C.~Zhao, A.~Thabet, and B.~Ghanem, ``Video self-stitching graph network for
  temporal action localization,'' \emph{arXiv preprint arXiv:2011.14598}, 2020.

\bibitem{nawhal2021activity}
M.~Nawhal and G.~Mori, ``Activity graph transformer for temporal action
  localization,'' \emph{arXiv preprint arXiv:2101.08540}, 2021.

\bibitem{tan2021relaxed}
J.~Tan, J.~Tang, L.~Wang, and G.~Wu, ``Relaxed transformer decoders for direct
  action proposal generation,'' \emph{arXiv preprint arXiv:2102.01894}, 2021.

\bibitem{laptev2007retrieving}
I.~Laptev and P.~P{\'e}rez, ``Retrieving actions in movies,'' in \emph{2007
  IEEE 11th International Conference on Computer Vision}.\hskip 1em plus 0.5em
  minus 0.4em\relax IEEE, 2007, pp. 1--8.

\bibitem{cao2010cross}
L.~Cao, Z.~Liu, and T.~S. Huang, ``Cross-dataset action detection,'' in
  \emph{2010 IEEE Computer Society Conference on Computer Vision and Pattern
  Recognition}.\hskip 1em plus 0.5em minus 0.4em\relax IEEE, 2010, pp.
  1998--2005.

\bibitem{jain2014action}
M.~Jain, J.~Van~Gemert, H.~J{\'e}gou, P.~Bouthemy, and C.~G. Snoek, ``Action
  localization with tubelets from motion,'' in \emph{Proceedings of the IEEE
  conference on computer vision and pattern recognition}, 2014, pp. 740--747.

\bibitem{oneata2014spatio}
D.~Oneata, J.~Revaud, J.~Verbeek, and C.~Schmid, ``Spatio-temporal object
  detection proposals,'' in \emph{European conference on computer
  vision}.\hskip 1em plus 0.5em minus 0.4em\relax Springer, 2014, pp. 737--752.

\bibitem{chen2015action}
W.~Chen and J.~J. Corso, ``Action detection by implicit intentional motion
  clustering,'' in \emph{Proceedings of the IEEE international conference on
  computer vision}, 2015, pp. 3298--3306.

\bibitem{van2015apt}
J.~C. Van~Gemert, M.~Jain, E.~Gati, C.~G. Snoek \emph{et~al.}, ``Apt: Action
  localization proposals from dense trajectories.'' in \emph{BMVC}, vol.~2,
  2015, p.~4.

\bibitem{puscas2015unsupervised}
M.~M. Puscas, E.~Sangineto, D.~Culibrk, and N.~Sebe, ``Unsupervised tube
  extraction using transductive learning and dense trajectories,'' in
  \emph{Proceedings of the IEEE international conference on computer vision},
  2015, pp. 1653--1661.

\bibitem{gkioxari2015finding}
G.~Gkioxari and J.~Malik, ``Finding action tubes,'' in \emph{Proceedings of the
  IEEE conference on computer vision and pattern recognition}, 2015, pp.
  759--768.

\bibitem{hou2017tube}
R.~Hou, C.~Chen, and M.~Shah, ``Tube convolutional neural network (t-cnn) for
  action detection in videos,'' in \emph{Proceedings of the IEEE international
  conference on computer vision}, 2017, pp. 5822--5831.

\bibitem{peng2016multi}
X.~Peng and C.~Schmid, ``Multi-region two-stream r-cnn for action detection,''
  in \emph{European conference on computer vision}.\hskip 1em plus 0.5em minus
  0.4em\relax Springer, 2016, pp. 744--759.

\bibitem{singh2017online}
G.~Singh, S.~Saha, M.~Sapienza, P.~H. Torr, and F.~Cuzzolin, ``Online real-time
  multiple spatiotemporal action localisation and prediction,'' in
  \emph{Proceedings of the IEEE International Conference on Computer Vision},
  2017, pp. 3637--3646.

\bibitem{saha2016deep}
S.~Saha, G.~Singh, M.~Sapienza, P.~H. Torr, and F.~Cuzzolin, ``Deep learning
  for detecting multiple space-time action tubes in videos,'' \emph{arXiv
  preprint arXiv:1608.01529}, 2016.

\bibitem{weinzaepfel2015learning}
P.~Weinzaepfel, Z.~Harchaoui, and C.~Schmid, ``Learning to track for
  spatio-temporal action localization,'' in \emph{Proceedings of the IEEE
  international conference on computer vision}, 2015, pp. 3164--3172.

\bibitem{yang2017spatio}
Z.~Yang, J.~Gao, and R.~Nevatia, ``Spatio-temporal action detection with
  cascade proposal and location anticipation,'' \emph{arXiv preprint
  arXiv:1708.00042}, 2017.

\bibitem{ye2019discovering}
Y.~Ye, X.~Yang, and Y.~Tian, ``Discovering spatio-temporal action tubes,''
  \emph{Journal of Visual Communication and Image Representation}, vol.~58, pp.
  515--524, 2019.

\bibitem{li2018videolstm}
Z.~Li, K.~Gavrilyuk, E.~Gavves, M.~Jain, and C.~G. Snoek, ``Videolstm
  convolves, attends and flows for action recognition,'' \emph{Computer Vision
  and Image Understanding}, vol. 166, pp. 41--50, 2018.

\bibitem{wang2016actionness}
L.~Wang, Y.~Qiao, X.~Tang, and L.~Van~Gool, ``Actionness estimation using
  hybrid fully convolutional networks,'' in \emph{Proceedings of the IEEE
  Conference on Computer Vision and Pattern Recognition}, 2016, pp. 2708--2717.

\bibitem{yu2015fast}
G.~Yu and J.~Yuan, ``Fast action proposals for human action detection and
  search,'' in \emph{Proceedings of the IEEE conference on computer vision and
  pattern recognition}, 2015, pp. 1302--1311.

\bibitem{kalogeiton2017action}
V.~Kalogeiton, P.~Weinzaepfel, V.~Ferrari, and C.~Schmid, ``Action tubelet
  detector for spatio-temporal action localization,'' in \emph{Proceedings of
  the IEEE International Conference on Computer Vision}, 2017, pp. 4405--4413.

\bibitem{gu2018ava}
C.~Gu, C.~Sun, D.~A. Ross, C.~Vondrick, C.~Pantofaru, Y.~Li,
  S.~Vijayanarasimhan, G.~Toderici, S.~Ricco, R.~Sukthankar \emph{et~al.},
  ``Ava: A video dataset of spatio-temporally localized atomic visual
  actions,'' in \emph{Proceedings of the IEEE Conference on Computer Vision and
  Pattern Recognition}, 2018, pp. 6047--6056.

\bibitem{yang2019step}
X.~Yang, X.~Yang, M.-Y. Liu, F.~Xiao, L.~S. Davis, and J.~Kautz, ``Step:
  Spatio-temporal progressive learning for video action detection,'' in
  \emph{Proceedings of the IEEE Conference on Computer Vision and Pattern
  Recognition}, 2019, pp. 264--272.

\bibitem{wu2019long}
C.-Y. Wu, C.~Feichtenhofer, H.~Fan, K.~He, P.~Krahenbuhl, and R.~Girshick,
  ``Long-term feature banks for detailed video understanding,'' in
  \emph{Proceedings of the IEEE Conference on Computer Vision and Pattern
  Recognition}, 2019, pp. 284--293.

\bibitem{sun2018actor}
C.~Sun, A.~Shrivastava, C.~Vondrick, K.~Murphy, R.~Sukthankar, and C.~Schmid,
  ``Actor-centric relation network,'' in \emph{Proceedings of the European
  Conference on Computer Vision (ECCV)}, 2018, pp. 318--334.

\bibitem{zhang2019structured}
Y.~Zhang, P.~Tokmakov, M.~Hebert, and C.~Schmid, ``A structured model for
  action detection,'' in \emph{Proceedings of the IEEE Conference on Computer
  Vision and Pattern Recognition}, 2019, pp. 9975--9984.

\bibitem{girdhar2019video}
R.~Girdhar, J.~Carreira, C.~Doersch, and A.~Zisserman, ``Video action
  transformer network,'' in \emph{Proceedings of the IEEE Conference on
  Computer Vision and Pattern Recognition}, 2019, pp. 244--253.

\bibitem{ulutan2020actor}
O.~Ulutan, S.~Rallapalli, M.~Srivatsa, C.~Torres, and B.~Manjunath, ``Actor
  conditioned attention maps for video action detection,'' in \emph{The IEEE
  Winter Conference on Applications of Computer Vision}, 2020, pp. 527--536.

\bibitem{tomei2019stage}
M.~Tomei, L.~Baraldi, S.~Calderara, S.~Bronzin, and R.~Cucchiara, ``Stage:
  Spatio-temporal attention on graph entities for video action detection,''
  \emph{arXiv preprint arXiv:1912.04316}, 2019.

\bibitem{ji2020action}
J.~Ji, R.~Krishna, L.~Fei-Fei, and J.~C. Niebles, ``Action genome: Actions as
  compositions of spatio-temporal scene graphs,'' in \emph{Proceedings of the
  IEEE/CVF Conference on Computer Vision and Pattern Recognition}, 2020, pp.
  10\,236--10\,247.

\bibitem{pan2020actor}
J.~Pan, S.~Chen, Z.~Shou, J.~Shao, and H.~Li, ``Actor-context-actor relation
  network for spatio-temporal action localization,'' \emph{arXiv preprint
  arXiv:2006.07976}, 2020.

\bibitem{tomei2021video}
M.~Tomei, L.~Baraldi, S.~Calderara, S.~Bronzin, and R.~Cucchiara, ``Video
  action detection by learning graph-based spatio-temporal interactions,''
  \emph{Computer Vision and Image Understanding}, vol. 206, p. 103187, 2021.

\bibitem{sun2015temporal}
C.~Sun, S.~Shetty, R.~Sukthankar, and R.~Nevatia, ``Temporal localization of
  fine-grained actions in videos by domain transfer from web images,'' in
  \emph{Proceedings of the 23rd ACM international conference on Multimedia},
  2015, pp. 371--380.

\bibitem{bojanowski2014weakly}
P.~Bojanowski, R.~Lajugie, F.~Bach, I.~Laptev, J.~Ponce, C.~Schmid, and
  J.~Sivic, ``Weakly supervised action labeling in videos under ordering
  constraints,'' in \emph{European Conference on Computer Vision}.\hskip 1em
  plus 0.5em minus 0.4em\relax Springer, 2014, pp. 628--643.

\bibitem{huang2016connectionist}
D.-A. Huang, L.~Fei-Fei, and J.~C. Niebles, ``Connectionist temporal modeling
  for weakly supervised action labeling,'' in \emph{European Conference on
  Computer Vision}.\hskip 1em plus 0.5em minus 0.4em\relax Springer, 2016, pp.
  137--153.

\bibitem{richard2017weakly}
A.~Richard, H.~Kuehne, and J.~Gall, ``Weakly supervised action learning with
  rnn based fine-to-coarse modeling,'' in \emph{Proceedings of the IEEE
  Conference on Computer Vision and Pattern Recognition}, 2017, pp. 754--763.

\bibitem{kuehne2017weakly}
H.~Kuehne, A.~Richard, and J.~Gall, ``Weakly supervised learning of actions
  from transcripts,'' \emph{Computer Vision and Image Understanding}, vol. 163,
  pp. 78--89, 2017.

\bibitem{narayan20193c}
S.~Narayan, H.~Cholakkal, F.~S. Khan, and L.~Shao, ``3c-net: Category count and
  center loss for weakly-supervised action localization,'' in \emph{Proceedings
  of the IEEE International Conference on Computer Vision}, 2019, pp.
  8679--8687.

\bibitem{schroeter2019weakly}
J.~Schroeter, K.~Sidorov, and D.~Marshall, ``Weakly-supervised temporal
  localization via occurrence count learning,'' \emph{arXiv preprint
  arXiv:1905.07293}, 2019.

\bibitem{wang2017untrimmednets}
L.~Wang, Y.~Xiong, D.~Lin, and L.~Van~Gool, ``Untrimmednets for weakly
  supervised action recognition and detection,'' in \emph{Proceedings of the
  IEEE conference on Computer Vision and Pattern Recognition}, 2017, pp.
  4325--4334.

\bibitem{paul2018w}
S.~Paul, S.~Roy, and A.~K. Roy-Chowdhury, ``W-talc: Weakly-supervised temporal
  activity localization and classification,'' in \emph{Proceedings of the
  European Conference on Computer Vision (ECCV)}, 2018, pp. 563--579.

\bibitem{islam2020weakly}
A.~Islam and R.~Radke, ``Weakly supervised temporal action localization using
  deep metric learning,'' in \emph{The IEEE Winter Conference on Applications
  of Computer Vision}, 2020, pp. 547--556.

\bibitem{carbonneau2018multiple}
M.-A. Carbonneau, V.~Cheplygina, E.~Granger, and G.~Gagnon, ``Multiple instance
  learning: A survey of problem characteristics and applications,''
  \emph{Pattern Recognition}, vol.~77, pp. 329--353, 2018.

\bibitem{rashid2020action}
M.~Rashid, H.~Kjellstrom, and Y.~J. Lee, ``Action graphs: Weakly-supervised
  action localization with graph convolution networks,'' in \emph{The IEEE
  Winter Conference on Applications of Computer Vision}, 2020, pp. 615--624.

\bibitem{jain2020actionbytes}
M.~Jain, A.~Ghodrati, and C.~G. Snoek, ``Actionbytes: Learning from trimmed
  videos to localize actions,'' in \emph{Proceedings of the IEEE/CVF Conference
  on Computer Vision and Pattern Recognition}, 2020, pp. 1171--1180.

\bibitem{gao2020woad}
M.~Gao, Y.~Zhou, R.~Xu, R.~Socher, and C.~Xiong, ``Woad: Weakly supervised
  online action detection in untrimmed videos,'' \emph{arXiv preprint
  arXiv:2006.03732}, 2020.

\bibitem{wen2016discriminative}
Y.~Wen, K.~Zhang, Z.~Li, and Y.~Qiao, ``A discriminative feature learning
  approach for deep face recognition,'' in \emph{European conference on
  computer vision}.\hskip 1em plus 0.5em minus 0.4em\relax Springer, 2016, pp.
  499--515.

\bibitem{huang2020relational}
L.~Huang, Y.~Huang, W.~Ouyang, L.~Wang \emph{et~al.}, ``Relational prototypical
  network for weakly supervised temporal action localization,'' 2020.

\bibitem{nguyen2019weakly}
P.~X. Nguyen, D.~Ramanan, and C.~C. Fowlkes, ``Weakly-supervised action
  localization with background modeling,'' in \emph{Proceedings of the IEEE
  International Conference on Computer Vision}, 2019, pp. 5502--5511.

\bibitem{shou2018autoloc}
Z.~Shou, H.~Gao, L.~Zhang, K.~Miyazawa, and S.-F. Chang, ``Autoloc:
  Weakly-supervised temporal action localization in untrimmed videos,'' in
  \emph{Proceedings of the European Conference on Computer Vision (ECCV)},
  2018, pp. 154--171.

\bibitem{liu2019weakly}
Z.~Liu, L.~Wang, Q.~Zhang, Z.~Gao, Z.~Niu, N.~Zheng, and G.~Hua, ``Weakly
  supervised temporal action localization through contrast based evaluation
  networks,'' in \emph{Proceedings of the IEEE International Conference on
  Computer Vision}, 2019, pp. 3899--3908.

\bibitem{shi2020weakly}
B.~Shi, Q.~Dai, Y.~Mu, and J.~Wang, ``Weakly-supervised action localization by
  generative attention modeling,'' in \emph{Proceedings of the IEEE/CVF
  Conference on Computer Vision and Pattern Recognition}, 2020, pp. 1009--1019.

\bibitem{nguyen2018weakly}
P.~Nguyen, T.~Liu, G.~Prasad, and B.~Han, ``Weakly supervised action
  localization by sparse temporal pooling network,'' in \emph{Proceedings of
  the IEEE Conference on Computer Vision and Pattern Recognition}, 2018, pp.
  6752--6761.

\bibitem{lee2020background}
P.~Lee, Y.~Uh, and H.~Byun, ``Background suppression network for
  weakly-supervised temporal action localization.'' in \emph{AAAI}, 2020, pp.
  11\,320--11\,327.

\bibitem{yuan2019marginalized}
Y.~Yuan, Y.~Lyu, X.~Shen, I.~W. Tsang, and D.-Y. Yeung, ``Marginalized average
  attentional network for weakly-supervised learning,'' \emph{arXiv preprint
  arXiv:1905.08586}, 2019.

\bibitem{liu2019completeness}
D.~Liu, T.~Jiang, and Y.~Wang, ``Completeness modeling and context separation
  for weakly supervised temporal action localization,'' in \emph{Proceedings of
  the IEEE Conference on Computer Vision and Pattern Recognition}, 2019, pp.
  1298--1307.

\bibitem{snell2017prototypical}
J.~Snell, K.~Swersky, and R.~Zemel, ``Prototypical networks for few-shot
  learning,'' in \emph{Advances in neural information processing systems},
  2017, pp. 4077--4087.

\bibitem{sohn2015learning}
K.~Sohn, H.~Lee, and X.~Yan, ``Learning structured output representation using
  deep conditional generative models,'' in \emph{Advances in neural information
  processing systems}, 2015, pp. 3483--3491.

\bibitem{singh2017hide}
K.~K. Singh and Y.~J. Lee, ``Hide-and-seek: Forcing a network to be meticulous
  for weakly-supervised object and action localization,'' in \emph{2017 IEEE
  international conference on computer vision (ICCV)}.\hskip 1em plus 0.5em
  minus 0.4em\relax IEEE, 2017, pp. 3544--3553.

\bibitem{zhong2018step}
J.-X. Zhong, N.~Li, W.~Kong, T.~Zhang, T.~H. Li, and G.~Li, ``Step-by-step
  erasion, one-by-one collection: a weakly supervised temporal action
  detector,'' in \emph{Proceedings of the 26th ACM international conference on
  Multimedia}, 2018, pp. 35--44.

\bibitem{zeng2019breaking}
R.~Zeng, C.~Gan, P.~Chen, W.~Huang, Q.~Wu, and M.~Tan, ``Breaking
  winner-takes-all: Iterative-winners-out networks for weakly supervised
  temporal action localization,'' \emph{IEEE Transactions on Image Processing},
  vol.~28, no.~12, pp. 5797--5808, 2019.

\bibitem{lin2017structured}
Z.~Lin, M.~Feng, C.~N.~d. Santos, M.~Yu, B.~Xiang, B.~Zhou, and Y.~Bengio, ``A
  structured self-attentive sentence embedding,'' \emph{arXiv preprint
  arXiv:1703.03130}, 2017.

\bibitem{sener2018unsupervised}
F.~Sener and A.~Yao, ``Unsupervised learning and segmentation of complex
  activities from video,'' in \emph{Proceedings of the IEEE Conference on
  Computer Vision and Pattern Recognition}, 2018, pp. 8368--8376.

\bibitem{kukleva2019unsupervised}
A.~Kukleva, H.~Kuehne, F.~Sener, and J.~Gall, ``Unsupervised learning of action
  classes with continuous temporal embedding,'' in \emph{Proceedings of the
  IEEE Conference on Computer Vision and Pattern Recognition}, 2019, pp.
  12\,066--12\,074.

\bibitem{gong2020learning}
G.~Gong, X.~Wang, Y.~Mu, and Q.~Tian, ``Learning temporal co-attention models
  for unsupervised video action localization,'' in \emph{Proceedings of the
  IEEE/CVF Conference on Computer Vision and Pattern Recognition}, 2020, pp.
  9819--9828.

\bibitem{chen2020action}
M.-H. Chen, B.~Li, Y.~Bao, G.~AlRegib, and Z.~Kira, ``Action segmentation with
  joint self-supervised temporal domain adaptation,'' in \emph{Proceedings of
  the IEEE/CVF Conference on Computer Vision and Pattern Recognition}, 2020,
  pp. 9454--9463.

\bibitem{ganin2015unsupervised}
Y.~Ganin and V.~Lempitsky, ``Unsupervised domain adaptation by
  backpropagation,'' in \emph{International conference on machine learning},
  2015, pp. 1180--1189.

\bibitem{ganin2016domain}
Y.~Ganin, E.~Ustinova, H.~Ajakan, P.~Germain, H.~Larochelle, F.~Laviolette,
  M.~Marchand, and V.~Lempitsky, ``Domain-adversarial training of neural
  networks,'' \emph{The Journal of Machine Learning Research}, vol.~17, no.~1,
  pp. 2096--2030, 2016.

\bibitem{caba2015activitynet}
F.~Caba~Heilbron, V.~Escorcia, B.~Ghanem, and J.~Carlos~Niebles, ``Activitynet:
  A large-scale video benchmark for human activity understanding,'' in
  \emph{Proceedings of the ieee conference on computer vision and pattern
  recognition}, 2015, pp. 961--970.

\bibitem{zhao2019hacs}
H.~Zhao, A.~Torralba, L.~Torresani, and Z.~Yan, ``Hacs: Human action clips and
  segments dataset for recognition and temporal localization,'' in
  \emph{Proceedings of the IEEE International Conference on Computer Vision},
  2019, pp. 8668--8678.

\bibitem{sigurdsson2016hollywood}
G.~A. Sigurdsson, G.~Varol, X.~Wang, A.~Farhadi, I.~Laptev, and A.~Gupta,
  ``Hollywood in homes: Crowdsourcing data collection for activity
  understanding,'' in \emph{European Conference on Computer Vision}.\hskip 1em
  plus 0.5em minus 0.4em\relax Springer, 2016, pp. 510--526.

\bibitem{stein2013combining}
S.~Stein and S.~J. McKenna, ``Combining embedded accelerometers with computer
  vision for recognizing food preparation activities,'' in \emph{Proceedings of
  the 2013 ACM international joint conference on Pervasive and ubiquitous
  computing}, 2013, pp. 729--738.

\bibitem{rohrbach2016recognizing}
M.~Rohrbach, A.~Rohrbach, M.~Regneri, S.~Amin, M.~Andriluka, M.~Pinkal, and
  B.~Schiele, ``Recognizing fine-grained and composite activities using
  hand-centric features and script data,'' \emph{International Journal of
  Computer Vision}, vol. 119, no.~3, pp. 346--373, 2016.

\bibitem{tang2019coin}
Y.~Tang, D.~Ding, Y.~Rao, Y.~Zheng, D.~Zhang, L.~Zhao, J.~Lu, and J.~Zhou,
  ``Coin: A large-scale dataset for comprehensive instructional video
  analysis,'' in \emph{Proceedings of the IEEE Conference on Computer Vision
  and Pattern Recognition}, 2019, pp. 1207--1216.

\bibitem{ji2019learning}
J.~Ji, K.~Cao, and J.~C. Niebles, ``Learning temporal action proposals with
  fewer labels,'' in \emph{Proceedings of the IEEE International Conference on
  Computer Vision}, 2019, pp. 7073--7082.

\bibitem{tarvainen2017mean}
A.~Tarvainen and H.~Valpola, ``Mean teachers are better role models:
  Weight-averaged consistency targets improve semi-supervised deep learning
  results,'' in \emph{Advances in neural information processing systems}, 2017,
  pp. 1195--1204.

\bibitem{gaidon2011actom}
A.~Gaidon, Z.~Harchaoui, and C.~Schmid, ``Actom sequence models for efficient
  action detection,'' in \emph{CVPR 2011}.\hskip 1em plus 0.5em minus
  0.4em\relax IEEE, 2011, pp. 3201--3208.

\bibitem{gaidon2013temporal}
------, ``Temporal localization of actions with actoms,'' \emph{IEEE
  transactions on pattern analysis and machine intelligence}, vol.~35, no.~11,
  pp. 2782--2795, 2013.

\bibitem{duchenne2009automatic}
O.~Duchenne, I.~Laptev, J.~Sivic, F.~Bach, and J.~Ponce, ``Automatic annotation
  of human actions in video,'' in \emph{2009 IEEE 12th International Conference
  on Computer Vision}.\hskip 1em plus 0.5em minus 0.4em\relax IEEE, 2009, pp.
  1491--1498.

\bibitem{tang2020comprehensive}
Y.~Tang, J.~Lu, and J.~Zhou, ``Comprehensive instructional video analysis: The
  coin dataset and performance evaluation,'' \emph{IEEE Transactions on Pattern
  Analysis and Machine Intelligence}, 2020.

\bibitem{yuan2017temporal}
Z.~Yuan, J.~C. Stroud, T.~Lu, and J.~Deng, ``Temporal action localization by
  structured maximal sums,'' in \emph{Proceedings of the IEEE Conference on
  Computer Vision and Pattern Recognition}, 2017, pp. 3684--3692.

\bibitem{dai2017temporal}
X.~Dai, B.~Singh, G.~Zhang, L.~S. Davis, and Y.~Qiu~Chen, ``Temporal context
  network for activity localization in videos,'' in \emph{Proceedings of the
  IEEE International Conference on Computer Vision}, 2017, pp. 5793--5802.

\bibitem{long2019gaussian}
F.~Long, T.~Yao, Z.~Qiu, X.~Tian, J.~Luo, and T.~Mei, ``Gaussian temporal
  awareness networks for action localization,'' in \emph{Proceedings of the
  IEEE Conference on Computer Vision and Pattern Recognition}, 2019, pp.
  344--353.

\bibitem{eun2019srg}
H.~Eun, S.~Lee, J.~Moon, J.~Park, C.~Jung, and C.~Kim, ``Srg: Snippet
  relatedness-based temporal action proposal generator,'' \emph{IEEE
  Transactions on Circuits and Systems for Video Technology}, 2019.

\bibitem{su2020bsn++}
H.~Su, ``Bsn++: Complementary boundary regressor with scale-balanced relation
  modeling for temporal action proposal generation,'' in \emph{Proceedings of
  the Asian Conference on Computer Vision}, 2020.

\bibitem{zhaobottom}
P.~Zhao, L.~Xie, C.~Ju, Y.~Zhang, Y.~Wang, and Q.~Tian, ``Bottom-up temporal
  action localization with mutual regularization.''

\bibitem{gao2020play}
L.~Gao, T.~Li, J.~Song, Z.~Zhao, and H.~T. Shen, ``Play and rewind:
  Context-aware video temporal action proposals,'' \emph{Pattern Recognition},
  p. 107477, 2020.

\bibitem{liu2021end}
X.~Liu, Q.~Wang, Y.~Hu, X.~Tang, S.~Bai, and X.~Bai, ``End-to-end temporal
  action detection with transformer,'' \emph{arXiv preprint arXiv:2106.10271},
  2021.

\bibitem{liu2020progressive}
Q.~Liu and Z.~Wang, ``Progressive boundary refinement network for temporal
  action detection.'' in \emph{AAAI}, 2020, pp. 11\,612--11\,619.

\bibitem{alwassel2020tsp}
H.~Alwassel, S.~Giancola, and B.~Ghanem, ``Tsp: Temporally-sensitive
  pretraining of video encoders for localization tasks,'' \emph{arXiv preprint
  arXiv:2011.11479}, 2020.

\bibitem{wang2021rgb}
C.~Wang, H.~Cai, Y.~Zou, and Y.~Xiong, ``Rgb stream is enough for temporal
  action detection,'' \emph{arXiv preprint arXiv:2107.04362}, 2021.

\bibitem{wu2021towards}
J.~Wu, P.~Sun, S.~Chen, J.~Yang, Z.~Qi, L.~Ma, and P.~Luo, ``Towards
  high-quality temporal action detection with sparse proposals,'' \emph{arXiv
  preprint arXiv:2109.08847}, 2021.

\bibitem{liu2021multi}
X.~Liu, Y.~Hu, S.~Bai, F.~Ding, X.~Bai, and P.~H. Torr, ``Multi-shot temporal
  event localization: a benchmark,'' in \emph{Proceedings of the IEEE/CVF
  Conference on Computer Vision and Pattern Recognition}, 2021, pp.
  12\,596--12\,606.

\bibitem{xu2019segregated}
Y.~Xu, C.~Zhang, Z.~Cheng, J.~Xie, Y.~Niu, S.~Pu, and F.~Wu, ``Segregated
  temporal assembly recurrent networks for weakly supervised multiple action
  detection,'' in \emph{Proceedings of the AAAI Conference on Artificial
  Intelligence}, vol.~33, 2019, pp. 9070--9078.

\bibitem{zhang2020adapnet}
X.-Y. Zhang, C.~Li, H.~Shi, X.~Zhu, P.~Li, and J.~Dong, ``Adapnet: Adaptability
  decomposing encoder-decoder network for weakly supervised action recognition
  and localization,'' \emph{IEEE transactions on neural networks and learning
  systems}, 2020.

\bibitem{yu2019temporal}
T.~Yu, Z.~Ren, Y.~Li, E.~Yan, N.~Xu, and J.~Yuan, ``Temporal structure mining
  for weakly supervised action detection,'' in \emph{Proceedings of the IEEE
  International Conference on Computer Vision}, 2019, pp. 5522--5531.

\bibitem{shen2020weakly}
Z.~Shen, F.~Wang, and J.~Dai, ``Weakly supervised temporal action localization
  by multi-stage fusion network,'' \emph{IEEE Access}, vol.~8, pp.
  17\,287--17\,298, 2020.

\bibitem{zhai2020two}
Y.~Zhai, L.~Wang, W.~Tang, Q.~Zhang, J.~Yuan, and G.~Hua, ``Two-stream
  consensus network for weakly-supervised temporal action localization,'' in
  \emph{European conference on computer vision}.\hskip 1em plus 0.5em minus
  0.4em\relax Springer, 2020, pp. 37--54.

\bibitem{yang2020equivalent}
L.~Yang, D.~Zhang, T.~Zhao, and J.~Han, ``Equivalent classification mapping for
  weakly supervised temporal action localization,'' \emph{arXiv preprint
  arXiv:2008.07728}, 2020.

\bibitem{min2020adversarial}
K.~Min and J.~J. Corso, ``Adversarial background-aware loss for
  weakly-supervised temporal activity localization,'' \emph{arXiv preprint
  arXiv:2007.06643}, 2020.

\bibitem{luo2020weakly}
Z.~Luo, D.~Guillory, B.~Shi, W.~Ke, F.~Wan, T.~Darrell, and H.~Xu,
  ``Weakly-supervised action localization with expectation-maximization
  multi-instance learning,'' \emph{arXiv preprint arXiv:2004.00163}, 2020.

\bibitem{lee2020backgrounduncertain}
P.~Lee, J.~Wang, Y.~Lu, and H.~Byun, ``Background modeling via uncertainty
  estimation for weakly-supervised action localization,'' \emph{arXiv preprint
  arXiv:2006.07006}, 2020.

\bibitem{ma2021weakly}
J.~Ma, S.~K. Gorti, M.~Volkovs, and G.~Yu, ``Weakly supervised action selection
  learning in video,'' in \emph{Proceedings of the IEEE/CVF Conference on
  Computer Vision and Pattern Recognition}, 2021, pp. 7587--7596.

\bibitem{huang2021modeling}
L.~Huang, Y.~Huang, W.~Ouyang, and L.~Wang, ``Modeling sub-actions for weakly
  supervised temporal action localization,'' \emph{IEEE Transactions on Image
  Processing}, vol.~30, pp. 5154--5167, 2021.

\bibitem{ding2020weakly}
X.~Ding, N.~Wang, X.~Gao, J.~Li, X.~Wang, and T.~Liu, ``Weakly supervised
  temporal action localization with segment-level labels,'' \emph{arXiv
  preprint arXiv:2007.01598}, 2020.

\bibitem{zhang2021cola}
C.~Zhang, M.~Cao, D.~Yang, J.~Chen, and Y.~Zou, ``Cola: Weakly-supervised
  temporal action localization with snippet contrastive learning,'' in
  \emph{Proceedings of the IEEE/CVF Conference on Computer Vision and Pattern
  Recognition}, 2021, pp. 16\,010--16\,019.

\bibitem{liu2021acsnet}
Z.~Liu, L.~Wang, Q.~Zhang, W.~Tang, J.~Yuan, N.~Zheng, and G.~Hua, ``Acsnet:
  Action-context separation network for weakly supervised temporal action
  localization,'' \emph{arXiv preprint arXiv:2103.15088}, 2021.

\bibitem{lee2020weakly}
P.~Lee, J.~Wang, Y.~Lu, and H.~Byun, ``Weakly-supervised temporal action
  localization by uncertainty modeling,'' \emph{arXiv preprint
  arXiv:2006.07006}, 2020.

\bibitem{qu2021acm}
S.~Qu, G.~Chen, Z.~Li, L.~Zhang, F.~Lu, and A.~Knoll, ``Acm-net: Action context
  modeling network for weakly-supervised temporal action localization,''
  \emph{arXiv preprint arXiv:2104.02967}, 2021.

\bibitem{narayan2020d2}
S.~Narayan, H.~Cholakkal, M.~Hayat, F.~S. Khan, M.-H. Yang, and L.~Shao,
  ``D2-net: Weakly-supervised action localization via discriminative embeddings
  and denoised activations,'' \emph{arXiv preprint arXiv:2012.06440}, 2020.

\bibitem{lin2019towards}
X.~Lin, Z.~Shou, and S.-F. Chang, ``Towards train-test consistency for
  semi-supervised temporal action localization,'' \emph{arXiv preprint
  arXiv:1910.11285}, 2019.

\bibitem{gongself}
G.~Gong, L.~Zheng, W.~Jiang, and Y.~Mu, ``Self-supervised video action
  localization with adversarial temporal transforms.''

\bibitem{rohrbach2012database}
M.~Rohrbach, S.~Amin, M.~Andriluka, and B.~Schiele, ``A database for fine
  grained activity detection of cooking activities,'' in \emph{2012 IEEE
  Conference on Computer Vision and Pattern Recognition}.\hskip 1em plus 0.5em
  minus 0.4em\relax IEEE, 2012, pp. 1194--1201.

\bibitem{nadolski2005optimizing}
R.~J. Nadolski, P.~A. Kirschner, and J.~J. Van~Merri{\"e}nboer, ``Optimizing
  the number of steps in learning tasks for complex skills,'' \emph{British
  Journal of Educational Psychology}, vol.~75, no.~2, pp. 223--237, 2005.

\bibitem{damen2018scaling}
D.~Damen, H.~Doughty, G.~Maria~Farinella, S.~Fidler, A.~Furnari, E.~Kazakos,
  D.~Moltisanti, J.~Munro, T.~Perrett, W.~Price \emph{et~al.}, ``Scaling
  egocentric vision: The epic-kitchens dataset,'' in \emph{Proceedings of the
  European Conference on Computer Vision (ECCV)}, 2018, pp. 720--736.

\bibitem{alayrac2016unsupervised}
J.-B. Alayrac, P.~Bojanowski, N.~Agrawal, J.~Sivic, I.~Laptev, and
  S.~Lacoste-Julien, ``Unsupervised learning from narrated instruction
  videos,'' in \emph{Proceedings of the IEEE Conference on Computer Vision and
  Pattern Recognition}, 2016, pp. 4575--4583.

\bibitem{chandola2007outlier}
V.~Chandola, A.~Banerjee, and V.~Kumar, ``Outlier detection: A survey,''
  \emph{ACM Computing Surveys}, vol.~14, p.~15, 2007.

\bibitem{chalapathy2019deep}
R.~Chalapathy and S.~Chawla, ``Deep learning for anomaly detection: A survey,''
  \emph{arXiv preprint arXiv:1901.03407}, 2019.

\bibitem{li2013anomaly}
W.~Li, V.~Mahadevan, and N.~Vasconcelos, ``Anomaly detection and localization
  in crowded scenes,'' \emph{IEEE transactions on pattern analysis and machine
  intelligence}, vol.~36, no.~1, pp. 18--32, 2013.

\bibitem{zhu2012context}
Y.~Zhu, N.~M. Nayak, and A.~K. Roy-Chowdhury, ``Context-aware activity
  recognition and anomaly detection in video,'' \emph{IEEE Journal of Selected
  Topics in Signal Processing}, vol.~7, no.~1, pp. 91--101, 2012.

\bibitem{sultani2018real}
W.~Sultani, C.~Chen, and M.~Shah, ``Real-world anomaly detection in
  surveillance videos,'' in \emph{Proceedings of the IEEE Conference on
  Computer Vision and Pattern Recognition}, 2018, pp. 6479--6488.

\bibitem{he2018anomaly}
C.~He, J.~Shao, and J.~Sun, ``An anomaly-introduced learning method for
  abnormal event detection,'' \emph{Multimedia Tools and Applications},
  vol.~77, no.~22, pp. 29\,573--29\,588, 2018.

\bibitem{sabokrou2018deep}
M.~Sabokrou, M.~Fayyaz, M.~Fathy, Z.~Moayed, and R.~Klette, ``Deep-anomaly:
  Fully convolutional neural network for fast anomaly detection in crowded
  scenes,'' \emph{Computer Vision and Image Understanding}, vol. 172, pp.
  88--97, 2018.

\bibitem{sabokrou2017deep}
M.~Sabokrou, M.~Fayyaz, M.~Fathy, and R.~Klette, ``Deep-cascade: Cascading 3d
  deep neural networks for fast anomaly detection and localization in crowded
  scenes,'' \emph{IEEE Transactions on Image Processing}, vol.~26, no.~4, pp.
  1992--2004, 2017.

\bibitem{sabokrou2018avid}
M.~Sabokrou, M.~Pourreza, M.~Fayyaz, R.~Entezari, M.~Fathy, J.~Gall, and
  E.~Adeli, ``Avid: Adversarial visual irregularity detection,'' in \emph{Asian
  Conference on Computer Vision}.\hskip 1em plus 0.5em minus 0.4em\relax
  Springer, 2018, pp. 488--505.

\bibitem{liu2019exploring}
K.~Liu and H.~Ma, ``Exploring background-bias for anomaly detection in
  surveillance videos,'' in \emph{Proceedings of the 27th ACM International
  Conference on Multimedia}, 2019, pp. 1490--1499.

\bibitem{xu2019temporal}
M.~Xu, M.~Gao, Y.-T. Chen, L.~S. Davis, and D.~J. Crandall, ``Temporal
  recurrent networks for online action detection,'' in \emph{Proceedings of the
  IEEE International Conference on Computer Vision}, 2019, pp. 5532--5541.

\bibitem{bettadapura2016leveraging}
V.~Bettadapura, C.~Pantofaru, and I.~Essa, ``Leveraging contextual cues for
  generating basketball highlights,'' in \emph{Proceedings of the 24th ACM
  international conference on Multimedia}, 2016, pp. 908--917.

\bibitem{heilbron2017scc}
F.~C. Heilbron, W.~Barrios, V.~Escorcia, and B.~Ghanem, ``Scc: Semantic context
  cascade for efficient action detection,'' in \emph{2017 IEEE Conference on
  Computer Vision and Pattern Recognition (CVPR)}.\hskip 1em plus 0.5em minus
  0.4em\relax IEEE, 2017, pp. 3175--3184.

\bibitem{felsen2017will}
P.~Felsen, P.~Agrawal, and J.~Malik, ``What will happen next? forecasting
  player moves in sports videos,'' in \emph{Proceedings of the IEEE
  international conference on computer vision}, 2017, pp. 3342--3351.

\bibitem{kapela2014real}
R.~Kapela, K.~McGuinness, A.~Swietlicka, and N.~E. O’Connor, ``Real-time
  event detection in field sport videos,'' in \emph{Computer vision in
  Sports}.\hskip 1em plus 0.5em minus 0.4em\relax Springer, 2014, pp. 293--316.

\bibitem{cioppa2018bottom}
A.~Cioppa, A.~Deliege, and M.~Van~Droogenbroeck, ``A bottom-up approach based
  on semantics for the interpretation of the main camera stream in soccer
  games,'' in \emph{Proceedings of the IEEE Conference on Computer Vision and
  Pattern Recognition Workshops}, 2018, pp. 1765--1774.

\bibitem{tsunoda2017football}
T.~Tsunoda, Y.~Komori, M.~Matsugu, and T.~Harada, ``Football action recognition
  using hierarchical lstm,'' in \emph{Proceedings of the IEEE conference on
  computer vision and pattern recognition workshops}, 2017, pp. 99--107.

\bibitem{cai2019temporal}
Z.~Cai, H.~Neher, K.~Vats, D.~A. Clausi, and J.~Zelek, ``Temporal hockey action
  recognition via pose and optical flows,'' in \emph{Proceedings of the IEEE
  Conference on Computer Vision and Pattern Recognition Workshops}, 2019, pp.
  0--0.

\bibitem{sanabria2019deep}
M.~Sanabria, F.~Precioso, and T.~Menguy, ``A deep architecture for multimodal
  summarization of soccer games,'' in \emph{Proceedings Proceedings of the 2nd
  International Workshop on Multimedia Content Analysis in Sports}, 2019, pp.
  16--24.

\bibitem{shukla2018automatic}
P.~Shukla, H.~Sadana, A.~Bansal, D.~Verma, C.~Elmadjian, B.~Raman, and M.~Turk,
  ``Automatic cricket highlight generation using event-driven and
  excitement-based features,'' in \emph{Proceedings of the IEEE Conference on
  Computer Vision and Pattern Recognition Workshops}, 2018, pp. 1800--1808.

\bibitem{tsagkatakis2017goal}
G.~Tsagkatakis, M.~Jaber, and P.~Tsakalides, ``Goal!! event detection in sports
  video,'' \emph{Electronic Imaging}, vol. 2017, no.~16, pp. 15--20, 2017.

\bibitem{turchini2019flexible}
F.~Turchini, L.~Seidenari, L.~Galteri, A.~Ferracani, G.~Becchi, and
  A.~Del~Bimbo, ``Flexible automatic football filming and summarization,'' in
  \emph{Proceedings Proceedings of the 2nd International Workshop on Multimedia
  Content Analysis in Sports}, 2019, pp. 108--114.

\bibitem{giancola2018soccernet}
S.~Giancola, M.~Amine, T.~Dghaily, and B.~Ghanem, ``Soccernet: A scalable
  dataset for action spotting in soccer videos,'' in \emph{Proceedings of the
  IEEE Conference on Computer Vision and Pattern Recognition Workshops}, 2018,
  pp. 1711--1721.

\bibitem{huang2006semantic}
C.-L. Huang, H.-C. Shih, and C.-Y. Chao, ``Semantic analysis of soccer video
  using dynamic bayesian network,'' \emph{IEEE Transactions on Multimedia},
  vol.~8, no.~4, pp. 749--760, 2006.

\bibitem{fontana2018action}
V.~Fontana, G.~Singh, S.~Akrigg, M.~Di~Maio, S.~Saha, and F.~Cuzzolin, ``Action
  detection from a robot-car perspective,'' \emph{arXiv preprint
  arXiv:1807.11332}, 2018.

\bibitem{yao2020and}
Y.~Yao, X.~Wang, M.~Xu, Z.~Pu, E.~Atkins, and D.~Crandall, ``When, where, and
  what? a new dataset for anomaly detection in driving videos,'' \emph{arXiv
  preprint arXiv:2004.03044}, 2020.

\bibitem{mahadevan2019av}
K.~Mahadevan, E.~Sanoubari, S.~Somanath, J.~E. Young, and E.~Sharlin,
  ``Av-pedestrian interaction design using a pedestrian mixed traffic
  simulator,'' in \emph{Proceedings of the 2019 on Designing Interactive
  Systems Conference}, 2019, pp. 475--486.

\bibitem{cao2019action}
D.~Cao, L.~Xu, and H.~Chen, ``Action recognition in untrimmed videos with
  composite self-attention two-stream framework,'' in \emph{Asian Conference on
  Pattern Recognition}.\hskip 1em plus 0.5em minus 0.4em\relax Springer, 2019,
  pp. 27--40.

\bibitem{shi2019weakly}
H.~Shi, X.~Zhang, and C.~Li, ``Weakly-supervised action recognition and
  localization via knowledge transfer,'' in \emph{Chinese Conference on Pattern
  Recognition and Computer Vision (PRCV)}.\hskip 1em plus 0.5em minus
  0.4em\relax Springer, 2019, pp. 205--216.

\bibitem{zhang2020zstad}
L.~Zhang, X.~Chang, J.~Liu, M.~Luo, S.~Wang, Z.~Ge, and A.~Hauptmann, ``Zstad:
  Zero-shot temporal activity detection,'' in \emph{Proceedings of the IEEE/CVF
  Conference on Computer Vision and Pattern Recognition}, 2020, pp. 879--888.

\bibitem{xu2018similarity}
H.~Xu, B.~Kang, X.~Sun, J.~Feng, K.~Saenko, and T.~Darrell, ``Similarity r-c3d
  for few-shot temporal activity detection,'' \emph{arXiv preprint
  arXiv:1812.10000}, 2018.

\bibitem{xu2020revisiting}
H.~Xu, X.~Sun, E.~Tzeng, A.~Das, K.~Saenko, and T.~Darrell, ``Revisiting
  few-shot activity detection with class similarity control,'' \emph{arXiv
  preprint arXiv:2004.00137}, 2020.

\bibitem{yang2018one}
H.~Yang, X.~He, and F.~Porikli, ``One-shot action localization by learning
  sequence matching network,'' in \emph{Proceedings of the IEEE Conference on
  Computer Vision and Pattern Recognition}, 2018, pp. 1450--1459.

\bibitem{huang2019decoupling}
Y.~Huang, Q.~Dai, and Y.~Lu, ``Decoupling localization and classification in
  single shot temporal action detection,'' in \emph{2019 IEEE International
  Conference on Multimedia and Expo (ICME)}.\hskip 1em plus 0.5em minus
  0.4em\relax IEEE, 2019, pp. 1288--1293.

\bibitem{zhang2020metal}
D.~Zhang, X.~Dai, and Y.-F. Wang, ``Metal: Minimum effort temporal activity
  localization in untrimmed videos,'' in \emph{Proceedings of the IEEE/CVF
  Conference on Computer Vision and Pattern Recognition}, 2020, pp. 3882--3892.

\end{thebibliography}
% \bibliography{reference}
\end{document}